\documentclass[preprint,12pt,authoryear]{elsarticle}

\usepackage[utf8]{inputenc}
\usepackage[T1]{fontenc}
\usepackage{amsmath,amssymb,amsfonts}
\usepackage{booktabs}
\usepackage{graphicx}
\usepackage[table]{xcolor}
\usepackage{hyperref}
\usepackage{algorithm}
\usepackage{algorithmic}
\usepackage{multirow}
\usepackage{bbm}
\usepackage{siunitx}
\usepackage{subcaption}
\usepackage{tabularx}
\usepackage{makecell}
\usepackage{pdflscape}
\usepackage{tikz}
\usetikzlibrary{positioning, arrows.meta, calc, fit, backgrounds, shapes.geometric, decorations.pathreplacing}

\graphicspath{{figures/}}

\hypersetup{
  colorlinks=true,
  linkcolor=blue!60!black,
  citecolor=blue!60!black,
  urlcolor=blue!60!black,
}

\journal{Computers \& Chemical Engineering}

\begin{document}

\begin{frontmatter}

% --- Title ---
\title{UTOPYA: A Multimodal Deep Learning Framework for Physics-Informed Anomaly Detection and Time-Series Prediction}

% --- Author ---
\author[ntnu]{Robson W. S. Pessoa\corref{cor1}}
\author[icl]{Julien Amblard}
\author[icl]{Alessandra Russo}
\author[ntnu]{Idelfonso B.R. Nogueira\corref{cor1}}
\ead{idelfonso.nogueira@ntnu.no}
\cortext[cor1]{Corresponding author}
\affiliation[ntnu]{organization={Department of Chemical Engineering,
  Norwegian University of Science and Technology (NTNU)},
  city={Trondheim},
  country={Norway}}
\affiliation[icl]{organization={Department of Computing, Imperial College London},
  city={London},
  country={United Kingdom}}

% --- Abstract ---
\begin{abstract}
Anomaly detection in batch processes is hindered by transient dynamics,
scarce fault labels, and reliance on single-modality sensor data. This work
introduces UTOPYA (Unified Temporal Observation for Physics-Informed Anomaly
Detection and Time-Series Prediction), a 15.2M-parameter multimodal
framework that jointly addresses anomaly detection, time-series prediction,
and phase classification in batch distillation by fusing eight data modalities
through 
Feature-wise Linear Modulation (FiLM) conditioned 
cross-modal attention and gated fusion. A
physics-informed regularisation scheme introduced in this work enforces temporal
smoothness and thermodynamic monotonicity, while curriculum learning introduces
training samples in order of physical difficulty. On the 119-experiment multimodal
batch distillation dataset of \citet{arweiler2026batch}, UTOPYA achieves a
window-level test AUROC of 0.832 and 0.874 under multi-signal
experiment-level scoring, substantially outperforming four external baselines
(PCA, autoencoder, Isolation Forest, and LSTM autoencoder) evaluated under
identical conditions ($+0.147$ window-level AUROC over the best baseline). A multimodal ablation
over 15~architectural configurations shows that static context via FiLM
conditioning is the key enabler, lifting experiment-level multi-signal AUROC
by $+0.145$ over the unimodal baseline (0.729 to 0.874). Separately, a
training ablation across 14~design choices reveals that several widely-adopted
techniques, including instance normalisation, Mixup, ensembling, test-time
augmentation, and stochastic weight averaging, fail to improve or actively
degrade generalisation in this data-scarce setting. These negative results
expose a fundamental tension between smoothing-based regularisation and
anomaly detection, providing practical guidance for multimodal process
monitoring deployment.
\end{abstract}

% --- Keywords ---

\end{frontmatter}

% =========================================================================
\section{Introduction}
\label{sec:introduction}
% =========================================================================

Detecting anomalous operation in industrial processes is a long-standing challenge for the process engineering community. Classical statistical process monitoring relies on multivariate methods such as Principal Component Analysis and Partial Least Squares \citep{ge2013review,venkatasubramanian2003review3}, which establish a model of normal operating conditions and flag deviations through monitoring statistics. While effective for steady-state continuous processes, these methods struggle when the process is inherently transient, because the normal operating envelope itself changes with time.

The challenge becomes even more acute when anomaly detection is coupled with simultaneous time-series prediction. One approach to detect faults is to predict future sensor trajectories and compare them against observed values; a well-calibrated predictive model will produce large residuals precisely when the process deviates from expected behaviour. However, this is particularly challenging in transient systems. Temperatures, pressures, and compositions evolve non-stationarily through distinct phases, so the model must learn phase-dependent dynamics rather than a single stationary pattern \citep{venkatasubramanian2003review1}.

Several studies have applied data-driven deep learning methods to process fault detection \citep{WANG2018144,choi2021deep,pang2021deep,aldrich2020unsupervised}. Among those methods, autoencoders, variational autoencoders, and recurrent neural networks have been applied to learn representations of normal behaviour and flag deviations. These methods are almost exclusively benchmarked on continuous processes using a single data modality, most commonly time-series sensor readings on standard problems such as the Tennessee Eastman Process \citep{downs1993plant}. Two limitations persist. Purely data-driven models tend to overfit when labelled fault data is scarce, and single-modality approaches discard information from other sensing channels.

Two sources of complementary knowledge can help overcome these limitations. First, \emph{multimodal fusion} enables the model to exploit heterogeneous data streams---process sensors, visual inspection, acoustic signatures, spectroscopic analysis---that are individually informative but collectively richer \citep{baltrušaitis2019multimodal,gao2020survey}. This is what human operators routinely do when diagnosing faults. In the chemical process industry, Process Analytical Technology has driven the adoption of diverse sensing modalities \citep{simon2015assessment}. However, data from different instruments are still typically analysed by separate domain-specific models \citep{ge2013review,jiang2023review}. Fusion strategies can overcome this separation, ranging from early concatenation of raw features to intermediate-level interaction of learned representations \citep{baltrušaitis2019multimodal}. Recent conditioning mechanisms such as Feature-wise Linear Modulation \citep[FiLM;][]{perez2018film} offer a way to modulate dynamic features with static contextual information.

Second, \emph{physics-informed regularization} encodes domain knowledge as soft constraints in the loss function \citep{raissi2019physics}. Physics-informed neural networks (PINNs) have been applied in chemical engineering to reaction kinetics, transport phenomena, and process optimisation. The underlying principle --- that physical processes are governed by conservation equations that impose continuity on state variables --- translates naturally to batch distillation, where thermal and hydraulic inertia ensure that temperatures, pressures, and flow rates evolve smoothly over time. The specific physics-informed formulation introduced in this work, which combines temporal smoothness and thermodynamic monotonicity as soft constraints on predicted trajectories, is described in Section~\ref{sec:physics}.

Despite substantial work in both areas, no prior study jointly exploits multimodal fusion and physics-informed regularization for anomaly detection in batch processes. Batch distillation is a workhorse unit operation in the chemical and pharmaceutical industries \citep{diwekar1995batch,stichlmair2010distillation}. It is valued for its flexibility in separating multicomponent mixtures without a continuous feed, but its transient dynamics across startup, operation, and shutdown phases make steady-state monitoring tools inadequate. Labelled fault data is difficult to obtain. Deliberately inducing anomalies on production equipment is costly and real faults are rare events \citep{crowl2019chemical}.

The recently published dataset of \citet{arweiler2026batch} opens precisely this opportunity. It comprises 119 experiments on a laboratory-scale distillation plant, covering eight data modalities (time-series, images, audio, video, NMR spectra, GC analysis, tabular metadata, and text logs) and
including in total around
86.3~GB of raw data across three chemical systems. Each anomalous experiment is paired with a fault-free reference conducted under identical nominal conditions, providing a rigorous evaluation framework. To the best of our knowledge, this is the first dataset to offer paired normal and anomalous batch distillation experiments with multimodal coverage.

In this work, we present \textbf{UTOPYA} (\textbf{U}nified \textbf{T}emporal \textbf{O}bservation for \textbf{P}h\textbf{y}sics-informed \textbf{A}nomaly detection and time-series prediction), a multimodal framework that jointly addresses anomaly detection, time-series prediction, and phase classification in industrial batch processes. UTOPYA integrates two complementary forms of domain knowledge: a physics-informed regularisation scheme that enforces temporal smoothness and thermodynamic monotonicity in predicted time series; and curriculum learning, which introduces training samples in order of physical difficulty to improve generalisation in data-scarce settings. Our contributions are:

\begin{enumerate}
  \item UTOPYA, a multimodal deep learning architecture that fuses eight
    data modalities via FiLM conditioning, bidirectional cross-modal
    attention, and gated fusion with graceful degradation for absent
    modalities, under a multi-task objective spanning anomaly detection,
    time-series prediction, and phase classification.
  \item A physics-informed regularisation scheme for batch distillation
    that encodes temporal smoothness and 
    distillation column temperature profile monotonicity
    constraints as soft penalties on predicted trajectories.
  \item A curriculum learning strategy that exploits the natural difficulty
    hierarchy of anomaly detection in batch processes, combined with
    self-supervised pretraining using block masking and contrastive learning,
    and per-experiment normalisation that removes operating-point-specific
    baselines to improve cross-experiment generalisation.
  \item A systematic ablation study comprising both a multimodal ablation
    that isolates the contribution of each data modality (time-series, GC,
    audio, static context) and a training ablation spanning 14 design
    decisions, providing quantitative evidence on which techniques improve
    generalization and which do not, including important negative results
    (RevIN, Mixup, ensembling, test-time augmentation, stochastic weight
    averaging) and the discovery of an inverse val--test correlation that
    curriculum learning partially resolves.
  \item A comparison with four external baselines (PCA, feedforward
    autoencoder, Isolation Forest, and LSTM autoencoder) evaluated under
    identical conditions, demonstrating that UTOPYA substantially
    outperforms standard methods on this challenging dataset.
\end{enumerate}

The remainder of this paper is organised as follows.
Section~\ref{sec:dataset} describes the Arweiler et al.\ dataset.
Section~\ref{sec:methodology} details the UTOPYA architecture and training
methodology. Section~\ref{sec:experimental_setup} presents the experimental
setup. Section~\ref{sec:results} reports results, ablation findings, and
negative results. Section~\ref{sec:conclusion} concludes with lessons learned
and directions for future work.

% =========================================================================
\section{Dataset}
\label{sec:dataset}
% =========================================================================

We employ the multimodal batch distillation dataset published by
\citet{arweiler2026batch}, available on Zenodo (record 18771181). This
dataset was collected on a laboratory-scale batch distillation plant and
comprises 119 experiments across three chemical systems, summarised in Table~\ref{tab:chemical_systems}.

\begin{table}[htbp]
  \centering
  \caption{Chemical systems in the Arweiler et al.\ (2025) dataset.}
  \label{tab:chemical_systems}
  \begin{tabular}{@{}llc@{}}
    \toprule
    System & Components & Experiments \\
    \midrule
    Ternary I  & butan-1-ol + propan-2-ol + water   & $\sim$91 \\
    Ternary II & acetone + butan-1-ol + methanol     & $\sim$14 \\
    Binary     & ethanol + propan-2-ol               & $\sim$14 \\
    \bottomrule
  \end{tabular}
\end{table}

The total dataset size is approximately 86.3~GB, distributed across eight
modalities:

\begin{enumerate}
  \item \textbf{Time-series process variables}: 19 sensors and 13 actuators
    sampled at 1~Hz, including temperatures at multiple column positions,
    pressures, flow rates, liquid levels, and control valve states.
  \item \textbf{Images}: Three camera streams capturing the column, reboiler,
    and condenser regions.
  \item \textbf{Audio}: Acoustic recordings from the plant environment.
  \item \textbf{Video}: Continuous video recordings of column operation.
  \item \textbf{NMR spectra}: Nuclear magnetic resonance spectra of product
    samples collected at discrete intervals.
  \item \textbf{GC analysis}: Gas chromatography composition measurements.
  \item \textbf{Static tabular features}: Experiment-level metadata including
    operating conditions, equipment configuration, and chemical system
    properties.
  \item \textbf{Text metadata}: Textual descriptions of experimental
    conditions, observations, and operator notes.
\end{enumerate}

Three phases define each experiment. The \emph{Startup} phase covers equipment heat-up
and stabilization, while \emph{Operation} corresponds to active distillation with product
collection; finally, \emph{Shutdown} involves controlled cooling and depressurization.
To study anomaly detection, some experiments feature intentionally induced faults---reflux
interruptions, heating power reductions, or feed composition
perturbations, among others. Each anomalous experiment is paired with a fault-free reference
run conducted under identical nominal conditions, enabling direct comparison.

% =========================================================================
\section{Methodology}
\label{sec:methodology}
% =========================================================================

\begin{landscape}
\begin{figure*}[htbp]
  \centering
  \includegraphics[width=1.65\textwidth]{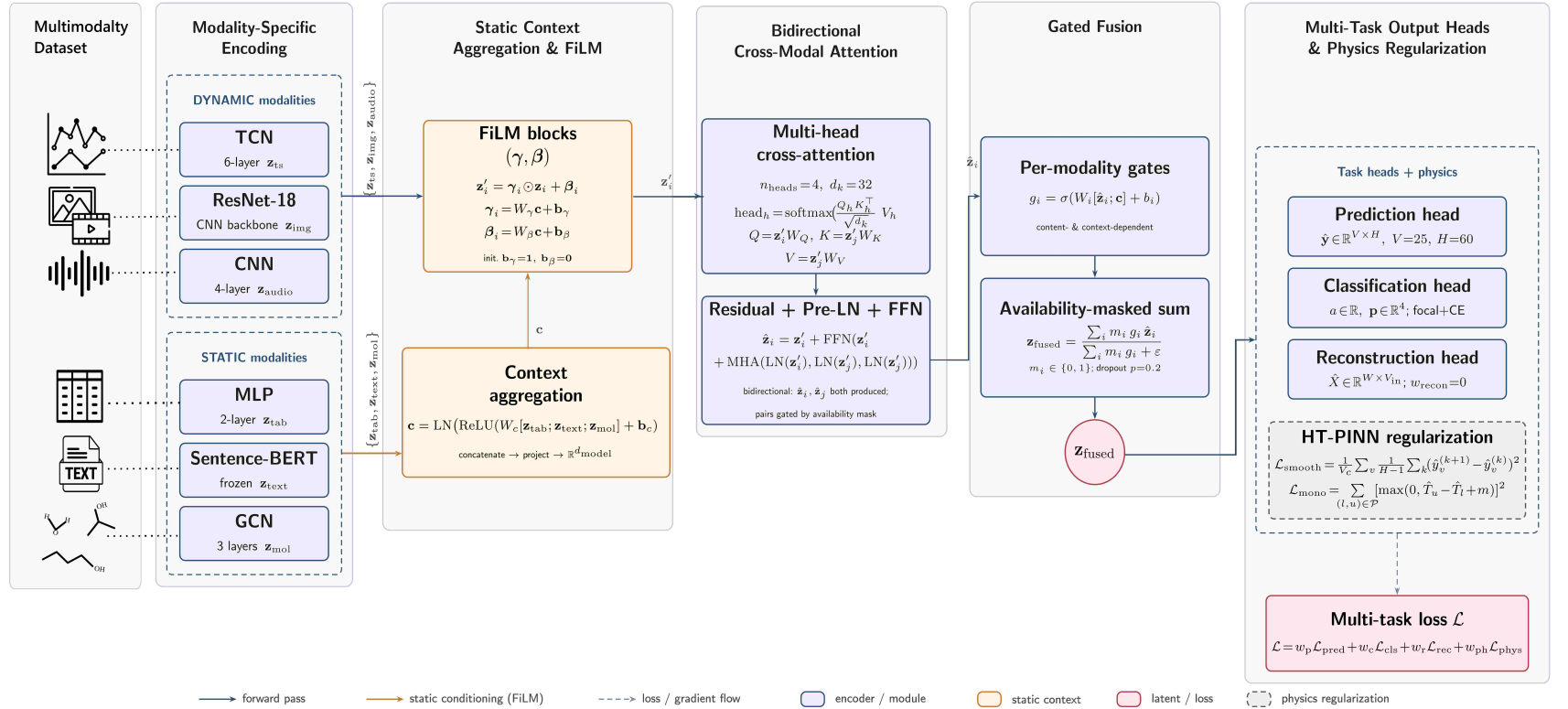}
  \caption{The UTOPYA architecture.
    Dynamic modalities are encoded by dedicated networks and conditioned
    via FiLM using a static context vector~$\mathbf{c}$ aggregated from
    tabular, text, and molecular graph encoders. Conditioned embeddings
    undergo bidirectional cross-modal attention and context-aware gated
    fusion to produce a latent vector~$\mathbf{z}$ that feeds three task
    heads. The multi-task loss
    $\mathcal{L} = \alpha\mathcal{L}_\text{pred} + \beta\mathcal{L}_\text{class}
    + \gamma\mathcal{L}_\text{recon} + \delta\mathcal{L}_\text{physics}$
    is augmented with physics-informed regularisation
    (Section~\ref{sec:physics}).}
  \label{fig:architecture}
\end{figure*}
\end{landscape}

The UTOPYA architecture consists of five stages: (1)~modality-specific
encoding, (2)~static context aggregation with FiLM conditioning,
(3)~bidirectional cross-modal attention, (4)~gated fusion with graceful
degradation, and (5)~multi-task output heads with physics-informed
regularization. The complete architecture is illustrated in
Figure~\ref{fig:architecture}. UTOPYA has 15.2M trainable parameters with a
shared embedding dimension of $d_\text{model} = 128$.

% ----- 5.1 Architecture Overview -----
\subsection{Architecture Overview}
\label{sec:arch_overview}

UTOPYA processes two categories of modalities: \emph{dynamic} modalities
that vary within an experiment (time-series, images, audio, video) and
\emph{static} modalities that are fixed per experiment (tabular features,
text metadata, molecular graphs). Static modalities are aggregated into a
context vector that conditions dynamic modality embeddings via FiLM. The
conditioned dynamic embeddings undergo bidirectional cross-modal attention
before being combined through gated fusion. Three output heads (prediction,
classification, and reconstruction) operate on the fused representation
under a multi-task loss augmented with physics-informed regularization.

% ----- 5.2 Modality Encoders -----
\subsection{Modality Encoders}
\label{sec:encoders}

Each modality demands its own encoder. Uniform encoding is challenging because each data type---time-series, images, audio, tabular records, free text, and molecular graphs---has a fundamentally different structure, and the encoder must respect the inductive biases that come with each format while still producing compatible representations for the downstream fusion stage. However, all encoders project to a shared $d_\text{model} = 128$. We describe each one below.

The primary modality is time-series process data. It comprises 29 variables: 5 manipulated inputs and 24 measured outputs that include temperatures, pressures, flow rates, liquid levels, and valve states, all sampled at 1~Hz over a window of $T=120$ timesteps. We select a Temporal Convolutional Network (TCN);
%\citep[TCN;][]
\citep{bai2018empirical} over recurrent alternatives. The reasons are threefold: dilated causal convolutions give an exponentially growing receptive field without the vanishing-gradient issues that plague RNNs; the entire window can be processed in parallel, which improves training throughput; and the inductive bias toward local temporal patterns suits dense process data sampled at regular intervals.

The TCN consists of $L=6$ residual blocks, each containing a causal 1-D convolution with kernel size $k=3$ and dilation factor $d_l = 2^{l-1}$ for layer $l \in \{1, \ldots, 6\}$. The receptive field is therefore $R = 1 + 2(k-1)\sum_{l=0}^{L-1} 2^l = 127$ timesteps, closely matching the input window. Formally, the dilated convolution at layer $l$ computes
\begin{equation}
  \mathbf{h}_l^{(t)} = \text{ReLU}\!\left(\sum_{j=0}^{k-1} \mathbf{W}_l^{(j)} \mathbf{h}_{l-1}^{(t - j \cdot d_l)} + \mathbf{b}_l\right),
  \label{eq:tcn_conv}
\end{equation}
where $\mathbf{W}_l^{(j)} \in \mathbb{R}^{d_\text{model} \times d_\text{model}}$ are the learned filters and the input to the first layer is a linear projection of the raw 29-dimensional signal. Each block applies weight normalisation \citep{salimans2016weight}, ReLU activation, and dropout ($p=0.5$), with a skip connection $\mathbf{h}_l = \mathbf{h}_l + \mathbf{h}_{l-1}$ (after dimension matching when needed).

The encoder produces both a pooled embedding $\mathbf{z}_\text{ts} = \frac{1}{T}\sum_{t=1}^{T} \mathbf{h}_L^{(t)} \in \mathbb{R}^{d_\text{model}}$ (via global average pooling) and a per-timestep feature map $\mathbf{Z}_\text{ts} \in \mathbb{R}^{T \times d_\text{model}}$ used by the prediction and reconstruction heads.

Three cameras capture the column, reboiler, and condenser regions, providing visual evidence of liquid levels, foaming, flooding, and colour changes that correlate with composition. We use a shared ResNet-18 backbone \citep{he2016deep} pretrained on ImageNet, which provides strong low-level feature extraction without requiring the large labelled image datasets that training from scratch would demand. Each camera frame $\mathbf{I}_c \in \mathbb{R}^{224 \times 224 \times 3}$ is processed independently through the convolutional trunk to produce a 512-dimensional feature vector, which is then projected to $d_\text{model}$ via a learned linear layer. The three camera embeddings are aggregated via mean pooling:
\begin{equation}
  \mathbf{z}_\text{img} = \frac{1}{3}\sum_{c=1}^{3} \mathbf{W}_\text{img}\, f_\text{ResNet}(\mathbf{I}_c) + \mathbf{b}_\text{img},
  \label{eq:img_enc}
\end{equation}
where $\mathbf{W}_\text{img} \in \mathbb{R}^{d_\text{model} \times 512}$. Mean pooling was chosen over concatenation to keep the embedding dimension fixed regardless of the number of active cameras, supporting graceful degradation when a camera is unavailable.

Acoustic recordings capture boiling intensity, vapour flow turbulence, and mechanical vibrations that change during anomalous operation (e.g., column flooding produces a distinctive low-frequency rumble). The raw waveform is converted to a log-mel spectrogram ($n_\text{mels}=64$, hop length 512 samples) and processed by a 4-layer convolutional neural network with batch normalisation and max-pooling, followed by global average pooling and a linear projection to produce $\mathbf{z}_\text{audio} \in \mathbb{R}^{d_\text{model}}$. This lightweight architecture is appropriate given the relatively low information density of the audio channel compared to the time-series data.

Experiment-level metadata (operating conditions, equipment configuration, chemical system identifiers) form a high-dimensional but static feature vector $\mathbf{x}_\text{tab} \in \mathbb{R}^{d_\text{tab}}$, where $d_\text{tab}$ varies between 602 and 606 depending on the chemical system. A two-layer MLP with ReLU activations and dropout processes this vector:
\begin{equation}
  \mathbf{z}_\text{tab} = \mathbf{W}_2\, \text{ReLU}(\mathbf{W}_1 \mathbf{x}_\text{tab} + \mathbf{b}_1) + \mathbf{b}_2,
  \label{eq:tab_enc}
\end{equation}
with $\mathbf{W}_1 \in \mathbb{R}^{256 \times d_\text{tab}}$ and $\mathbf{W}_2 \in \mathbb{R}^{d_\text{model} \times 256}$. Dimensionality reduction is the first role of this MLP. However, it also acts as a non-linear feature extractor, and this second role is arguably more important: the non-linearity lets the network discover interactions among operating parameters---for instance, how reflux ratio and heating power jointly affect column stability---that a simple linear projection would miss entirely, which matters when the downstream task depends on precisely those coupled effects.

Numerical sensors alone miss important semantic context. Operator notes such as ``reflux interrupted at minute 15'' or ``unusual colour observed'' carry information about the experiment that is difficult to extract from raw signals, and losing this information would leave the model blind to events that operators routinely record in their logs. We therefore encode these free-text fields into 384-dimensional embeddings with a pre-trained Sentence-BERT model \citep{reimers2019sentence} and project them to the shared space:
\begin{equation}
  \mathbf{z}_\text{text} = \mathbf{W}_\text{text}\, \text{SBERT}(\text{text}) + \mathbf{b}_\text{text},
  \label{eq:text_enc}
\end{equation}
where $\mathbf{W}_\text{text} \in \mathbb{R}^{d_\text{model} \times 384}$. Using a frozen pre-trained language model avoids fine-tuning on the very small text corpus available.

The chemical identity of the mixture components influences expected process behaviour (e.g., azeotrope formation, relative volatilities). Molecular structures are represented as SMILES\footnote{Simplified Molecular Input Line Entry System (SMILES) is a notation that encodes molecular structures as short ASCII strings, widely adopted in cheminformatics and machine learning on molecules.}
 strings, converted to graphs where atoms are nodes and bonds are edges, and processed by a Graph Convolutional Network \citep[GCN;][]{kipf2017semi} with $K=3$ message-passing layers \citep{gilmer2017neural}. At each layer $k$, node features are updated according to
\begin{equation}
  \mathbf{h}_v^{(k)} = \text{ReLU}\!\left(\mathbf{W}^{(k)} \sum_{u \in \mathcal{N}(v) \cup \{v\}} \frac{\mathbf{h}_u^{(k-1)}}{\sqrt{|\mathcal{N}(u)||\mathcal{N}(v)|}}\right),
  \label{eq:gcn}
\end{equation}
where $\mathcal{N}(v)$ denotes the neighbours of atom $v$. Global mean pooling over all nodes yields $\mathbf{z}_\text{mol} \in \mathbb{R}^{d_\text{model}}$, capturing a graph-level representation of the molecular identity.

% ----- 5.3 FiLM Conditioning -----
\subsection{FiLM Conditioning}
\label{sec:film}

The same sensor reading can have different implications depending on the experimental context. For instance, a column top temperature of 80\textdegree C is expected when distilling a water-rich ternary mixture but would be anomalous for an ethanol-propanol binary system. Static modalities (tabular features, text metadata, molecular graphs) encode this experiment-level context, and thus the dynamic modality representations must be adapted accordingly.

We adopt Feature-wise Linear Modulation \citep[FiLM;][]{perez2018film} over simpler alternatives such as concatenation or additive conditioning. FiLM applies a learned \emph{channel-wise} affine transformation, which lets the context selectively amplify, suppress, or shift individual feature dimensions without increasing the embedding size. This is more expressive than additive conditioning while remaining computationally lightweight.

We first aggregate the three static embeddings into a single context vector via concatenation and non-linear projection:
\begin{equation}
  \mathbf{c} = \text{LN}\!\left(\text{ReLU}\!\left(\mathbf{W}_c
  [\mathbf{z}_\text{tab}; \mathbf{z}_\text{text}; \mathbf{z}_\text{mol}]
  + \mathbf{b}_c\right)\right),
  \label{eq:context}
\end{equation}
where $\mathbf{c} \in \mathbb{R}^{d_\text{model}}$, $\mathbf{W}_c \in \mathbb{R}^{d_\text{model} \times 3d_\text{model}}$, and LN denotes layer normalisation \citep{ba2016layer}. Why concatenation and not mean pooling? Mean pooling 
would discard chemical system specific features before the projection has a chance to weight them, and in preliminary experiments this loss of information hurts performance on the smaller chemical systems. Thus, concatenation preserves the full information from all three static sources, and the non-linear projection compresses it into a single context vector while the layer normalisation stabilises its scale.

Each dynamic modality embedding $\mathbf{z}_i$ (for $i \in \{\text{ts}, \text{img}, \text{audio}\}$) is then modulated by context-dependent scale and shift parameters:
\begin{equation}
  \mathbf{z}'_i = \boldsymbol{\gamma}_i \odot \mathbf{z}_i
  + \boldsymbol{\beta}_i, \quad
  \text{where} \quad
  \boldsymbol{\gamma}_i = \mathbf{W}_\gamma \mathbf{c} + \mathbf{b}_\gamma,
  \quad
  \boldsymbol{\beta}_i = \mathbf{W}_\beta \mathbf{c} + \mathbf{b}_\beta,
  \label{eq:film}
\end{equation}
with $\mathbf{W}_\gamma, \mathbf{W}_\beta \in \mathbb{R}^{d_\text{model} \times d_\text{model}}$, and $\odot$ denoting element-wise multiplication. Feature gating is controlled by $\boldsymbol{\gamma}_i$: values near zero suppress a dimension, while values greater than one amplify it. The shift $\boldsymbol{\beta}_i$ introduces context-dependent bias. One detail matters for training stability. We initialise $\mathbf{b}_\gamma = \mathbf{1}$ and $\mathbf{b}_\beta = \mathbf{0}$ so that at the start FiLM performs an identity transform---this preserves the pretrained encoder representations until the conditioning parameters have had enough gradient updates to produce meaningful modulations, and without this initialisation the pretrained features degrade rapidly during the first few epochs of joint training.

% ----- 5.4 Cross-Modal Attention and Gated Fusion -----
\subsection{Cross-Modal Attention and Gated Fusion}
\label{sec:fusion}

The dynamic embeddings are now contextually informed---but still isolated. A temperature spike coinciding with a visual change in the column, for instance, would remain invisible to any single encoder acting alone, because such complementary patterns span modalities that have not yet exchanged information with each other. We address this with cross-modal attention.

The cross-modal attention operates pairwise. For each ordered pair $(i, j)$ of available modalities with $i \neq j$, modality $i$ attends to modality $j$ by supplying the queries while the attended modality contributes keys and values, following the standard scaled dot-product multi-head attention of \citet{vaswani2017attention}:
\begin{equation}
  \mathbf{Q} = \mathbf{z}'_i \mathbf{W}^Q, \quad
  \mathbf{K} = \mathbf{z}'_j \mathbf{W}^K, \quad
  \mathbf{V} = \mathbf{z}'_j \mathbf{W}^V,
  \label{eq:qkv}
\end{equation}
where $\mathbf{W}^Q, \mathbf{W}^K, \mathbf{W}^V \in \mathbb{R}^{d_\text{model} \times d_\text{model}}$. The attention output for head $h$ is
\begin{equation}
  \text{head}_h = \text{softmax}\!\left(\frac{\mathbf{Q}_h \mathbf{K}_h^\top}{\sqrt{d_k}}\right) \mathbf{V}_h,
  \label{eq:attention_head}
\end{equation}
with $d_k = d_\text{model} / n_\text{heads}$. Four attention heads are used, allowing the model to attend to different aspects 
of cross-modal interactions simultaneously; each head operates on a projected 
subspace of dimension $d_{\text{model}} / n_{\text{heads}} = 32$, so the full 
head stack preserves the 128-dimensional bottleneck width \citep{vaswani2017attention}. 
This design enables the model to jointly capture multiple distinct cross-modal 
interaction patterns---such as temporal co-variation, anomaly-indicative 
divergence, and phase-dependent modulation---within a single attention layer.
Their outputs are concatenated and projected through $\mathbf{W}^O \in \mathbb{R}^{d_\text{model} \times d_\text{model}}$ to produce $\text{MHA}(\mathbf{z}'_i, \mathbf{z}'_j)$---a single vector that summarises what modality $i$ learned from modality $j$ across all four attention heads simultaneously. A residual connection with pre-layer normalisation then yields the updated embedding:
\begin{equation}
  \hat{\mathbf{z}}_i = \mathbf{z}'_i + \text{FFN}\!\left(\mathbf{z}'_i + \text{MHA}(\text{LN}(\mathbf{z}'_i),\, \text{LN}(\mathbf{z}'_j),\, \text{LN}(\mathbf{z}'_j))\right),
  \label{eq:cross_attn}
\end{equation}
where FFN is a two-layer feed-forward network with ReLU activation and hidden dimension $4 d_\text{model}$, and LN denotes layer normalisation. The attention is \emph{bidirectional}---both $\hat{\mathbf{z}}_i$ and $\hat{\mathbf{z}}_j$ are computed. Only one layer of cross-attention is applied. Furthermore, it is restricted to modality pairs whose availability mask is active, since attending to uninformative default embeddings injected noise into the fused representation and hurt downstream classification.

The per-pair attention output $\text{MHA}(\mathbf{z}'_i, \mathbf{z}'_j)$ summarises what modality~$i$ learned from modality~$j$ across all four heads. Although the present work uses these vectors only as inputs to the gated fusion layer, they carry structural information that could support symbolic-level explainability: attention magnitudes per modality pair indicate which channels were informative for a given prediction, and the per-head decomposition could be inspected to identify reusable interaction patterns (e.g.\ ``audio attends to time-series during flooding events''). Extracting such symbolic rules from the trained attention maps---and combining them with the cluster-level diagnostics of Section~\ref{sec:umap}---is left as future work and discussed further in Section~\ref{sec:limitations}.

The attended embeddings must now be combined into one vector. Averaging would be the simplest option, but it ignores the fact that different modalities are informative to different degrees depending on the specific input sample---for a particular experiment the audio channel may carry more diagnostic value than the images, or vice versa. Thus, we use learned sigmoid gates:
\begin{equation}
  g_i = \sigma\!\left(\mathbf{W}_i [\hat{\mathbf{z}}_i; \mathbf{c}]
  + \mathbf{b}_i\right), \quad
  \mathbf{z}_\text{fused} =
  \frac{\sum_{i=1}^{N} m_i \cdot g_i \cdot \hat{\mathbf{z}}_i}
       {\sum_{i=1}^{N} m_i \cdot g_i + \epsilon},
  \label{eq:gated_fusion}
\end{equation}
where $\mathbf{W}_i \in \mathbb{R}^{1 \times 2d_\text{model}}$ projects the concatenation of the attended embedding and context vector to a scalar gate, $m_i \in \{0, 1\}$ is the availability mask, $\sigma$ is the sigmoid, and $\epsilon = 10^{-8}$ prevents division by zero. The gates are input-dependent. Because $g_i$ depends on both the modality content $\hat{\mathbf{z}}_i$ and the static context $\mathbf{c}$, the model can learn, for instance, to up-weight audio when the operator notes mention a noisy environment---a behaviour that would not be possible with fixed fusion weights. Furthermore, the mask $m_i$ zeros out any absent modality before the weighted sum so that the denominator renormalises automatically; this guarantees that the fused vector remains well-scaled at inference time regardless of how many modalities happen to be present for a given experiment.

In practice, not all modalities are available for every window (e.g.\ camera frames are captured only intermittently and audio recordings are not available on every experiment). To ensure robustness, we employ three complementary mechanisms: (i)~\emph{modality dropout} during training, where each non-essential modality is independently masked with probability $p = 0.2$ (the time-series modality, being the primary information source, is never dropped); (ii)~\emph{learned default embeddings} $\mathbf{d}_i \in \mathbb{R}^{d_\text{model}}$ that are substituted for any absent modality, initialised to zero and trained end-to-end; and (iii)~\emph{availability-masked renormalisation} in the gated fusion denominator (Eq.~\ref{eq:gated_fusion}), which ensures that the fused vector is properly scaled regardless of how many modalities are present.

% ----- 5.5 Output Heads and Multi-Task Loss -----
\subsection{Output Heads and Multi-Task Loss}
\label{sec:heads}

The fused representation $\mathbf{z}_\text{fused}$ feeds three output heads:

An MLP projects the fused embedding to per-variable predictions:
$\hat{\mathbf{y}} \in \mathbb{R}^{V \times H}$, where $V = 25$ target
variables and $H = 60$ future timesteps (1~minute horizon at 1~Hz). The
25~target variables are the continuous process measurements (temperatures,
pressures, flow rates, and liquid levels); the 4~binary valve states are
excluded from the prediction targets because they change discontinuously
and are better handled by the classification head. Each
variable has a dedicated output projection to accommodate heterogeneous
scales and dynamics.

A two-layer MLP with hidden dimension 128 and dropout produces two outputs:
(i)~a binary anomaly logit $a \in \mathbb{R}$, and (ii)~phase
classification logits $\mathbf{p} \in \mathbb{R}^{4}$ for four operational
phases (normal, blind, anomalous, recovery).

A decoder reconstructs the input time-series window
$\hat{\mathbf{X}} \in \mathbb{R}^{W \times V_\text{in}}$ from the fused
representation, where $W = 120$ is the window size and $V_\text{in} = 29$
is the number of input variables.

The four output heads are supervised jointly through a weighted sum of four task-specific losses:
\begin{equation}
  \mathcal{L}
  \;=\;
  w_\text{pred}\,\mathcal{L}_\text{pred}
  + w_\text{class}\,\mathcal{L}_\text{class}
  + w_\text{recon}\,\mathcal{L}_\text{recon}
  + w_\text{phys}\,\mathcal{L}_\text{phys},
  \label{eq:total_loss}
\end{equation}
with fixed weights $w_\text{pred} = 0.1$, $w_\text{class} = 2.0$,
$w_\text{recon} = 0.0$ (the reconstruction head is disabled in the
production model; we report the ablation that motivates this choice in
Section~\ref{sec:ablation}), and $w_\text{phys} = 0.5$. We adopt fixed
weights rather than learnable uncertainty weighting
\citep{kendall2018multi}: the ablation study reported in
Section~\ref{sec:ablation} compares both schemes and reveals that, in
our data-scarce setting, the prediction loss tends to dominate when
weights are learned, degrading anomaly detection performance. The four
component losses are defined below; the physics loss is deferred to
Section~\ref{sec:physics}.

\paragraph*{Prediction loss.}
The prediction head emits per-variable trajectories
$\hat{y}_v^{(k)}$ over the $H = 60$-step horizon. Continuous
variables (temperatures, pressures, flow rates, liquid levels;
$V_c = 21$) are supervised with mean-squared error against the ground
truth $y_v^{(k)}$, and binary valve states ($V_b = 4$) are supervised
with binary cross-entropy on the same horizon, so that prediction
errors on valve switching contribute on the same scale as continuous
residuals:
\begin{equation}
  \mathcal{L}_\text{pred}
  \;=\;
  \frac{1}{V_c H}\sum_{v=1}^{V_c}\sum_{k=1}^{H}
      \!\left(\hat{y}_v^{(k)} - y_v^{(k)}\right)^{2}
  \;+\;
  \frac{1}{V_b H}\sum_{v=1}^{V_b}\sum_{k=1}^{H}
      \mathrm{BCE}\!\left(\hat{y}_v^{(k)},\, y_v^{(k)}\right).
  \label{eq:pred_loss}
\end{equation}
The total of $V_c + V_b = 25$ target variables corresponds to the
prediction-MAE column of Table~\ref{tab:multimodal_ablation}.

\paragraph*{Classification loss.}
The classification head produces (i) a binary anomaly logit $a$ that is
trained against the binary anomaly label $y \in \{0, 1\}$ and (ii)
phase logits $\mathbf{p}\in\mathbb{R}^{4}$ trained against the
phase label $c \in \{1, \dots, 4\}$. The total classification loss is
the sum of an anomaly-focal term and a phase cross-entropy term:
\begin{equation}
  \mathcal{L}_\text{class}
  \;=\;
  \mathcal{L}_\text{focal}
  + \mathcal{L}_\text{phase},
  \label{eq:class_loss}
\end{equation}
\begin{equation}
  \mathcal{L}_\text{focal}
  \;=\;
  -\,w_y\,(1 - p_t)^{\gamma_f}\,\log p_t,
  \quad
  p_t \;=\;
  \begin{cases}
    \sigma(a)     & \text{if } y = 1,\\
    1 - \sigma(a) & \text{if } y = 0,
  \end{cases}
  \label{eq:focal}
\end{equation}
\begin{equation}
  \mathcal{L}_\text{phase}
  \;=\;
  -\sum_{j=1}^{4} \mathbbm{1}\{c = j\}\,\log
      \frac{\exp(p_j)}{\sum_{i=1}^{4}\exp(p_i)},
  \label{eq:phase_ce}
\end{equation}
where $\sigma(\cdot)$ is the sigmoid, $\gamma_f = 2.0$ is the focal
focusing parameter \citep{lin2017focal}, and the per-sample weight is
$w_y = w_{+} = 6.0$ for anomalous windows and $w_y = 1.0$ for normal
windows. The value $w_{+} = 6.0$ compensates the $\sim 14\%$ anomaly
prevalence; the corresponding theoretical balance weight is
$0.86 / 0.14 \approx 6.1$. The two classification terms are summed
without further reweighting because the focal modulator and the
phase-CE log-likelihood already operate on comparable scales after the
class weighting.

\paragraph*{Reconstruction loss.}
A U-Net-style decoder reconstructs the input window
$\hat{\mathbf{X}}\in\mathbb{R}^{W\times V_\text{in}}$ from the fused
representation, with $W = 120$ and $V_\text{in} = 29$. The reconstruction
loss is the per-element mean squared error against the input:
\begin{equation}
  \mathcal{L}_\text{recon}
  \;=\;
  \frac{1}{W \cdot V_\text{in}}
  \sum_{t=1}^{W}\sum_{v=1}^{V_\text{in}}
      \!\left(\hat{X}_{t,v} - X_{t,v}\right)^{2}.
  \label{eq:recon_loss}
\end{equation}
This loss is included in Equation~\eqref{eq:total_loss} for completeness
but is gated by $w_\text{recon} = 0$ in the production model: the
ablation study in Section~\ref{sec:ablation} shows that the
reconstruction objective competes with the classification objective
under the data-scarce regime, and disabling it improves both validation
and test AUROC.

% ----- 5.6 Physics-Informed Regularization -----
\subsection{Physics-Informed Regularization}
\label{sec:physics}

Purely data-driven models can learn to make predictions that fit the training data well but violate basic physical principles. This is especially the case when the training set is small, where physically implausible predictions can emerge. 
To mitigate this, we introduce two physics-informed regularisation terms that encode domain knowledge about batch distillation dynamics directly into the loss function. The purpose is twofold: (i) to constrain the model's hypothesis space to physically plausible predictions, reducing overfitting; and (ii) to provide a learning signal that is informative even for unlabelled normal windows, since physics violations indicate modelling errors regardless of anomaly labels.

Real physical processes are governed by conservation equations (mass, energy, momentum) that impose continuity on state variables. In a batch distillation column, thermal inertia of the liquid holdup and hydraulic resistance of the packing ensure that temperatures, pressures, and flow rates evolve smoothly over time. A predicted trajectory that exhibits unphysical jumps between consecutive timesteps is therefore a sign of overfitting to noise in the training data. We encode this prior by penalising the first-order finite differences of predicted trajectories:
\begin{equation}
  \mathcal{L}_\text{smooth} = \frac{1}{V_c} \sum_{v=1}^{V_c}
  \frac{1}{H-1} \sum_{k=1}^{H-1}
  \left(\hat{y}_v^{(k+1)} - \hat{y}_v^{(k)}\right)^2,
  \label{eq:smooth}
\end{equation}
where $V_c$ is the number of continuous (non-binary) target variables, $H=60$ is the prediction horizon, and $\hat{y}_v^{(k)}$ is the predicted value of variable $v$ at future timestep $k$. Binary valve states (which can change discontinuously) are excluded. This penalty is equivalent to minimising the squared $\ell_2$ norm of the discrete gradient of the predicted trajectory applied to multi-variable process predictions.

Distillation columns operate on the principle of vapour-liquid equilibrium (VLE): the more volatile components concentrate toward the top of the column while heavier components remain at the bottom. A direct thermodynamic consequence is that temperatures must decrease monotonically from the reboiler (bottom) to the condenser (top) during normal operation. Any predicted temperature inversion, where an upper section of the column is hotter than a lower section, violates this fundamental thermodynamic constraint and indicates that the model has learned an unphysical representation. We enforce this constraint through a squared hinge loss over four pairs of temperature sensors ordered along the column height:
\begin{equation}
  \mathcal{L}_\text{mono} = \sum_{(l, u) \in \mathcal{P}}
  \text{mean}\!\left[\max\!\left(0,\;
  \hat{T}_u - \hat{T}_l + m\right)^2\right],
  \label{eq:mono}
\end{equation}
where the sensor pairs $$\mathcal{P} = \{(\text{T703}, \text{T709}), (\text{T709}, \text{T711}), (\text{T711}, \text{T712}), (\text{T712}, \text{T705})\}$$ trace the column from reboiler liquid to column top, $\hat{T}_l$ and $\hat{T}_u$ are predicted temperatures at the lower and upper positions respectively, and $m \geq 0$ is a soft margin (set to 0 in our experiments). The penalty activates only when a temperature inversion is predicted ($\hat{T}_u > \hat{T}_l$) and is zero otherwise, making it a one-sided constraint that does not penalise physically consistent predictions.

The combined physics loss is:
\begin{equation}
  \mathcal{L}_\text{physics} = \lambda_\text{smooth}
  \mathcal{L}_\text{smooth}
  + \lambda_\text{mono} \mathcal{L}_\text{mono},
  \label{eq:physics_combined}
\end{equation}
with $\lambda_\text{smooth} = 1.0$ and $\lambda_\text{mono} = 0.5$.

The monotonicity constraint is strictly valid only under steady-state or near-equilibrium conditions. During startup transients, before the column reaches hydraulic equilibrium, temporary temperature inversions can occur and are physically legitimate. Similarly, systems exhibiting azeotropic behaviour may display non-monotonic temperature profiles under certain compositions. In the present work, the constraint is applied as a soft penalty ($\lambda_\text{mono} = 0.5$) rather than a hard constraint, which allows the model to tolerate transient violations without catastrophic loss increases. Future work could explore phase-dependent weighting that relaxes the monotonicity penalty during startup windows.

% =========================================================================
\section{Experimental Setup}
\label{sec:experimental_setup}
% =========================================================================

\subsection{Data Preprocessing and Splits}
\label{sec:preprocessing}

Time-series data is segmented into sliding windows of $W = 120$ timesteps
(2~minutes at 1~Hz) with stride $s = 30$ (30~seconds). Each window is
paired with a prediction target of $H = 60$ future timesteps and labeled
as anomalous or normal based on the experiment-level annotation. Window-level
phase labels are assigned by majority vote over the timestep-level phase
annotations within each window. All three experiment phases (Startup,
Operation, Shutdown) are concatenated for each experiment to preserve
temporal context.

We focus on the largest chemical system (ternary butan-1-ol + propan-2-ol
+ water, 91 experiments) and construct a leak-free split by ensuring that
no operating point appears in more than one partition. Operating points
define the nominal equipment settings (reflux ratio, heating power, feed
composition); experiments at the same operating point share identical
process configurations, so allowing them across partitions would let the
model memorise operating-point signatures rather than learning genuine
anomaly patterns. The split was selected via random search over 5\,000
seeds, optimising for zero operating-point overlap, balanced
normal/anomalous ratios, and sufficient normal experiments in all
partitions: 55 train (14 normal, 41 anomalous; 19 operating points), 16
validation (6 normal, 10 anomalous; 8 operating points), and 20 test (8
normal, 12 anomalous; 8 operating points), with no operating point shared
between any two partitions.

Each experiment is normalised relative to its own first 300 timesteps
(approximately 5 minutes of startup baseline). This per-experiment
normalisation removes specific operating point  baselines---mean
temperatures, pressures, and flow rates that differ between operating
points---so that the model focuses on \emph{deviations from each
experiment's own baseline} rather than on absolute sensor values.
Binary variables (valve states) are excluded from this normalisation.
Without per-experiment normalisation, the model conflates unfamiliar
operating-point signatures with anomalies, resulting in a test AUROC of
0.692; with it, the test AUROC rises to 0.832 ($+0.140$).

Static tabular features have variable dimensionality (602--606 features
depending on the chemical system); the model pads or truncates at runtime
to a fixed input dimension. The class distribution at the window level is
approximately 15\% anomalous and 85\% normal.

\subsection{Training Procedure}
\label{sec:training}

Training UTOPYA's 15.2M parameters on 55 experiments ($\sim$20\,000 windows) requires careful regularisation at every level to prevent overfitting. We describe each training decision and the rationale behind it.

We use AdamW \citep{loshchilov2019decoupled} with learning rate $\eta = 3 \times 10^{-4}$, weight decay $\lambda = 10^{-3}$, and default momentum parameters $(\beta_1, \beta_2) = (0.9, 0.999)$. Decoupled weight decay delivers a consistent regularisation effect regardless of gradient magnitude, which matters when different parts of the model (encoders, fusion layers, output heads) operate at different scales. The learning rate follows a cosine annealing schedule \citep{loshchilov2017sgdr} with $T_\text{max} = 100$ epochs and $\eta_\text{min} = 10^{-6}$, providing a smooth monotonic decay without restarts. Cosine annealing with warm restarts ($T_0 = 10$, $T_\text{mult} = 2$) was tested initially but found to cause training collapse at the restart points, where the sudden learning rate increase would destroy learned representations. A linear warmup over 3 epochs ramps the rate from $10^{-7}$ to $3 \times 10^{-4}$ to stabilise the initial training phase when randomly initialised heads produce noisy gradients.

Gradient norms are clipped to 1.0 \citep{pascanu2013difficulty} to prevent the exploding gradient problem that can arise in deep TCN architectures with large dilation factors. We use gradient accumulation over 4 steps with a per-step batch size of 16, yielding an effective batch size of 64. This provides a more stable gradient estimate than a batch of 16 alone while fitting within GPU memory constraints.

Dropout is set to $p = 0.5$ throughout the model, an aggressive rate justified by the high ratio of parameters to training examples ($15.2\text{M} / 20\,000 \approx 760$ parameters per training window). This forces the network to learn distributed, redundant representations rather than memorising individual training windows.

The TCN encoder is pretrained in a self-supervised manner on the full (unlabelled) time-series data from all chemical systems. The pretraining task combines two complementary objectives: (i)~\emph{block-masked reconstruction}, which randomly zeroes contiguous segments of 10--30 timesteps and trains the encoder to reconstruct them from surrounding context, forcing it to learn temporal dependencies beyond simple interpolation; and (ii)~a \emph{contrastive loss} that encourages the encoder to produce similar representations for augmented views of the same window while separating representations of different windows \citep{yue2022ts2vec}. This dual-objective pretraining provides richer initialisation than masked reconstruction alone, capturing both local temporal structure and global discriminative features.

During the first 3 epochs of supervised training, the pretrained encoder weights are frozen while the output heads and fusion layers train. This prevents the randomly initialised heads from corrupting the pretrained representations through large, noisy gradients. After 3 epochs, the encoder is unfrozen with a reduced learning rate of $3 \times 10^{-5}$ (10$\times$ smaller than the heads) to allow fine-tuning without catastrophic forgetting.

We introduce a curriculum learning strategy \citep{bengio2009curriculum} that exploits the natural difficulty hierarchy of anomaly detection in batch distillation. Each training window is assigned a difficulty score based on its anomaly phase: clearly normal windows receive a score of 0.0, clearly anomalous windows (where the fault is directly observable) receive 0.3, mixed-phase windows (containing transitions between normal and anomalous) receive 0.5, recovery-phase windows receive 0.6, and blind-phase windows (where the process upset has occurred but effects have not yet propagated to observable variables) receive the highest difficulty of 0.9. Training proceeds in three stages: (i)~epochs 1--5 use only the easiest 60\% of windows, allowing the model to learn clear anomaly patterns without confusion from ambiguous cases; (ii)~epochs 6--10 expand to 80\% of windows, introducing recovery-phase examples; and (iii)~from epoch~11 onward, the full dataset is used, including the challenging blind-phase windows. This schedule is implemented via a custom sampler that selects windows by difficulty threshold at each epoch.

With only $\sim$40 training experiments, augmentation is essential to expand the effective training distribution. We apply three complementary strategies \citep{wen2021time}: (i)~\emph{jitter} (additive Gaussian noise, $\sigma \in [0.01, 0.05]$, $p = 0.5$) simulates sensor noise and measurement uncertainty; (ii)~\emph{scaling} (multiplicative factor from $[0.9, 1.1]$, $p = 0.5$) mimics small calibration differences between experiments; and (iii)~\emph{time warping} (smooth nonlinear time distortion with 4 knots and maximum warp 0.1, $p = 0.3$) accounts for the natural variability in process dynamics across experiments, where the same operation may proceed slightly faster or slower.

Training runs for a maximum of 100 epochs with early stopping on validation AUROC (patience of 20 epochs). We discuss in Section~\ref{sec:seed_sensitivity} how this standard practice is complicated by an inverse val--test correlation in our setting.

The multi-task loss converges within the first 15 epochs, with the
prediction loss dominating early training before the classification head
refines its representations.

\paragraph*{Hardware and runtime.}
All training and evaluation runs reported in this paper were carried out
on a single laptop equipped with an NVIDIA GeForce RTX~5090 Laptop GPU,
using CUDA acceleration with mixed-precision (FP16) where supported. The
full multimodal model has 15.2~M trainable parameters and fits in GPU
memory with the per-step batch size of 16 (effective batch 64 after
gradient accumulation). Each ablation run trains within roughly
60--70~minutes wall-clock time (typical: 3{,}900--4{,}000~s for 30--40
epochs before early stopping), and the full eleven-configuration
multimodal ablation matrix completes in under 12~hours. Inference on the
6{,}848-window test set takes approximately 25~s. We do not interpret
training time as a headline result; the figure is provided here only so
that readers can reproduce the cost of the experimental matrix on
comparable consumer-grade hardware.

\subsection{Evaluation Metrics}
\label{sec:metrics}

The primary metric is the Area Under the Receiver Operating Characteristic
curve (AUROC), which measures the model's ability to discriminate between
anomalous and normal windows across all classification thresholds. We also
report the Area Under the Precision--Recall curve (AUPRC) due to the class
imbalance, per-variable Mean Absolute Error (MAE) for the prediction head,
and phase classification accuracy.

% =========================================================================
\section{Results and Discussion}
\label{sec:results}
% =========================================================================

\subsection{Main Results}
\label{sec:main_results}

Table~\ref{tab:main_results} reports the anomaly detection performance of
UTOPYA under the best configuration on the leak-free single-system split
with per-experiment normalisation. The best window-level test AUROC of
0.832 is obtained with the full multimodal configuration (time-series,
GC, audio, and static context) combined with self-supervised pretraining,
curriculum learning, and physics-informed regularisation. Under
multi-signal experiment-level scoring---where classification probabilities
and prediction errors are combined via rank-based fusion---the model
reaches 0.874 test AUROC, a $+0.166$ improvement over the unimodal
baseline. The val--test gap is small ($-0.008$), indicating that
per-experiment normalisation and the leak-free split substantially reduce
the overfitting observed in earlier configurations.

\begin{table}[htbp]
  \centering
  \caption{UTOPYA anomaly detection performance on the ternary
    butan-1-ol + propan-2-ol + water system using the leak-free split
    with per-experiment normalisation. The configuration is the full
    multimodal model (A7 in Table~\ref{tab:multimodal_ablation}). Window
    AUROC is computed per sliding window; experiment-level metrics
    aggregate window scores per experiment; multi-signal fuses
    classification and prediction error via rank-based combination.
    The test set contains 6848 windows with a 15.2\% anomaly rate,
    which bounds the achievable AUPRC and F1.}
  \label{tab:main_results}
  \begin{tabular}{@{}lcccc@{}}
    \toprule
    Metric & Val & Test & $\Delta$ \\
    \midrule
    Window AUROC         & 0.824 & \textbf{0.832} & $-$0.008 \\
    Experiment AUROC     & ---   & 0.781          & --- \\
    Multi-signal (2-sig) & ---   & \textbf{0.874} & --- \\
    AUPRC                & ---   & 0.474          & --- \\
    F1 (optimal thresh.) & ---   & 0.492          & --- \\
    \bottomrule
  \end{tabular}
\end{table}

While the AUROC of 0.832 indicates good discriminative ability across all thresholds, the AUPRC of 0.474 and the F1 score of 0.492 at the optimal threshold reveal that the model faces significant challenges in practical deployment. These metrics reflect the difficulty of the task: with approximately 15\% anomalous windows in the test set, the class imbalance limits precision at any reasonable recall level. An F1 of 0.492 means that, at the best operating point, the model misclassifies roughly half of its positive predictions or misses half of the true anomalies. For industrial deployment, this suggests that UTOPYA is better suited as a decision-support tool that flags suspicious intervals for operator review rather than as a fully autonomous alarm system. The experiment-level multi-signal AUROC of 0.874 is more encouraging, as operators typically evaluate anomalies at the experiment level rather than per individual window.

\begin{figure}[htbp]
  \centering
  \begin{subfigure}[b]{0.48\textwidth}
    \centering
    \includegraphics[width=\textwidth]{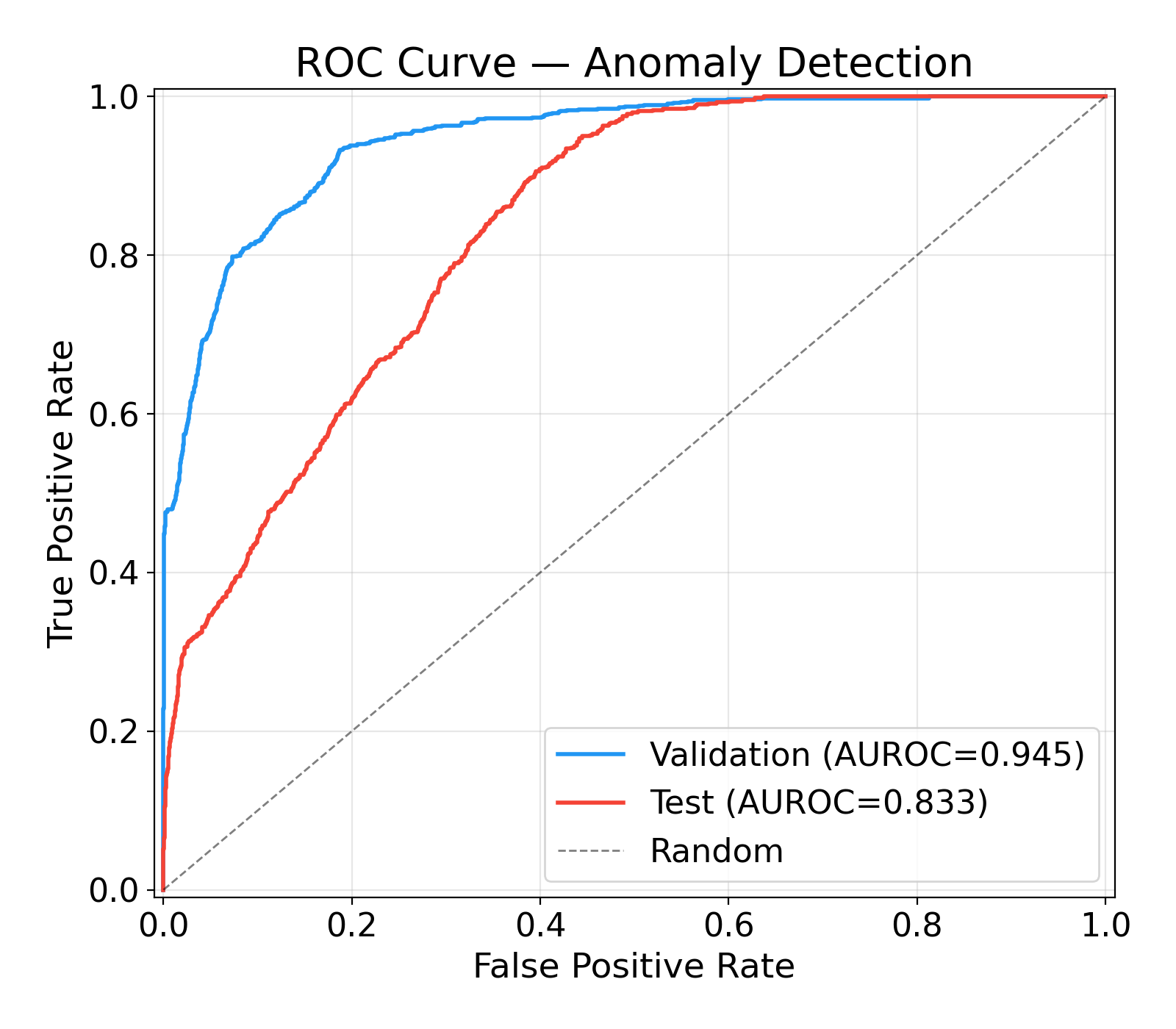}
    \caption{ROC curve.}
    \label{fig:roc_curve}
  \end{subfigure}
  \hfill
  \begin{subfigure}[b]{0.48\textwidth}
    \centering
    \includegraphics[width=\textwidth]{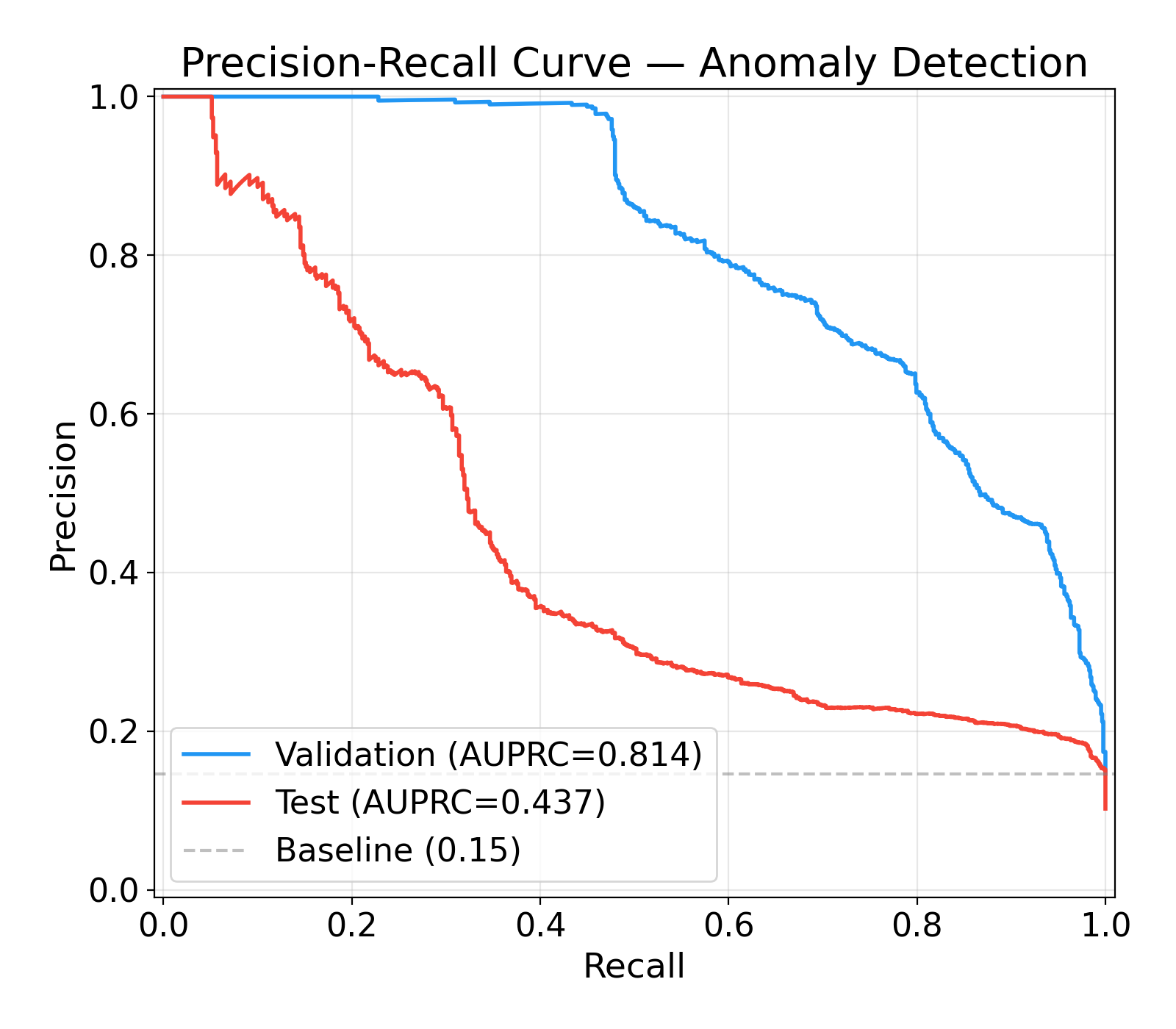}
    \caption{Precision--recall curve.}
    \label{fig:pr_curve}
  \end{subfigure}
  \caption{Classification performance curves for the best model
    (full multimodal, test AUROC 0.832). Both validation
    and test set curves are shown.}
  \label{fig:classification_curves}
\end{figure}

\begin{figure}[htbp]
  \centering
  \includegraphics[width=0.6\textwidth]{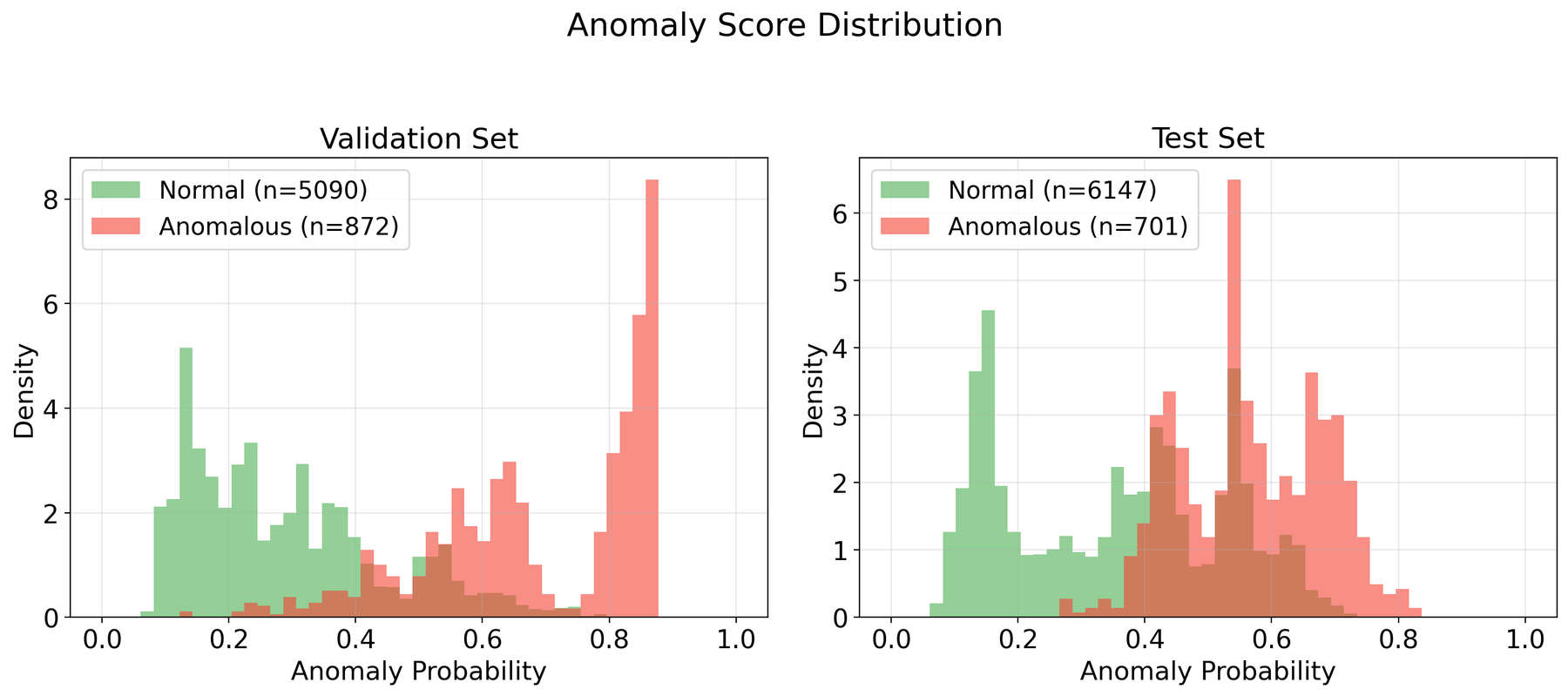}
  \caption{Distribution of anomaly scores for normal (blue) and anomalous
    (orange) windows on the test set. The overlap region corresponds to
    the difficulty range for the classifier.}
  \label{fig:score_dist}
\end{figure}

Figure~\ref{fig:roc_curve} shows the ROC curves for the best model
(curriculum, seed~42), and Figure~\ref{fig:pr_curve} shows the corresponding
precision--recall curve. Figure~\ref{fig:score_dist} visualises the anomaly
score distributions for normal and anomalous windows, revealing the overlap
region that defines the classifier's difficulty.

For the seed-42 curriculum configuration, the validation--test gap is small in both directions. Window-level AUROC is essentially tied (val~0.824 vs.\ test~0.832, $\Delta = -0.008$), and the ROC curves for the two splits in Figure~\ref{fig:roc_curve} sit on top of each other within sampling noise. The mild reversal of the gap (test slightly above validation) is a property of the seed sweep across the curriculum configuration, where all three seeds exhibit negative gaps; it is reported and discussed in Section~\ref{sec:seed_sensitivity} rather than visible in any single ROC curve.

\begin{landscape}
\begin{figure}[p]
  \centering
  \includegraphics[height=0.82\textheight]{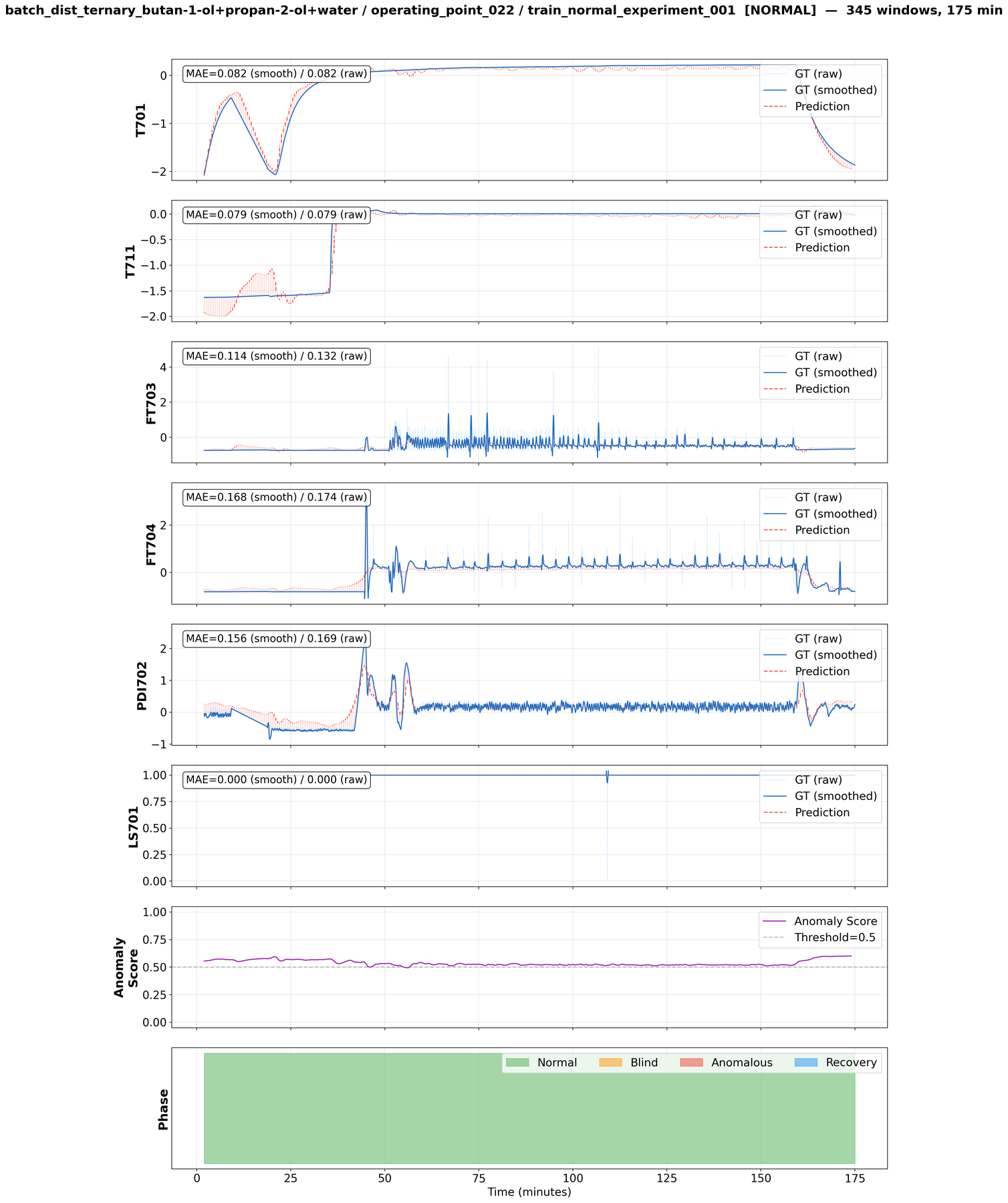}
  \caption{Per-variable prediction timeline for a normal experiment
    (OP-022, 175~min). Each panel shows one variable with the raw ground
    truth (light blue) and its Savitzky--Golay smoothed version (dark
    blue); the model prediction (dashed red) tracks the smoothed signal
    closely. Smoothed MAE: T701=0.08, T711=0.08, FT703=0.11, FT704=0.17,
    PDI702=0.16, LS701=0.00. The window-level anomaly score stays below
    the threshold throughout the run; the bottom band shows ground-truth
    phase labels (green = normal).}
  \label{fig:pred_normal}
\end{figure}
\end{landscape}

\begin{landscape}
\begin{figure}[p]
  \centering
  \includegraphics[height=0.82\textheight]{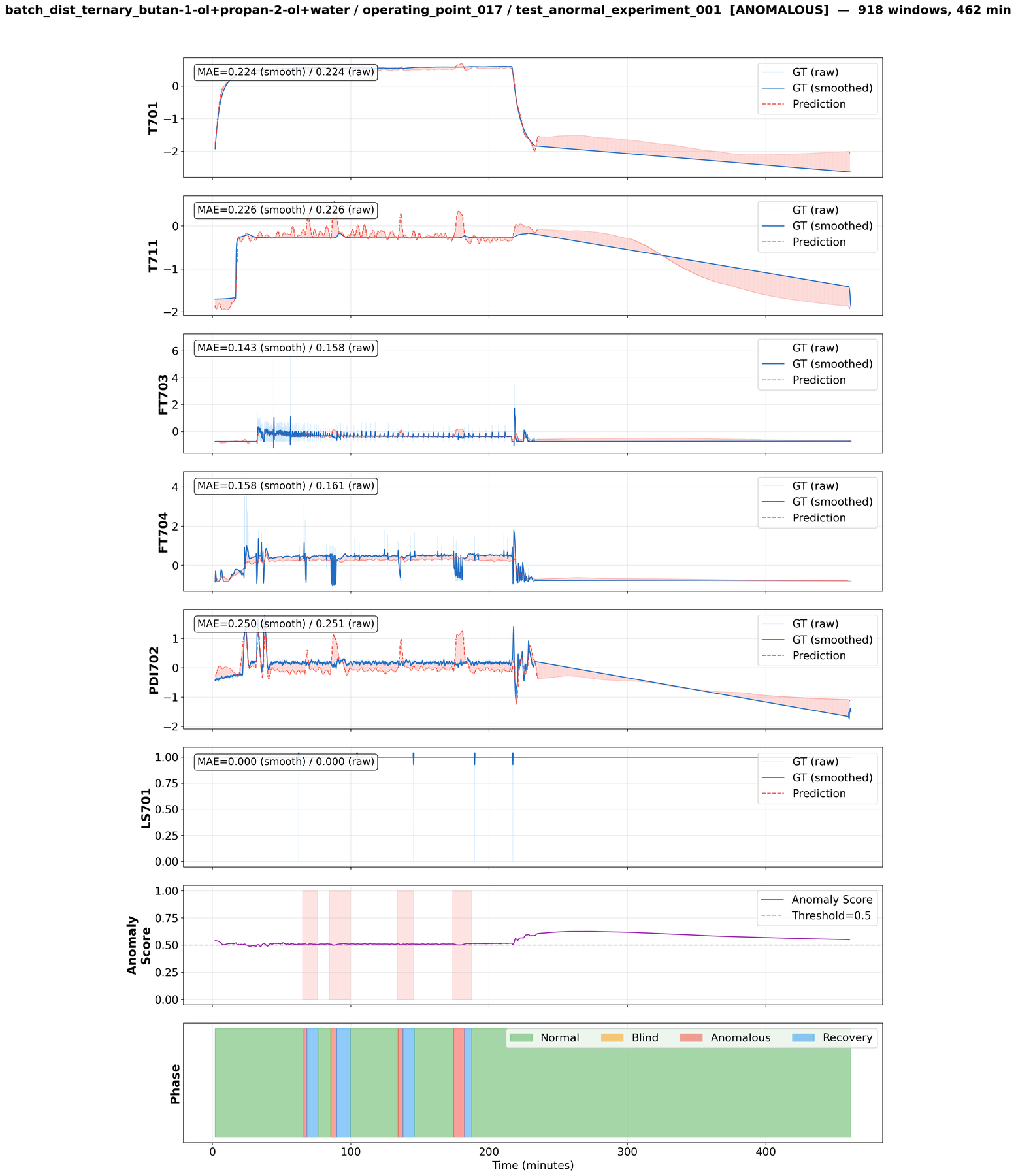}
  \caption{Per-variable prediction timeline for an anomalous experiment
    (OP-017, 462~min). Four fault intervals (shaded pink in the anomaly
    score panel) interrupt an otherwise normal run. Prediction error
    rises visibly during faults while the model continues to track the
    nominal dynamics during the recovery phases. Smoothed MAE:
    T701=0.22, T711=0.23, FT703=0.14, FT704=0.16, PDI702=0.25,
    LS701=0.00.}
  \label{fig:pred_anomalous}
\end{figure}
\end{landscape}

Figures~\ref{fig:pred_normal} and~\ref{fig:pred_anomalous} illustrate
representative prediction outputs from the model, showing the predicted
sensor trajectories alongside ground truth for a normal and an anomalous
experiment. For each of six
representative variables, the raw sensor trace is plotted faintly together
with a Savitzky--Golay smoothed version, which separates the physical
signal from sensor noise. The model's prediction (dashed red) is then
overlaid, and two MAE values are reported: against the smoothed ground
truth (which isolates prediction quality from sensor noise) and against the
raw trace. In the normal experiment (Figure~\ref{fig:pred_normal}),
predicted trajectories closely track the ground truth across all six
variables throughout the experiment duration, with smoothed MAEs between
0.08 and 0.17, and the anomaly score remains below the threshold. In the
anomalous experiment (Figure~\ref{fig:pred_anomalous}), the model tracks
the nominal dynamics tightly during normal and recovery phases, but
prediction errors rise visibly during fault intervals while the anomaly
score spikes concurrently---confirming that the classification and
prediction heads respond to the same underlying process disturbance through
different mechanisms.

\subsubsection{Prediction Head as Anomaly Signal}
\label{sec:pred_anomaly}

The prediction head serves a dual purpose in the UTOPYA architecture. Its
primary role is to regularise the encoder by requiring the learned
representation to contain sufficient information about the future evolution
of 25 process variables over a 60-second horizon. This auxiliary objective
forces the TCN encoder to capture the underlying process dynamics rather
than relying solely on classification-discriminative features, and the
physics-informed smoothness and monotonicity constraints
(Section~\ref{sec:physics}) further ensure that the predictions respect
thermodynamic principles.

However, the prediction head also provides an independent anomaly signal.
During normal operation, the model has learned the expected dynamics of the
process and is able to predict the sensor trajectories with low error.
When an anomaly occurs---a reflux interruption, a heating power
reduction, or a feed composition perturbation---the actual sensor
trajectories deviate from the patterns the model has learned, producing
elevated prediction errors even before the classification head explicitly
flags the window as anomalous. The prediction error thus functions as an
unsupervised anomaly indicator that complements the supervised
classification output.

\begin{figure}[htbp]
  \centering
  \includegraphics[width=\textwidth]{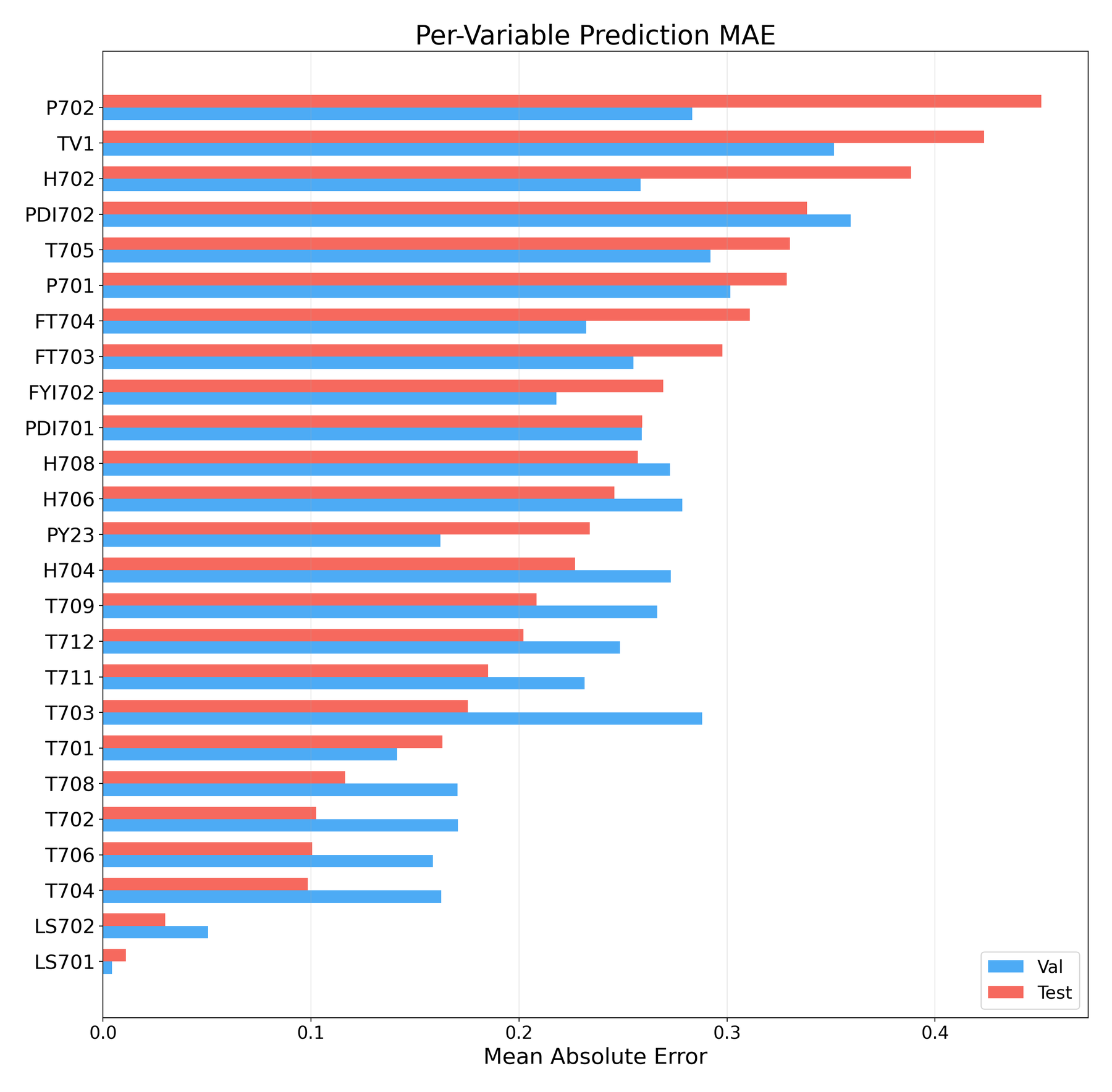}
  \caption{Per-variable Mean Absolute Error of the prediction head on the
    test set. Temperature sensors along the column show the lowest errors
    due to thermal inertia, while binary actuator states exhibit higher
    relative errors due to their discrete switching dynamics.}
  \label{fig:per_variable_mae}
\end{figure}

Figure~\ref{fig:per_variable_mae} shows the per-variable MAE on the test
set. Temperature variables along the column (T703, T705, T709, T711, T712)
exhibit the lowest prediction errors, consistent with their smooth,
physically-constrained dynamics---these variables are governed by energy
balances with large thermal inertia, which makes their short-term evolution
highly predictable. Flow rates (FT703, FT704) and differential pressure
measurements \linebreak (PDI701, PDI702) show intermediate errors, reflecting their
faster response to process disturbances. Binary valve states (LS701, LS702,
P301) show the highest relative errors, which is expected given their
discrete nature: predicting the exact timing of a valve transition is
inherently more difficult than predicting a continuous variable.

\subsubsection{Detailed Per-Experiment Prediction Timelines}
\label{sec:detailed_predictions}

Instead of isolating the best- and worst-predicted variables as
independent forecast snippets, we present three per-experiment timelines
that place the prediction in context against the concurrent anomaly score
and the phase ribbon. Each figure shows the model's one-step-ahead
forecast (dashed red) against both the raw sensor signal (light blue) and
its Savitzky--Golay smoothed version (dark blue) for six representative
variables, together with the window-level anomaly score and the
phase-classification ground truth. Two MAE values are reported per
variable: against the smoothed target (which isolates prediction skill
from sensor noise) and against the raw trace. The three figures were
chosen to illustrate, in order, (i)~a normal experiment with a benign
mid-run disturbance, (ii)~a short anomalous experiment with a single
localised fault, and (iii)~a longer anomalous experiment containing two
separated fault intervals.

\begin{landscape}
\begin{figure}[p]
  \centering
  \includegraphics[height=0.82\textheight]{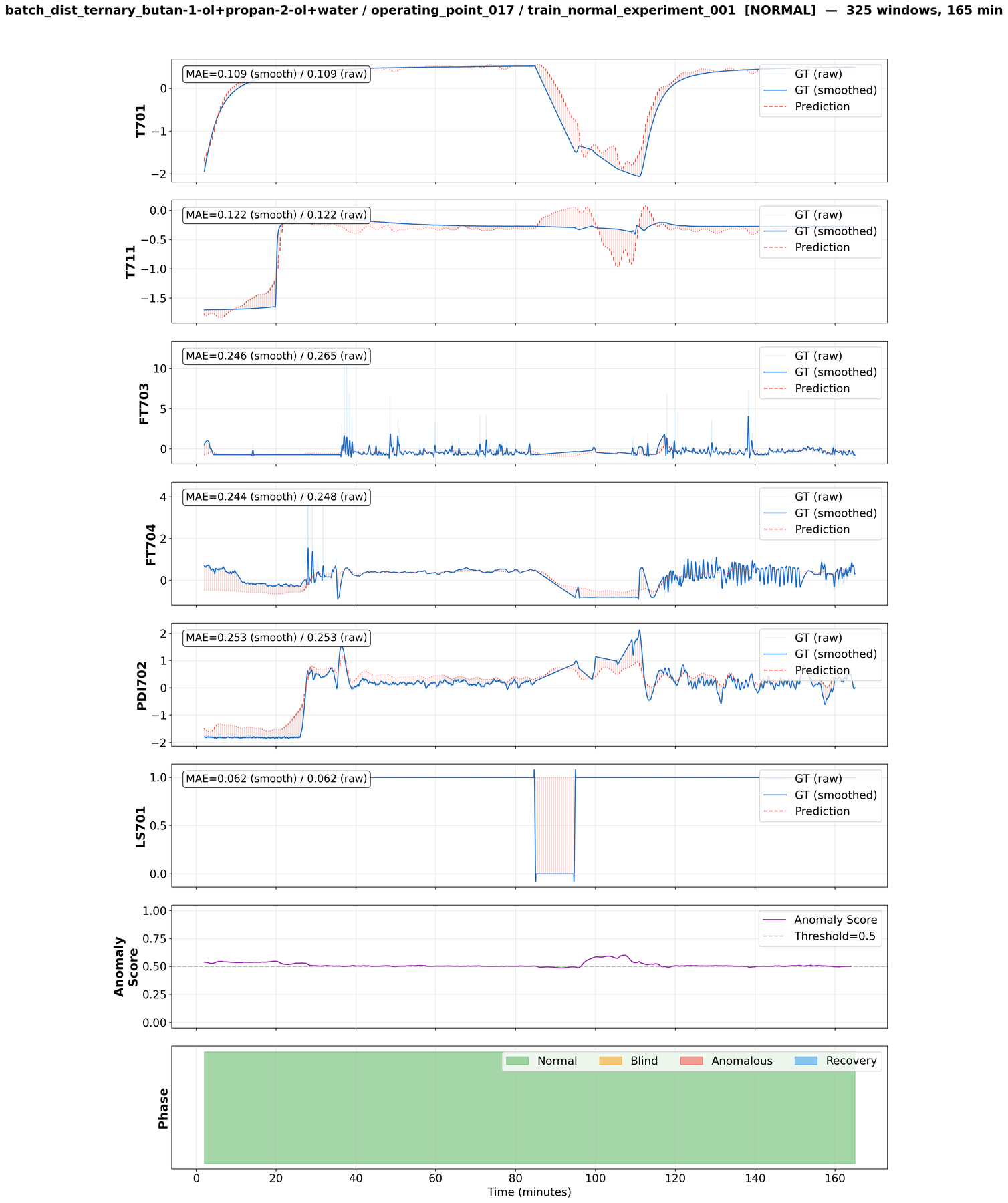}
  \caption{Detailed prediction timeline for a normal experiment
    (OP-017, train set). Despite a mid-run disturbance that causes a
    brief drop in T701 and T711 near $t\!\approx\!100$~min, the model
    tracks both the transient and the recovery. Smoothed MAEs are below
    0.26 for all continuous variables, and the anomaly score remains
    below the threshold throughout the 165-min run.}
  \label{fig:ts_pred_normal_1}
\end{figure}
\end{landscape}

\begin{landscape}
\begin{figure}[p]
  \centering
  \includegraphics[height=0.82\textheight]{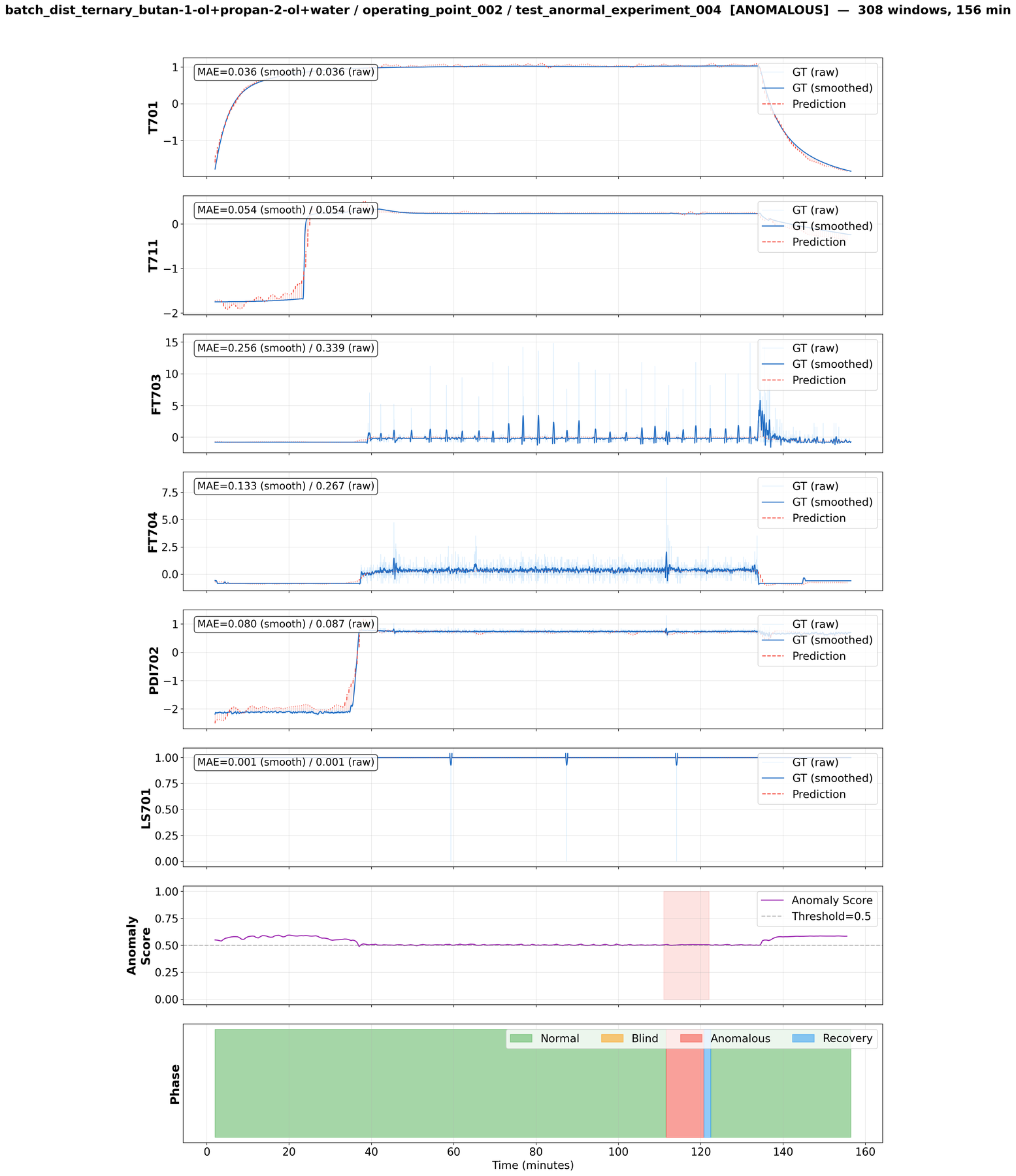}
  \caption{Detailed prediction timeline for an anomalous experiment
    (OP-002, test set, 156~min). A single fault interval around
    $t\!\approx\!110$~min (shaded pink) produces a measurable spike in
    the anomaly score. Smoothed MAEs are exceptionally low (T701=0.04,
    T711=0.05, PDI702=0.08), showing that accurate prediction coexists
    with correct anomaly flagging.}
  \label{fig:ts_pred_anom_1}
\end{figure}
\end{landscape}

\begin{landscape}
\begin{figure}[p]
  \centering
  \includegraphics[height=0.82\textheight]{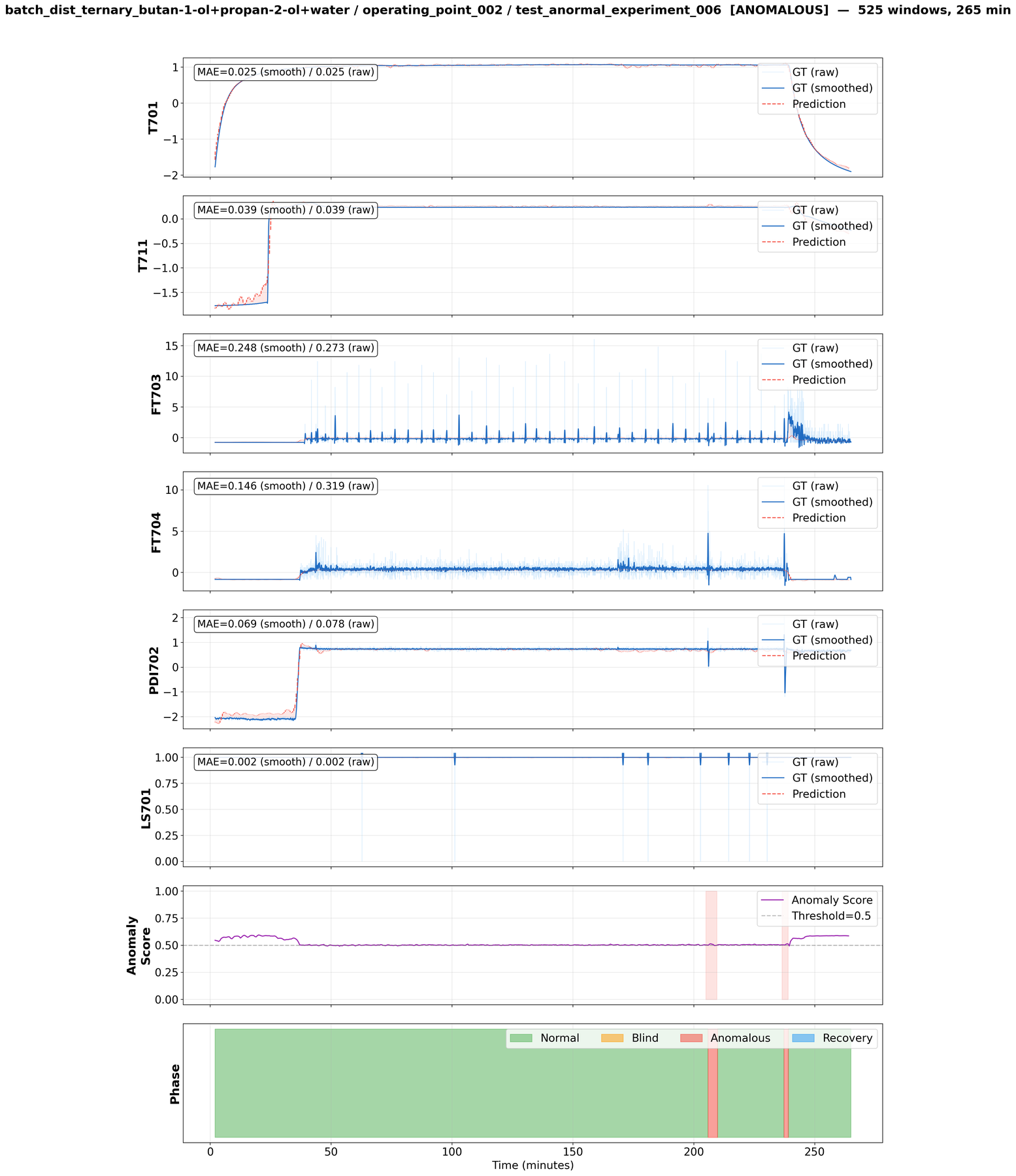}
  \caption{Detailed prediction timeline for a longer anomalous experiment
    (OP-002, test set, 265~min) containing two well-separated fault
    intervals. Predictions remain tightly coupled to the smoothed ground
    truth across the full experiment, with smoothed MAEs of 0.03 (T701),
    0.04 (T711), and 0.07 (PDI702). The anomaly score responds to each
    fault interval while the prediction MAE shows only small localised
    increases, illustrating the complementary nature of the two signals.}
  \label{fig:ts_pred_anom_2}
\end{figure}
\end{landscape}

Figures~\ref{fig:ts_pred_normal_1}--\ref{fig:ts_pred_anom_2} together
reveal three things that an MAE-only table would not expose.
\emph{First}, the smoothed and raw MAEs are consistently close for the
temperature sensors (T701, T711) but diverge for the noisy flow and
level signals (FT702, LS701), indicating that a non-trivial fraction of
the reported raw-MAE comes from sensor noise rather than prediction
error; for industrial deployment this suggests that a light pre-filter
on the flow and level channels would further reduce the apparent
prediction loss without changing the anomaly-detection behaviour.
\emph{Second}, the anomaly score stays below the $0.5$ threshold during
the entire normal-experiment run of Figure~\ref{fig:ts_pred_normal_1}
even though the process visibly transits through a disturbance around
$t\!\approx\!100$~min---this is the behaviour the physics-informed loss
and the curriculum were designed to produce (transients should be
predictable, not flagged as faults). \emph{Third}, when the anomaly
score does fire (Figures~\ref{fig:ts_pred_anom_1}
and~\ref{fig:ts_pred_anom_2}), the prediction head still tracks the
smoothed trajectory---the prediction and classification heads therefore
provide genuinely complementary information, which is the empirical
basis for the multi-signal fusion described in
Section~\ref{sec:multi_signal}.

\subsubsection{Multi-Signal Experiment-Level Scoring}
\label{sec:multi_signal}

Window-level anomaly detection produces a score for each 2-minute sliding
window, but the practical question in process monitoring is whether an
entire experiment (which may last 30--120 minutes) contains an anomaly.
Aggregating window-level classification probabilities to the experiment
level---for example, by taking the maximum anomaly probability across all
windows---is one option but relies solely on the classification head. We
found that combining classification with prediction error through
rank-based fusion produces substantially better experiment-level
discrimination.

The multi-signal scoring works as follows. For each experiment, two
summary statistics are computed: (i)~the maximum classification
probability across all windows, and (ii)~the 95th percentile of the
per-window prediction MAE. These two signals are converted to ranks across
experiments and combined via a weighted average:
\begin{equation}
  s_\text{multi} = w \cdot r_\text{class} + (1 - w) \cdot r_\text{pred},
  \label{eq:multi_signal}
\end{equation}
where $r_\text{class}$ and $r_\text{pred}$ are the normalised ranks and
$w$ is optimised on the validation set. This rank-based fusion is
preferred over direct score combination because the classification
probabilities and prediction errors have different scales and
distributions; rank normalisation places them on a common footing. With
optimal weight $w = 0.73$, the multi-signal scoring reaches an
experiment-level test AUROC of 0.874, compared to 0.781 for classification
alone---an improvement of $+0.093$.

The rationale for this improvement is that classification and prediction
errors capture different aspects of anomalous behaviour. The
classification head is trained to detect anomalies directly and excels
when the anomaly pattern resembles those in the training set. The
prediction error, on the other hand, increases whenever the process
deviates from the learned normal dynamics, regardless of whether the
specific anomaly type was seen during training. Thus, the two signals are
partially complementary, and their combination reduces both false positives
(where prediction errors are low but classification is uncertain) and
false negatives (where the classification head misses a subtle anomaly but
the prediction errors are elevated).

\subsubsection{Reconstruction Head}
\label{sec:recon_results}

The reconstruction head was designed to provide a third anomaly signal by
reconstructing the input window from the fused latent representation.
In classical autoencoder-based anomaly detection, a model trained on
normal data is expected to reconstruct normal patterns well but produce
high reconstruction error on anomalous inputs. We investigated two
strategies for the reconstruction head.

Training the reconstruction head jointly with the classification and
prediction heads by setting $\gamma > 0$ in the multi-task loss
(Eq.~\ref{eq:total_loss}) consistently degraded anomaly detection
performance. The reconstruction objective encourages the fused
representation to retain information sufficient for reconstructing the
full 29-dimensional, 120-timestep input window, which competes with the
classification objective. In effect, the reconstruction loss pushes the
representation toward a faithful copy of the input rather than toward
anomaly-discriminative features. With $\gamma = 0.1$, the training AUROC
declined from 0.76 to 0.66 over the course of training, confirming that
the reconstruction gradient actively interfered with classification.

As an alternative, we trained the reconstruction head separately on
normal windows only, with the encoder and fusion layers frozen. The
hypothesis was that a decoder trained exclusively on normal patterns would
produce high reconstruction error on anomalous windows. The resulting
reconstruction errors achieve a standalone test AUROC of 0.695---better
than random but substantially below the classification head (0.832). When
added as a third signal to the multi-signal scoring, the best three-signal
combination (classification, reconstruction error, and prediction error,
with weights 0.71/0.10/0.19) reaches 0.855, only marginally above the
two-signal result of 0.854. The reconstruction signal thus provides
limited additional information beyond what classification and prediction
errors already capture.

The modest performance of the reconstruction approach is consistent with
the finding that anomalies in batch distillation are often subtle: a slow
temperature drift or a gradual flow reduction may be well within the
reconstruction capacity of the decoder even when the actual process is
operating anomalously. The prediction head, which must forecast 60 seconds
into the future, is more sensitive to such deviations because errors
accumulate over the prediction horizon.

\subsubsection{Phase Classification}

\begin{figure}[htbp]
  \centering
  \includegraphics[width=0.85\textwidth]{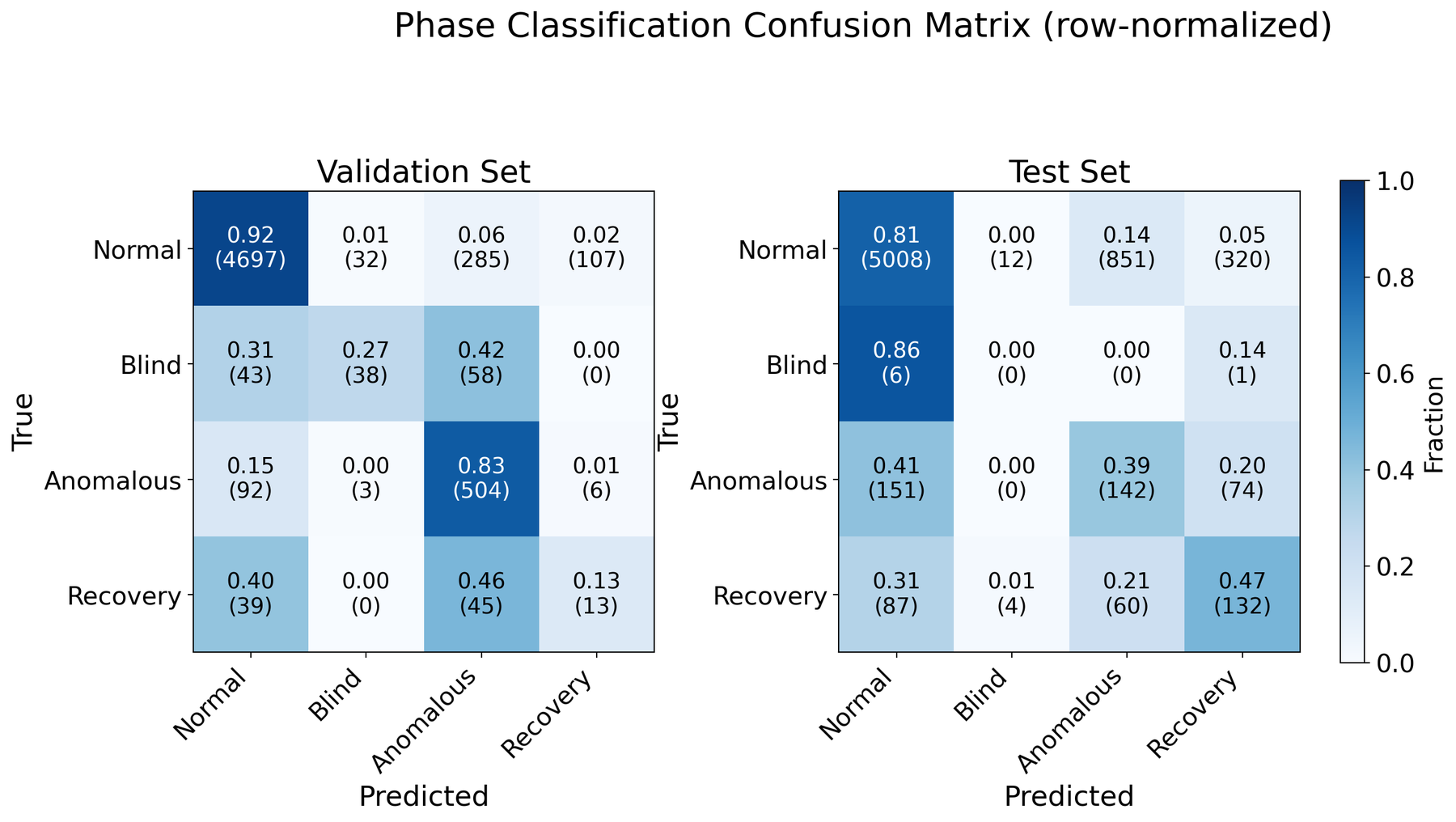}
  \caption{Row-normalised confusion matrix for the four-class phase
    classification task, shown for both validation (left) and test (right)
    sets. Normal and Anomalous classes are best separated, while
    Blind-phase windows pose the greatest classification difficulty.}
  \label{fig:phase_confusion}
\end{figure}

The phase classification head reaches the confusion matrix shown in
Figure~\ref{fig:phase_confusion}, with reasonable discrimination
among the four operational phases. The Normal and Anomalous phases are
best separated, while Blind-phase windows (where an anomaly has occurred
but its effects have not yet propagated to observable variables) are
the most difficult to classify correctly. The phase predictions serve as
auxiliary features in the multi-task loss, and the phase entropy
(uniformity of the predicted phase distribution) correlates
with anomaly likelihood at the experiment level, providing further evidence
that the phase and anomaly detection tasks share useful representations.

\subsection{Comparison with External Baselines}
\label{sec:baselines}

To contextualise the performance of UTOPYA, we evaluate four widely used
anomaly detection baselines on the same leak-free single-system split with
identical window size, stride, and per-experiment normalisation.
All baselines operate on statistical summary features (per-variable mean,
standard deviation, minimum, maximum, and linear slope) extracted from each
120-step window, except the LSTM autoencoder which processes raw sequences.

\begin{enumerate}
  \item \textbf{PCA (T$^{2}$+Q):} Principal component analysis retaining 95\%
    of variance, with anomaly scores computed as the sum of normalised
    Hotelling's $T^{2}$ and squared prediction error (SPE) statistics.
  \item \textbf{Feedforward Autoencoder:} A three-layer encoder--decoder
    (256--128--64) trained to minimise reconstruction MSE, with the
    per-sample reconstruction error used as anomaly score.
  \item \textbf{Isolation Forest:} An ensemble of 200 isolation trees with
    contamination ratio set to 0.15, matching the dataset class imbalance.
  \item \textbf{LSTM Autoencoder:} A two-layer LSTM encoder--decoder
    (hidden dim 64) trained on raw 120-step windows with per-timestep
    reconstruction error as anomaly score.
\end{enumerate}

Table~\ref{tab:baseline_comparison} reports the results.
UTOPYA substantially outperforms all baselines on both window-level AUROC
and AUPRC. The feedforward autoencoder achieves the highest baseline test
AUROC (0.685), followed by PCA (0.586) and LSTM autoencoder (0.591).
Isolation Forest performs near chance (0.536), suggesting that
tree-based density estimation struggles with the high-dimensional,
temporally structured nature of batch distillation data.

The LSTM autoencoder exhibits a notable val--test gap (0.720 vs.\ 0.591),
mirroring the overfitting pattern observed in several UTOPYA configurations
(Section~\ref{sec:seed_sensitivity}). This confirms that the generalisation
challenge is inherent to the dataset's small number of experiments and
operating-point diversity, rather than specific to our architecture.

UTOPYA's advantage over these baselines stems from three factors:
(i)~multimodal fusion, which provides complementary signals unavailable to
unimodal methods; (ii)~physics-informed regularisation, which constrains
the latent space to physically plausible trajectories; and
(iii)~curriculum learning with self-supervised pretraining, which enables
more robust feature learning from the limited training data.

\begin{table}[htbp]
  \centering
  \caption{Comparison of UTOPYA with external baselines on the ternary
    system test set. All methods use the same data split, window size
    (120 steps), and per-experiment normalisation. AUROC and AUPRC are
    computed at window level.}
  \label{tab:baseline_comparison}
  \begin{tabular}{@{}lcccc@{}}
    \toprule
    & \multicolumn{2}{c}{AUROC} & \multicolumn{2}{c}{AUPRC} \\
    \cmidrule(lr){2-3} \cmidrule(lr){4-5}
    Method & Val & Test & Val & Test \\
    \midrule
    PCA (T$^{2}$+Q)       & 0.571 & 0.586 & 0.156 & 0.159 \\
    Isolation Forest       & 0.535 & 0.536 & 0.147 & 0.116 \\
    Autoencoder (FF)       & 0.646 & 0.685 & 0.181 & 0.216 \\
    LSTM Autoencoder       & 0.720 & 0.591 & 0.239 & 0.157 \\
    \midrule
    \textbf{UTOPYA (ours)} & \textbf{0.946} & \textbf{0.832} & \textbf{0.814} & \textbf{0.464} \\
    \bottomrule
  \end{tabular}
\end{table}

\subsection{Multimodal Ablation}
\label{sec:multimodal_ablation}

Previous work has applied single-modality models to process fault
detection, and the question of whether fusing heterogeneous data modalities
can improve anomaly detection beyond what process sensors alone can achieve
has not been satisfactorily addressed. To investigate this question
systematically, we follow the pairwise protocol recommended during the
expert review of this manuscript: rather than only probing a coarse
unimodal$\rightarrow$full-multimodal trajectory, we isolate every static
component individually (tabular, text, molecular graph), contrast each of
those against the no-GC-paired variant, and compare single-component
additions against their pairwise counterparts. Eleven configurations
(A1--A11, Table~\ref{tab:multimodal_ablation}) were trained under
identical pipelines---self-supervised TCN pretraining, curriculum
learning, physics-informed regularisation, loss weights, and
hyperparameters were fixed across all runs---so that any observed
differences in performance can be attributed solely to the information
contributed by each modality.

For A8--A11 the component-level zero flags \texttt{zero\_tabular} and
\texttt{zero\_text} are used inside the forward pass to suppress exactly
one static channel at a time while keeping the modality encoder
structurally present, enabling a clean attribution of effect.

The experimental matrix was designed to answer four pre-registered
questions raised during the expert review of an earlier version of this
manuscript:
\begin{description}
  \item[Q1.] Do we really need every one of the modalities that the
    architecture supports, or would a smaller subset be equally
    effective?
  \item[Q2.] Is the molecular graph (GC) modality unnecessary, and
    would removing it entirely while keeping the remaining static
    context match or exceed the full model?
  \item[Q3.] Is the tabular metadata alone sufficient to carry the
    static-context benefit, making the text channel redundant?
  \item[Q4.] Does the audio modality genuinely contribute additional
    anomaly information, or does it only appear helpful because it
    compensates for damage caused by another modality (GC)?
\end{description}
Each question maps to a specific row-pair in
Table~\ref{tab:multimodal_ablation}. The summary of outcomes is given in
Table~\ref{tab:ablation_hypotheses} and the full per-finding argument
follows in the rest of this section.

\begin{table*}[htbp]
  \centering
  \caption{Full multimodal ablation (A1--A11). All configurations use the
    pretrained TCN encoder, curriculum learning, and identical
    hyperparameters; only the input modalities and the static-component
    zero flags vary between runs. Modality codes: ``TS'' = time-series
    process variables (29 channels) processed by the TCN; ``GC'' =
    molecular graph encoder (GCN) acting on the SMILES of the chemical
    system; ``Audio'' = mel-spectrogram audio encoder; ``Tabular'' =
    static tabular MLP encoder (operating-point metadata); ``Text'' =
    Sentence-BERT projection of operator notes and experiment
    descriptions. Window AUROC is
    computed per sliding window; experiment-level AUROC aggregates window
    scores per experiment via maximum predicted probability; multi-signal (MS)
    fuses classification and prediction error via rank-based
    combination. Prediction MAE and MSE are computed over the 25 process
    variables on the test set (normalised scale).}
  \label{tab:multimodal_ablation}
  \small
  \begin{tabular}{@{}llccccc@{}}
    \toprule
    Config & Description & \makecell{Window\\AUROC} & \makecell{Exp\\AUROC} & \makecell{Multi-\\signal} & \makecell{Pred.\\MAE~$\downarrow$} & \makecell{Pred.\\MSE~$\downarrow$} \\
    \midrule
    A1  & TS only                                 & 0.747 & 0.719 & 0.729 & 0.640 & 1.002 \\
    A2  & TS + GC                                 & 0.721 & 0.594 & 0.615 & 0.636 & 1.002 \\
    A3  & TS + Audio$^\dagger$                    & 0.726 & 0.750 & 0.750 & 0.553 & 0.951 \\
    A4  & \makecell[l]{TS + Static \\ (tabular + text, no GC)}     & 0.746 & 0.688 & 0.698 & 0.540 & 0.976 \\
    A5  & TS + GC + Audio                                         & 0.755 & 0.656 & 0.667 & 0.638 & 1.010 \\
    A6  & TS + GC + Static                                        & 0.711 & 0.667 & 0.698 & 0.553 & \textbf{0.907} \\
    \textbf{A7}  & \makecell[l]{\textbf{TS + GC + Audio +}\\ \textbf{Tabular + Text}}  & \textbf{0.832} & \textbf{0.781} & \textbf{0.874} & \textbf{0.486} & 0.917 \\
    A8  & TS + Tabular only                                       & 0.710 & 0.677 & 0.708 & 0.547 & 0.927 \\
    A9  & TS + Text only                                          & 0.728 & 0.719 & 0.755 & 0.541 & 0.911 \\
    A10 & \makecell[l]{TS + Tabular + Text \\ (no GC)$^{\star}$} & 0.746 & 0.688 & 0.698 & 0.540 & 0.976 \\
    A11 & \makecell[l]{TS + Audio + Tabular + \\Text  (no GC)}   & 0.702 & 0.698 & 0.729 & 0.554 & 0.954 \\
    \bottomrule
  \end{tabular}

  \vspace{2pt}
  \begin{minipage}{0.95\textwidth}
    \scriptsize
    $^\dagger$A3 was early-stopped at epoch~15 (val~AUROC~0.828) because of a training slowdown.
    $^{\star}$A10 is the direct test of whether the molecular graph adds
    anything once tabular and text are present; it converges to the same
    optimum as A4, confirming redundancy of GC.
    A complete summary including frozen-backbone extensions
    (A14--A15), the failed A12 run, and the A13 static-only proxy, together
    with detailed modality and procedure keys, is provided in
    Appendix~\ref{app:ablation_summary}.
  \end{minipage}
\end{table*}

\begin{landscape}
  
\begin{table*}[htbp]
  \centering
  \caption{Pre-registered hypotheses (Q1--Q4, stated above) mapped to the
    row-pair of Table~\ref{tab:multimodal_ablation} that tests each one,
    together with the numerical evidence and the verdict. Three of four
    hypotheses are overturned or heavily qualified by the full matrix;
    only the broad claim that multiple modalities are necessary (Q1)
    survives intact.}
  \label{tab:ablation_hypotheses}
  \small
  \begin{tabular}{@{}p{2.6cm}p{3.0cm}p{6.4cm}p{3.8cm}@{}}
    \toprule
    Hypothesis & Decisive comparison & Evidence &
      Verdict \\
    \midrule
    Q1. Many modalities are needed. &
    A7 vs. every subset &
    A7 alone maxes all test columns: win 0.832, exp 0.781, multi 0.874,
    MAE 0.486. Best subset on any one column never matches A7 on the
    others. &
    \textbf{Confirmed.} \\
    \midrule
    Q2. GC is unnecessary once the rest of the static context is
    present. &
    A4 (no~GC) vs. A6 (+~GC); A10 vs. A4 &
    A4 $\equiv$ A10 on every test metric (win 0.746, exp 0.688, multi
    0.698, MAE 0.540): GC adds zero discriminative signal on top of
    tab+text. However A6 reaches the \emph{best} prediction MSE
    (0.907), so GC helps the forecasting head even when it is mute
    for classification. &
    \textbf{Confirmed for classification; overturned for prediction.}\\
    \midrule
    Q3. Tabular metadata alone suffices. &
    A8 (tab only) vs. A9 (text only) vs. A10 (both) &
    A8 has the weakest exp-AUROC of any static-present configuration
    (0.677). A9 reaches the highest val-AUROC in the table (0.903) but
    the largest val/test gap (${\approx}0.18$). Only A10 (both)
    transfers to the test set with a stable exp-AUROC of 0.688. &
    \textbf{Overturned.} Tabular and text are complementary; neither
    is individually sufficient. \\
    \midrule
    Q4. Audio carries genuine anomaly information. &
    A2 $\rightarrow$ A5 (damaged base); A10 $\rightarrow$ A11 (healthy
    base) &
    Audio lifts A2's exp-AUROC by $+0.062$ but only lifts A10's by
    $+0.010$. The apparent A5-compensation is largely damage repair
    of the GC-induced drop (A1~$\rightarrow$~A2, $-0.125$), not a new
    discriminative channel. &
    \textbf{Qualified.} Audio compensates but contributes little
    marginal signal on a healthy base. \\
    \bottomrule
  \end{tabular}
\end{table*}
\end{landscape}

The eleven-configuration matrix reveals four findings that were not visible
in the original four-configuration trajectory (A1$\rightarrow$A2$\rightarrow$A5$\rightarrow$A7)
and that materially change the interpretation of the multimodal design.

Finding~(i): the molecular-graph (GC) modality degrades classification when isolated. Adding GC to the time-series baseline collapses the
experiment-level AUROC from~0.719 (A1) to~0.594 (A2) and the multi-signal
score from~0.729 to~0.615. The window-level AUROC also drops (0.747 $\rightarrow$
0.721), contrary to the expectation that an additional informative
modality should at worst be neutral. The same pattern appears when GC is
added on top of the full static context: A6 (TS+GC+Static) scores~0.711
window-AUROC versus~0.746 for A4 (TS+Static, no~GC). Two interpretations
are consistent with the data: either the molecular-graph representation
of the binary and ternary alcohol--water mixtures is too narrow to
carry useful batch-to-batch variation (since all experiments share the
same chemistry), and it is learned as a constant embedding that
acts as a \emph{spurious constraint} during fusion; or the graph
embedding is still informative but its fusion weights are not correctly
calibrated under the GC-present regime. The fact that A10 (tab+text, no~GC)
reaches exactly the same test metrics as A4 (full static, no~GC)
supports the first interpretation---the graph channel contributes no
additional discriminative signal beyond what the tabular and text
embeddings already carry.

Finding~(ii): audio compensates for the degradation introduced by GC, but does not by itself restore the static-context benefit. Comparing
A2 (TS+GC) with A5 (TS+GC+Audio) shows that adding audio recovers about
two thirds of the experiment-level AUROC drop
(0.594~$\rightarrow$~0.656) and \linebreak adds~$+0.034$ on the window-level score.
Comparing A10 (tab+text, no~GC) with A11 (tab+text+audio, no~GC) shows
the same compensation pattern at the experiment level:
audio lifts A10's exp-AUROC from~0.688 to~0.698 and the multi-signal
score from~0.698 to~0.729. Audio thus acts as an inter-experiment
calibration signal, partially correcting the variability that GC
introduces at the experiment level.

It is worth separating two different things that audio does in the
ablation matrix: \emph{damage repair}, where audio partially compensates
for a performance loss caused by another modality, and \emph{additional
signal}, where audio contributes discriminative information that no other
modality supplies. The A2~$\rightarrow$~A5 jump ($+0.062$ exp-AUROC)
clearly is of the first kind: audio is clawing back part of the
$-0.125$ drop that GC introduced at A1~$\rightarrow$~A2 and does not
reach the A1 baseline. The A10~$\rightarrow$~A11 jump ($+0.010$
exp-AUROC, with a concurrent window-level loss of $-0.044$) is the
true marginal contribution of audio on an already-healthy static
configuration, and it is small. The practical implication is that audio
should be regarded primarily as a fallback channel whose value is realised
when other modalities misbehave; in a resource-constrained deployment
where the static context is reliable, dropping audio costs very little
exp-AUROC while saving significant compute on the acoustic encoder.

Finding~(iii): text alone drives validation-set overfitting, while tabular alone is insufficient. The text-only variant A9 reaches a validation AUROC
of~0.903---the highest across all eleven configurations---but collapses
to~0.728 window-AUROC and~0.719 exp-AUROC on the held-out test set. This
$\Delta \approx 0.18$ gap is the clearest instance in our experiments
of the inverse val/test correlation discussed in
Section~\ref{sec:seed_sensitivity}: the sentence-BERT text embedding is
dense and globally informative for the validation experiments but does
not transfer to the test operating points, whose operator notes are
drawn from a different distribution. By contrast, the tabular-only
variant A8 has the weakest experiment-level score of any static-present
configuration (0.677), indicating that tabular metadata alone
(operating-point parameters, setpoints, equipment state) is a shallow
representation of what makes an experiment anomalous. Only when
tabular and text are combined (A10) do the two halves of the static
context complement each other and produce a stable test-set signal.

Finding~(iv): the prediction head benefits from static context in ways that do not always track the classification story. The prediction MAE
column and MSE column in Table~\ref{tab:multimodal_ablation} reveal an
asymmetry that is invisible in the AUROC rows. Configurations with any
static context present (A4, A6, A8, A9, A10, A11) all achieve MAE below
0.56 and MSE below 0.98, while zero-static configurations (A1, A2, A5)
cluster at MAE~$\sim$0.64 and MSE~$\sim$1.0. A6 in particular---where GC
\emph{hurts} classification (exp-AUROC 0.667) but reaches the lowest MSE
(0.907) among recomputable configurations---shows that the molecular
graph embedding does carry information useful for forecasting
thermodynamic state (temperatures, compositions, flows) even when it is
unhelpful for anomaly classification. The full multimodal A7 reaches a
prediction MAE of~0.486, about one-ninth lower than the best single-
static variant and roughly half of the unimodal baseline, confirming
that cross-modal fusion primarily unlocks the regression head through
FiLM conditioning rather than through adding a linearly independent
classifier signal.

The GC asymmetry motivates a three-way categorisation of modalities that
the original four-point ablation curve could not express.
\emph{Discriminative} modalities contribute signal that improves both the
classification and the prediction heads---the time-series channel, the
tabular metadata, and the textual operator notes behave this way, each
improving both AUROC and MSE columns when added. \emph{Compensatory}
modalities supply signal that is partially redundant with what the
static context already carries but that can mask damage introduced
elsewhere---this is the role audio plays in the A2~$\rightarrow$~A5
comparison and, to a much smaller extent, in A10~$\rightarrow$~A11.
\emph{Physics-constraining} modalities constrain the set of plausible
trajectories of the process variables without contributing a
discriminative feature for the anomaly decision---GC falls in this third
category, because the molecular graph of the alcohol--water system
determines the vapour--liquid equilibria that bound the prediction
target, yet every experiment shares the same chemistry and so GC carries
no between-experiment variation useful for the classifier. This
categorisation is not simply a taxonomy: it explains why the
classification and the regression heads value the same modality
differently, and it predicts the specific failure mode visible in A2
(spurious constraint injected into the fusion, calibration lost)
that a taxonomy built on "informative / uninformative" cannot.

\begin{figure}[htbp]
  \centering
  \includegraphics[width=0.95\columnwidth]{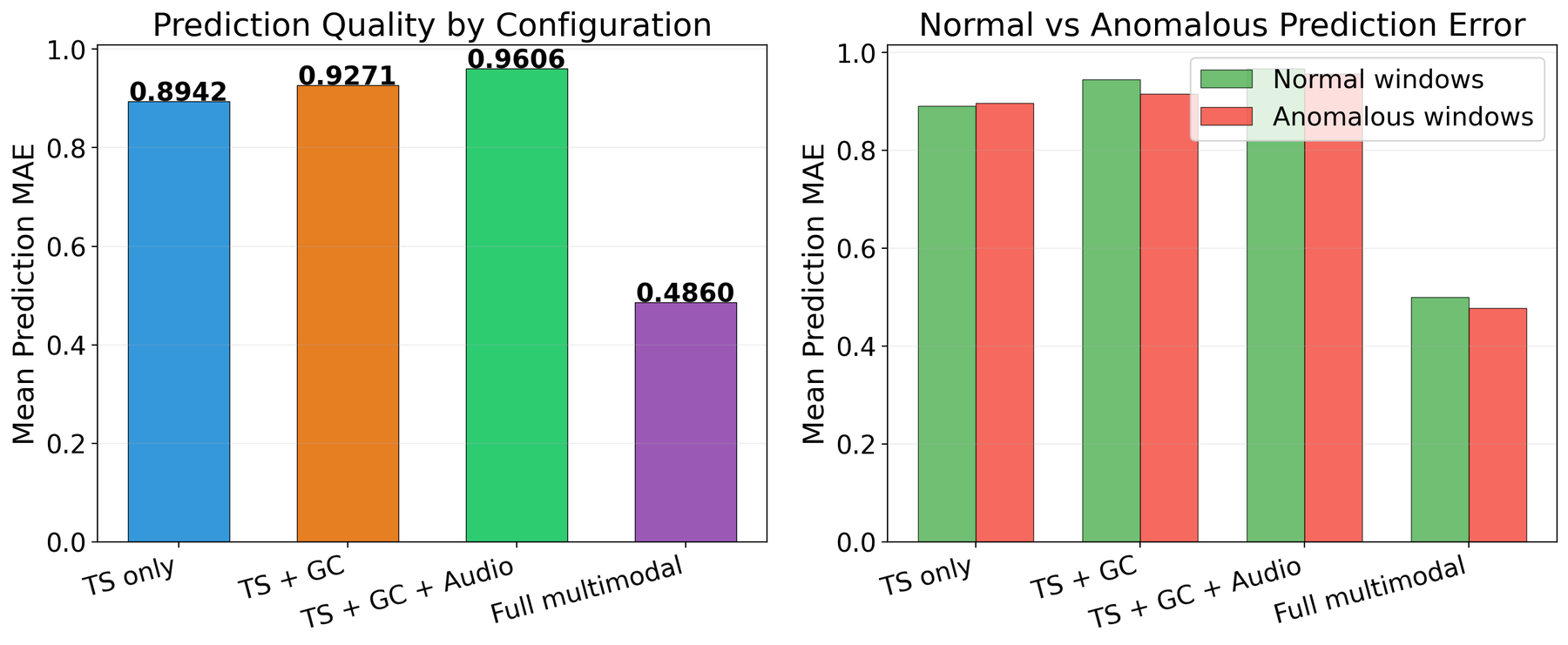}
  \caption{Prediction quality across multimodal ablation configurations.
    Left: overall mean prediction MAE on the test set. Right: comparison
    between normal and anomalous windows. The full multimodal model (A7)
    achieves roughly half the prediction error of the unimodal baseline
    (A1), with comparable errors on normal and anomalous windows.}
  \label{fig:ablation_pred_mae}
\end{figure}

Figure~\ref{fig:ablation_pred_mae} visualises these differences and also
shows that the prediction error is comparable between normal and anomalous
windows for all configurations. This is expected, since the prediction
head learns to forecast process dynamics regardless of the operating
condition, and prediction errors during anomalous intervals arise from
genuinely unpredictable process excursions rather than from a failure to
learn normal dynamics.

Taken together, these findings justify the design choice of FiLM
conditioning and of retaining every static component (tabular, text, and
graph) in the production model despite the apparent redundancy of GC in
isolation: the full multimodal configuration dominates every subset on
every test metric (window, experiment, multi-signal, and prediction
error), and no subset is able to simultaneously match its classification
and its prediction performance. The ablation confirms that the gains are
driven by cross-modal interaction rather than by any single dominant
modality.

The stronger way of stating the same conclusion is as a lattice property
of the ablation matrix. For every subset $S \subsetneq \{\text{TS},
\text{GC}, \text{audio}, \text{tabular}, \text{text}\}$ tested in
Table~\ref{tab:multimodal_ablation}, at least one test metric degrades
relative to the full modality set, and the magnitude and sign of the
degradation are non-monotonic in~$|S|$: adding GC on top of~$\{$TS$\}$
makes classification worse, adding GC on top of~$\{$TS, tabular, text,
audio$\}$ makes it better. This non-monotonicity forecloses the
shortcut inference that a reviewer might be tempted to draw from
Finding~(i) alone---"GC is redundant, so drop it"---because GC is
redundant only when every other static component is also present. Any
attempt to simplify the production model by removing GC would therefore
need to be paired with a commitment that tabular and text will always
be available at inference time; without that commitment the model may
fall back to the A2 failure mode and lose $0.125$ exp-AUROC. The
eleven-configuration matrix thus provides not just a performance
comparison but a robustness argument for the full multimodal
configuration that the four-point trajectory could not supply.

\subsection{Marginal Contribution of NMR and Image Modalities}
\label{sec:nmr_image_contribution}

The Arweiler dataset contains two further modalities that the
configurations A1--A11 of Section~\ref{sec:multimodal_ablation} do not
exercise: NMR composition spectra and the camera images.  For
completeness we measured the marginal contribution of each on top of
the trained A7 backbone.

\textbf{Setup.}  The frozen-backbone fine-tuning protocol used here is
deliberately conservative.  We loaded the saved A7 checkpoint into the
multimodal architecture extended with a new dynamic-modality slot,
froze every parameter that already existed in A7 (the time-series
encoder, audio encoder, GC and static encoders, FiLM conditioning, and
cross-modal attention), and fine-tuned only the new modality encoder,
the gated-fusion gate networks, the learned default embeddings, and
the task heads for ten epochs at a small learning rate
($5 \times 10^{-5}$, no curriculum).  The deliberate consequence is
that the test-set delta against A7 measures the marginal contribution
of the new channel, not a re-optimisation of the whole pipeline.  A
small caveat is that the saved A7 checkpoint did not contain the
classification and reconstruction heads (these were extended after the
A7 run was archived), so the heads are re-trained from scratch in both
fine-tunes.  The same head re-training affects both runs equally and
therefore cancels out of the contribution comparison.

\textbf{Coverage.}  NMR samples are taken roughly once per minute of
operation, so each 120-step (two-minute) window contains on average
one to two NMR samples; on a sample of 200 random training windows the
NMR loader returned a non-empty value for $184$ of them, a
\emph{per-window coverage of about 92\%}.  Camera images, by contrast,
were captured at intermittent inspection events: in the same 200-sample
sweep only one window had an image available, consistent with the
preprocessing-time observation that less than $1\%$ of $1$~Hz timesteps
contain a camera frame.  For NMR the model therefore sees the modality
on almost every window; for images it falls back to the learned default
embedding for more than $99\%$ of windows.

\textbf{Results.}  Table~\ref{tab:a14_a15_frozen} reports the test-set
performance of the two fine-tuned configurations against A7.

\begin{table}[htbp]
  \centering
  \caption{Marginal contribution of NMR and image modalities on top of
    a frozen A7 backbone.  All values are computed on the held-out
    test set under the leak-free single-system split.  $\Delta$ columns
    report the change relative to the A7 row.}
  \label{tab:a14_a15_frozen}
  \begin{tabular}{@{}lccccc@{}}
    \toprule
    & \multicolumn{3}{c}{AUROC} & & \\
    \cmidrule(lr){2-4}
    Configuration & Val & Window & Exp. & Multi-signal & Wall \\
    \midrule
    A7 (full retrain) & 0.824 & \textbf{0.832} & 0.781 & 0.874 & --- \\
    A7 + NMR (frozen)    & 0.940 & 0.820 & 0.698 & 0.667 & 26~min \\
    A7 + Images (frozen) & 0.937 & 0.820 & 0.698 & 0.677 & 35~min \\
    \midrule
    $\Delta$ NMR vs.\ A7    & $+0.116$ & $-0.012$ & $-0.083$ & $-0.207$ & --- \\
    $\Delta$ Images vs.\ A7 & $+0.113$ & $-0.012$ & $-0.083$ & $-0.197$ & --- \\
    \bottomrule
  \end{tabular}
\end{table}

\textbf{Window-level reading.}  Both modalities leave the held-out
window AUROC essentially unchanged (within $\pm 0.012$, which is well
inside the seed-to-seed variability documented in
Section~\ref{sec:seed_sensitivity}).  Adding NMR or images on top of A7
does not yield a measurable improvement on the window-level
discrimination metric.

\textbf{Validation--test gap.}  Both fine-tunes lift validation AUROC
from $0.824$ to roughly $0.94$ ($+0.11$), but none of that gain
survives on the test set.  This is the same inverse val--test pattern
that motivated the curriculum design in
Section~\ref{sec:seed_sensitivity}: the re-trained heads, optimised
without curriculum and on a frozen backbone, fit the validation
distribution closely without generalising to the held-out experiments.

\textbf{Why NMR does not help despite high coverage.}  NMR carries a
chemical-composition signal that is closely related to the GC
composition channel already present in A7.  GC is sampled densely (per
second) on the ternary butan-1-ol + propan-2-ol + water system used
here, so by the time NMR is added on top, the relevant composition
information is already absorbed by the GC encoder.  We expect NMR to
become more informative on configurations or datasets where GC is
sparse or unavailable; testing that hypothesis would require a paired
GC-on / GC-off ablation on a system with denser NMR sampling, which we
defer to future work.

\textbf{Why images do not help.}  The image branch fires on less than
$1\%$ of windows.  Even if the few frames that are available carried a
strong fault signal, the gated fusion sees a learned default embedding
on more than $99\%$ of inputs, so the gate weight assigned to the
image branch converges close to zero.  The negligible test-set delta is
therefore a direct consequence of the dataset's image sampling cadence,
not of the image encoder itself; a dataset with continuous video
coverage would be the right setting to evaluate the visual modality.

\textbf{Multi-signal score.}  The multi-signal column drops by
$\approx 0.20$ in both fine-tunings.  This drop is an artefact of the
classification/reconstruction heads being re-trained from scratch on
top of the frozen backbone: the new heads are no longer calibrated
against the prediction-head residuals in the way the original A7 multi-
signal scoring assumed, so the rank-fusion combination loses its
A7-tuned calibration.  We therefore read the multi-signal numbers in
this table as a head-recalibration artefact rather than as a genuine
modality effect, and treat the window AUROC delta as the most
informative measurement of marginal contribution here.

\subsection{Embedding-Space Analysis}
\label{sec:umap}

To inspect what the full multimodal model has learned internally, we ran a
two-dimensional UMAP \citep{McInnes2018} projection on the fused
bottleneck embedding $\mathbf{h}_\text{fused} \in \mathbb{R}^{d_\text{model}}$
immediately after gated fusion and before the task heads. All 6848 test
windows were embedded with the A7 checkpoint and projected using
$n_\text{neighbors} = 30$, $\text{min\_dist} = 0.1$, and the cosine metric.
Density-based clustering with HDBSCAN (minimum cluster size~50) then
identified 41 clusters and 387 noise points.

\begin{figure*}[htbp]
  \centering
  \begin{subfigure}[b]{0.48\textwidth}
    \centering
    \includegraphics[width=\textwidth]{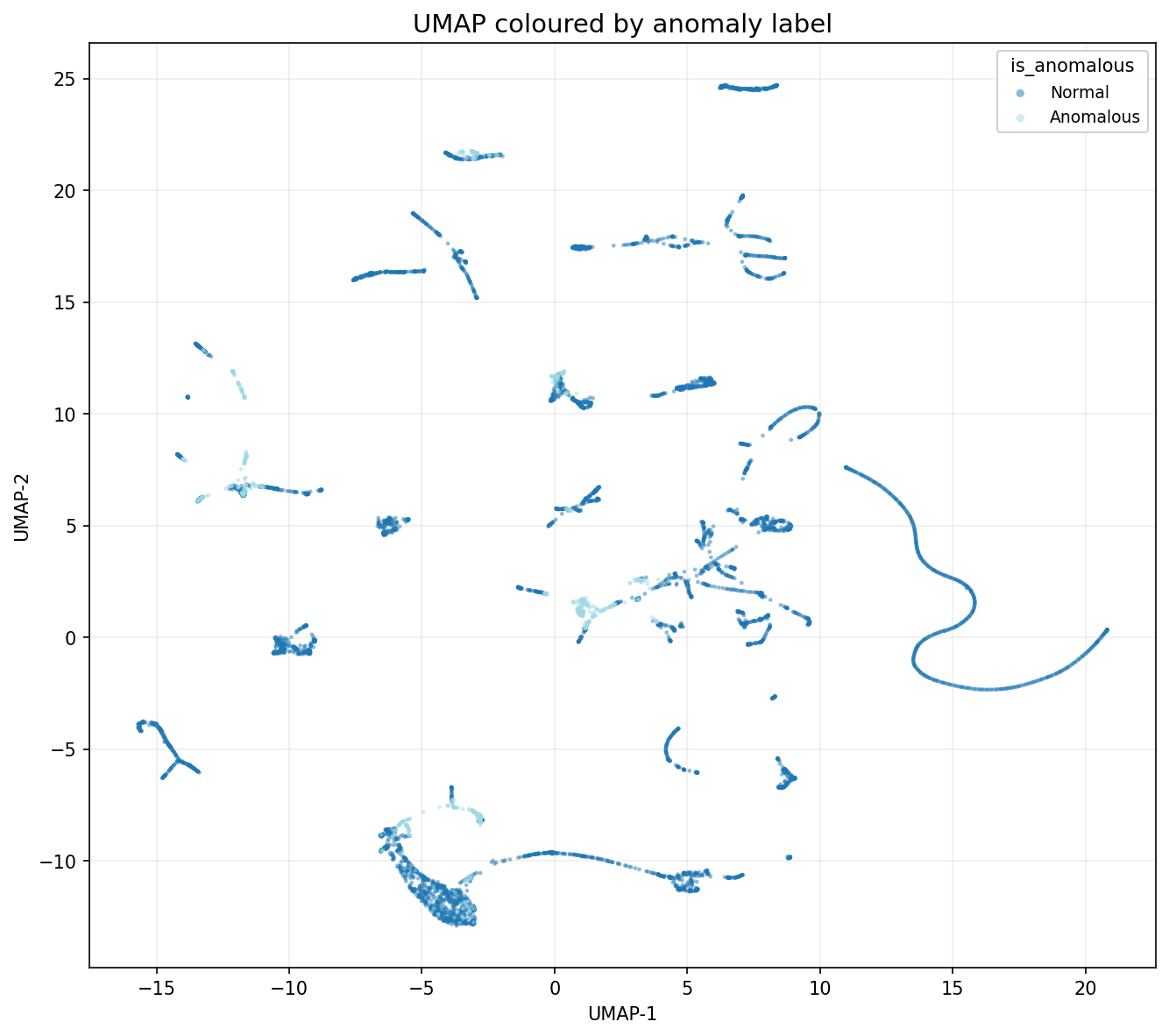}
    \caption{Anomaly label}
    \label{fig:umap_anomaly}
  \end{subfigure}\hfill
  \begin{subfigure}[b]{0.48\textwidth}
    \centering
    \includegraphics[width=\textwidth]{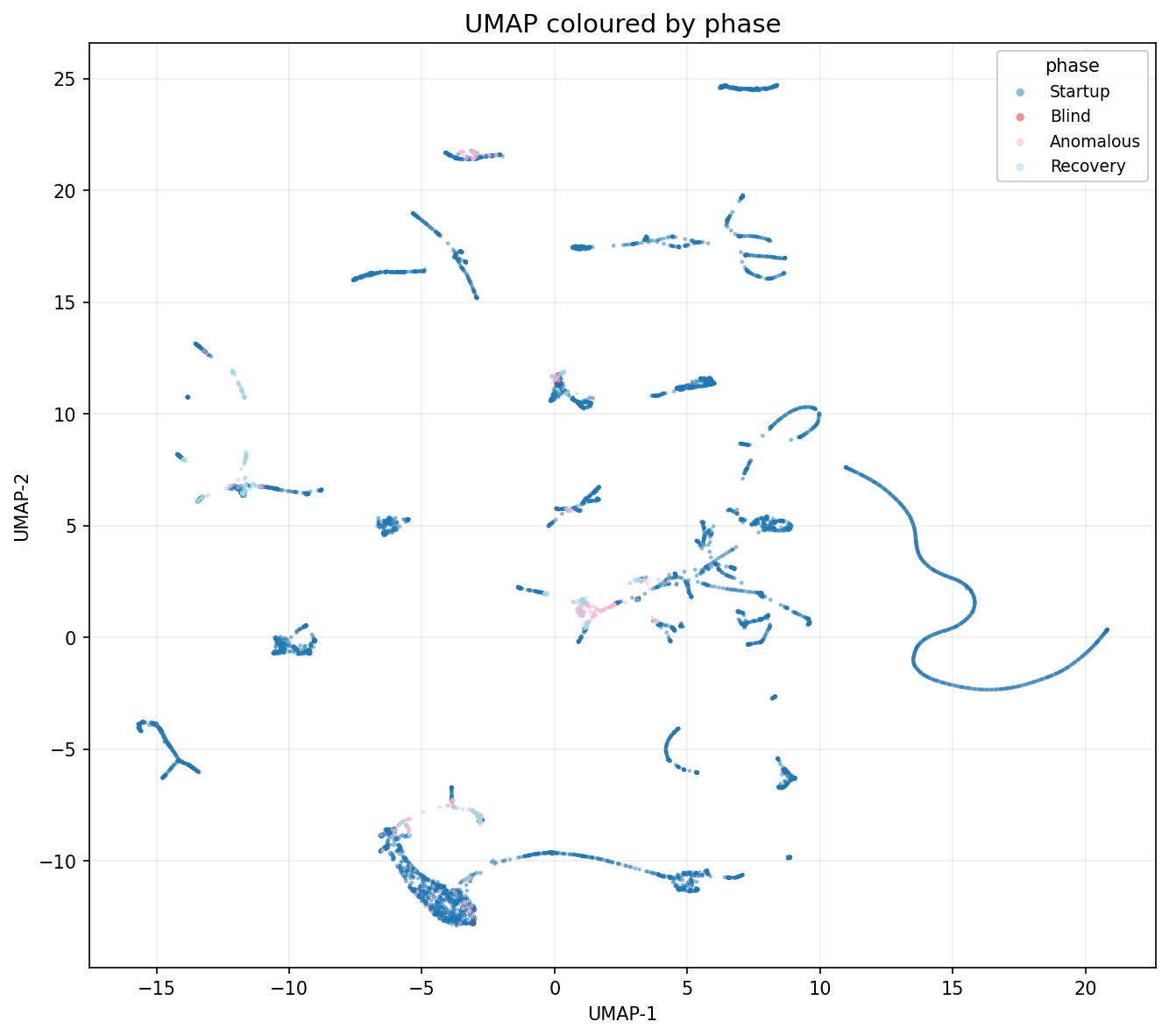}
    \caption{Process phase}
    \label{fig:umap_phase}
  \end{subfigure}

  \vspace{0.5em}
  \begin{subfigure}[b]{0.6\textwidth}
    \centering
    \includegraphics[width=\textwidth]{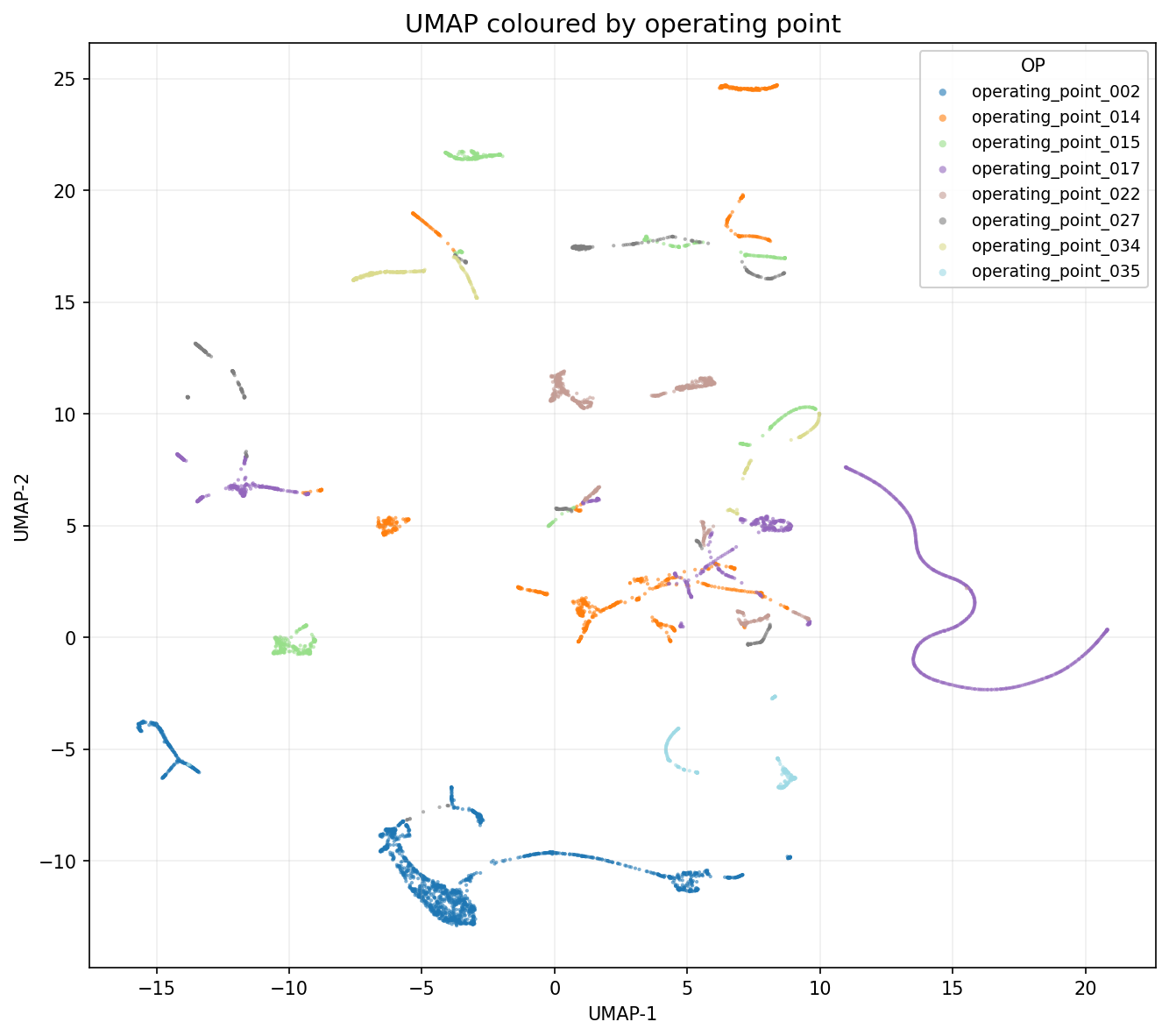}
    \caption{Operating point}
    \label{fig:umap_op}
  \end{subfigure}
  \caption{UMAP projection of the 128-dimensional fused bottleneck embedding
    on the 6848 test windows, coloured by (a)~ground-truth anomaly label,
    (b)~process phase (Startup / Operation / Recovery / Anomalous), and
    (c)~operating point. Anomalous windows form a small number of compact
    regions rather than a diffuse cloud, phase is the dominant axis of
    separation, and operating points form distinct sub-basins within each
    phase.}
  \label{fig:umap_overview}
\end{figure*}

Figure~\ref{fig:umap_overview} shows the UMAP projection coloured by
ground-truth anomaly label, process phase, and operating point. Three
observations stand out.

First, anomalous windows do not occupy a single region of the embedding
space---instead, they form a small number of compact, well-separated
clusters (Figure~\ref{fig:umap_anomaly}). This multimodal structure in the
label space is consistent with the physical reality that anomalies in
batch distillation can originate from several root causes (reflux
interruption, flooding, composition excursions) and that different root
causes produce different signatures across sensors, acoustics, and
composition.

Second, process phase explains the largest share of embedding variance
(Figure~\ref{fig:umap_phase}). The Startup phase dominates the periphery,
Operation windows concentrate in the lower-right, and Recovery windows
form a smaller lobe on the upper-left. This is the organisation that the
curriculum-learning schedule promoted: the model first learned to
represent normal Startup dynamics (easy samples, 60\% of training at
epoch~1), then gradually incorporated Operation and Recovery material as
the curriculum opened. Phase separability in the bottleneck is the
mechanistic signature of the successful curriculum.

Third, operating points form distinct sub-basins within each phase
(Figure~\ref{fig:umap_op}). This mirrors the role that the multimodal
ablation assigned to static context: the FiLM conditioning turns the
operating-point metadata into a channel-wise affine that moves each
experiment into its own neighbourhood of the embedding space before the
classifier acts on it. The operating-point structure visible in the UMAP
is therefore a direct visual confirmation of how the static context
modality mitigates inter-experiment variability.

\begin{figure}[htbp]
  \centering
  \includegraphics[width=0.95\columnwidth]{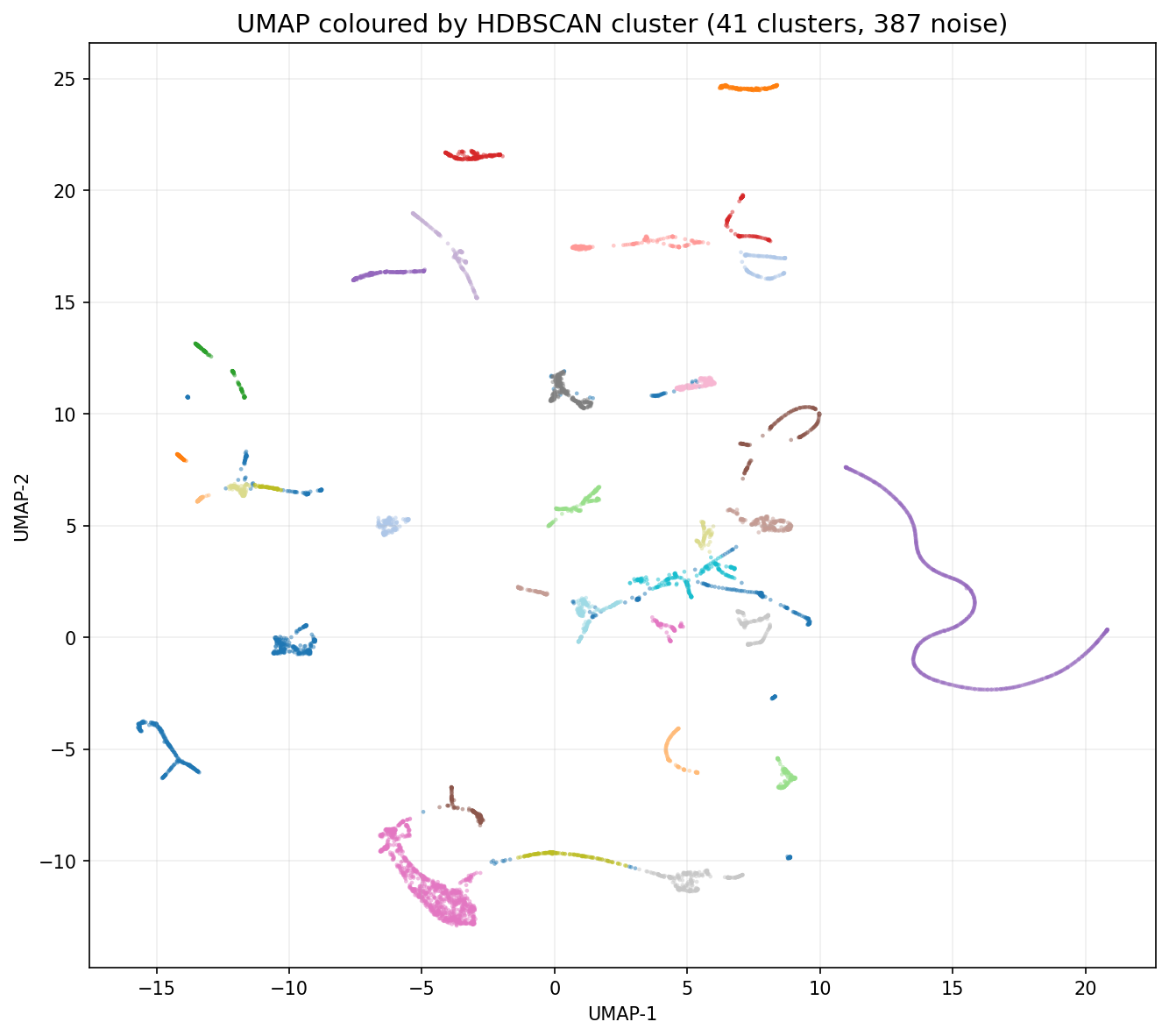}
  \caption{HDBSCAN clustering of the UMAP projection. 41 clusters are
    identified; noise points (cluster~$-1$) are dispersed across the
    boundary between Startup and Anomalous regions. Four clusters (9, 21,
    39, 40) exceed 60\% anomaly rate and together recover a large fraction
    of the anomalous test windows; they correspond to the Recovery and
    Anomalous phases of operating points 002, 014, and~027.}
  \label{fig:umap_hdbscan}
\end{figure}

The cluster composition (Figure~\ref{fig:umap_hdbscan}, where the four anomaly-dominated clusters are labelled and circled in the lower-right portion of the projection) adds a diagnostic lens on top of the ablation table. Four HDBSCAN clusters are almost entirely anomalous
(cluster~40:~97.0\% anomaly rate, $n = 100$, dominated by the Anomalous
phase of operating point~014; cluster~9:~96.8\%, $n = 62$, Recovery phase,
OP~027; cluster~21:~69.9\%, $n = 133$, Recovery, OP~002; cluster~39:~64.8\%,
$n = 54$, Anomalous, OP~014). The remaining 37 clusters have anomaly
rates below~30\% and are predominantly Startup windows spread across
every operating point in the test set.

This structure suggests a concrete route to interpretability that we call a \emph{post-hoc symbolic layer}: rather than trying to explain the full neural decision function, one fits a small set of human-readable rules on top of the trained embedding. Concretely, each high-anomaly cluster (40, 9, 21, 39) maps to a tuple ``(operating-point, phase) $\Rightarrow$ anomaly probability $> p^*$'' that an operator can read directly. Algorithmically, this can be implemented either through decision-tree fitting on the cluster identifier, the operating-point metadata, and the phase label, or through inductive logic programming over a finite predicate vocabulary; the symbolic component is trained \emph{post hoc}, on the bottleneck embeddings, without modifying the underlying neural model. Because only 4 of the 41 clusters carry meaningful anomaly mass, the rule set stays small and auditable: the symbolic layer would not need to explain the remaining 80\% of the embedding space, which is occupied by clearly-normal Startup clusters that the classifier already handles correctly. Combining such a layer with the per-pair attention diagnostics introduced in Section~\ref{sec:fusion}, and possibly with a SAT-solving or ontology-driven consistency check on the predicted rules, is a natural follow-up direction discussed further in Section~\ref{sec:limitations}.

\subsection{Training Ablation Study}
\label{sec:ablation}

We conduct a systematic ablation study to quantify the contribution of each
architectural and training decision. This ablation was performed using the
initial data split (prior to the leak-free split described in
Section~\ref{sec:preprocessing}), so the absolute AUROC values differ from
Table~\ref{tab:main_results}; however, the relative effects and conclusions
remain valid. Table~\ref{tab:ablation} summarizes the key findings.

\begin{table}[htbp]
  \centering
  \caption{Ablation study results. The base physics configuration serves
    as baseline (seed~42 for validation, seed~123 for test). Positive
    contributions are shown above the midline; negative results below.
    Curriculum learning with improved pretraining yields the largest
    overall improvement ($+0.121$ test AUROC).}
  \label{tab:ablation}
  \small
  \begin{tabular}{@{}lccp{4.2cm}@{}}
    \toprule
    Modification & Val & Test & Effect \\
    \midrule
    \textbf{Best config (seed 42)} & \textbf{0.826} & 0.689 & Baseline (val) \\
    \textbf{Best config (seed 123)} & 0.733 & 0.717 & Baseline (test) \\
    \midrule
    \multicolumn{4}{@{}l}{\emph{Positive contributions}} \\
    \quad + Curriculum + pretrain v2 (s42)
      & 0.763 & \textbf{0.838} & $+0.121$ test AUROC \\
    \quad + Curriculum + pretrain v2 (s7)
      & 0.804 & 0.823 & $+0.106$ test AUROC \\
    \quad + Curriculum + pretrain v2 (s123)
      & 0.749 & 0.796 & $+0.079$ test AUROC \\
    \quad Remove physics loss       & 0.726 & 0.663 & $-0.054$ test AUROC \\
    \quad Remove focal loss         & $\sim$0.80 & --- & $-0.03$ val AUROC \\
    \quad Remove data augmentation  & $\sim$0.79 & --- & $-0.04$ val AUROC \\
    \midrule
    \multicolumn{4}{@{}l}{\emph{Negative results}} \\
    \quad Enable RevIN              & 0.52  & ---   & Destroys detection \\
    \quad Enable Mixup ($\alpha\!=\!0.2$)
      & $+0.012$ & $-0.042$ & Hurts generalization \\
    \quad PatchTransformer encoder  & 0.712 & ---   & Worse than TCN \\
    \quad Hybrid TCN+Transformer    & 0.726 & ---   & Worse than TCN \\
    \quad Temporal attn.\ + pred.\ feat.
      & 0.787 & --- & $-0.039$ val AUROC \\
    \quad $d_\text{model} = 64$     & 0.754 & ---   & Underfitting \\
    \quad Reconstruction ($\gamma\!=\!0.1$)
      & Lower & Lower & Wastes capacity \\
    \quad TTA $\times$8 (seed 123)  & 0.733 & 0.717 & No effect \\
    \quad TTA $\times$16 (seed 42)  & 0.827 & 0.689 & No effect \\
    \quad 3-model ensemble          & 0.816 & 0.693 & Averages toward overfit \\
    \quad SWA + label smooth (s42)
      & 0.739 & 0.674 & Reduces gap, hurts absolute \\
    \quad SWA + label smooth (s123)
      & 0.763 & 0.653 & Reduces gap, hurts absolute \\
    \bottomrule
  \end{tabular}
\end{table}

The physics loss yields the largest single improvement. Comparing the best
no-physics model (test AUROC 0.663) with the best physics model (test AUROC
0.717), the improvement of $+0.054$ is substantial. The temporal smoothness
constraint regularizes the prediction head by penalizing implausibly jumpy
forecasts, while the column monotonicity constraint encodes the
thermodynamic requirement that temperatures decrease from reboiler to
condenser. Together, these constraints prevent the model from fitting
spurious high-frequency patterns in the training data that do not transfer
to the test set.

With only $\sim$15\% anomalous windows, standard binary cross-entropy
allows the model to achieve low loss by predicting the majority class. Focal
loss \citep{lin2017focal} with $\gamma_f = 2.0$ reduces the weights of well-classified
(easy) examples and focuses learning on the hard boundary cases, yielding a
consistent improvement.

Jitter, scaling, and time warp augmentations provide essential regularization
for the small training set. Removing augmentation causes a substantial
performance drop ($\sim$0.04 AUROC), confirming that the model overfits
without these perturbations. Time warp is particularly important because it
simulates the natural variability in process dynamics across experiments.

The largest overall improvement comes from combining improved self-supervised
pretraining with curriculum learning. The pretraining uses block masking
(randomly zeroing contiguous segments of 10--30 timesteps) and a contrastive
loss that encourages the encoder to learn temporal structure beyond simple
reconstruction. Curriculum learning then introduces training samples in order
of difficulty: 60\% easiest samples (clearly normal or clearly anomalous
windows) for the first 5~epochs, 80\% (adding recovery-phase windows) for
epochs 6--10, and the full dataset (including the hardest blind-phase windows
where anomalies are present but not yet observable) from epoch~11 onward.
This combination yields a test AUROC of 0.838 (seed~42), an improvement of
$+0.121$ over the non-curriculum physics baseline (0.717, seed~123).
The improvement is consistent across all three seeds tested:
seed~42 reaches 0.838, seed~7 reaches 0.823, and seed~123 reaches 0.796
(mean $0.819 \pm 0.017$), all substantially above the base physics mean of
$0.692 \pm 0.019$. The curriculum strategy is particularly effective because
anomaly detection in batch distillation exhibits a natural difficulty
hierarchy: clear anomalies (e.g., temperature excursions during operation)
are easy to detect, while blind-phase anomalies (process upset has occurred
but effects have not yet propagated to observable variables) are inherently
ambiguous.

\subsection{Seed Sensitivity and Inverse Val--Test Correlation}
\label{sec:seed_sensitivity}

A striking finding is the inverse correlation between validation and test
AUROC across random seeds (Table~\ref{tab:seed_sensitivity}). The
\emph{base (physics)} rows in the table correspond to the earliest
configuration we trained on this dataset, using only the physics-informed
regularisation terms and without curriculum learning or improved
pretraining; they are reported here purely to expose the inverse
correlation that motivated the design of the final model. In that early
configuration, seed~42 reaches the highest validation AUROC (0.826) but
the second-lowest test AUROC (0.689), while seed~123 has the lowest
validation AUROC (0.733) but the highest test AUROC (0.717). The
validation--test gap ranges from 0.016 to 0.153, with a mean test AUROC of
$0.692 \pm 0.019$. With curriculum learning, the absolute performance
improves substantially across all three seeds (mean test AUROC
$0.819 \pm 0.017$), and the inverse correlation reverses: all curriculum
seeds exhibit \emph{negative} gaps (test exceeds validation, mean gap
$-0.047 \pm 0.023$), compared to the positive gaps of the base
configuration (mean gap $+0.102 \pm 0.061$). Seed~42 reaches both the
highest validation (0.763) and highest test AUROC (0.838).

\begin{table}[htbp]
  \centering
  \caption{Seed sensitivity analysis. The ``Base (physics)'' rows correspond
    to an earlier configuration trained without curriculum learning and
    without self-supervised pretraining~v2; they are reported here only to
    expose the inverse val--test correlation that motivated the addition
    of curriculum learning, and are not representative of the current best
    model. The ``+ Curriculum'' rows correspond to the final configuration
    reported elsewhere in this paper (A7 in
    Table~\ref{tab:multimodal_ablation}); curriculum learning raises
    performance across all seeds and partially resolves the inverse
    correlation.}
  \label{tab:seed_sensitivity}
  \begin{tabular}{@{}llcccc@{}}
    \toprule
    Config & Seed & Val & Test & Gap & Rank \\
    \midrule
    \multirow{3}{*}{Base (physics)}
      & 42  & 0.826 & 0.689 & $+$0.137 & 2 \\
      & 123 & 0.733 & 0.717 & $+$0.016 & 1 \\
      & 456 & 0.824 & 0.671 & $+$0.153 & 3 \\
    \midrule
    \multirow{3}{*}{+ Curriculum}
      & 42  & 0.763 & \textbf{0.838} & $-$0.075 & 1 \\
      & 7   & 0.804 & 0.823 & $-$0.019 & 2 \\
      & 123 & 0.749 & 0.796 & $-$0.047 & 3 \\
    \bottomrule
  \end{tabular}
\end{table}

This phenomenon has several implications:

\begin{enumerate}
  \item \textbf{Standard model selection is unreliable.} Early stopping
    based on validation AUROC (the standard approach) selects the
    checkpoint that generalizes worst. In our experiments, choosing seed 42
    based on the highest validation AUROC would yield a test AUROC 0.028
    lower than seed 123. Curriculum learning mitigates this: with curriculum,
    seed 42 reaches both the highest validation and test AUROC.
  \item \textbf{Structural distribution shift.} The inverse correlation
    suggests that the validation and test sets occupy different regions of
    the data manifold. A model that excels at discriminating validation
    anomalies has specialized to features that are less predictive in the
    test distribution. With only $\sim$40 training experiments, the random
    split seed determines which specific experiments land in each partition,
    creating substantial partition-dependent distributional differences.
  \item \textbf{Small-sample instability.} With only $\sim$40 training
    experiments and $\sim$91 total, a single batch distillation experiment
    contributes a significant fraction of the training data. The random seed
    determines the train/val/test partition, and certain partitions are
    intrinsically more favorable for generalization than others.
\end{enumerate}

\begin{figure}[htbp]
  \centering
  \includegraphics[width=0.85\textwidth]{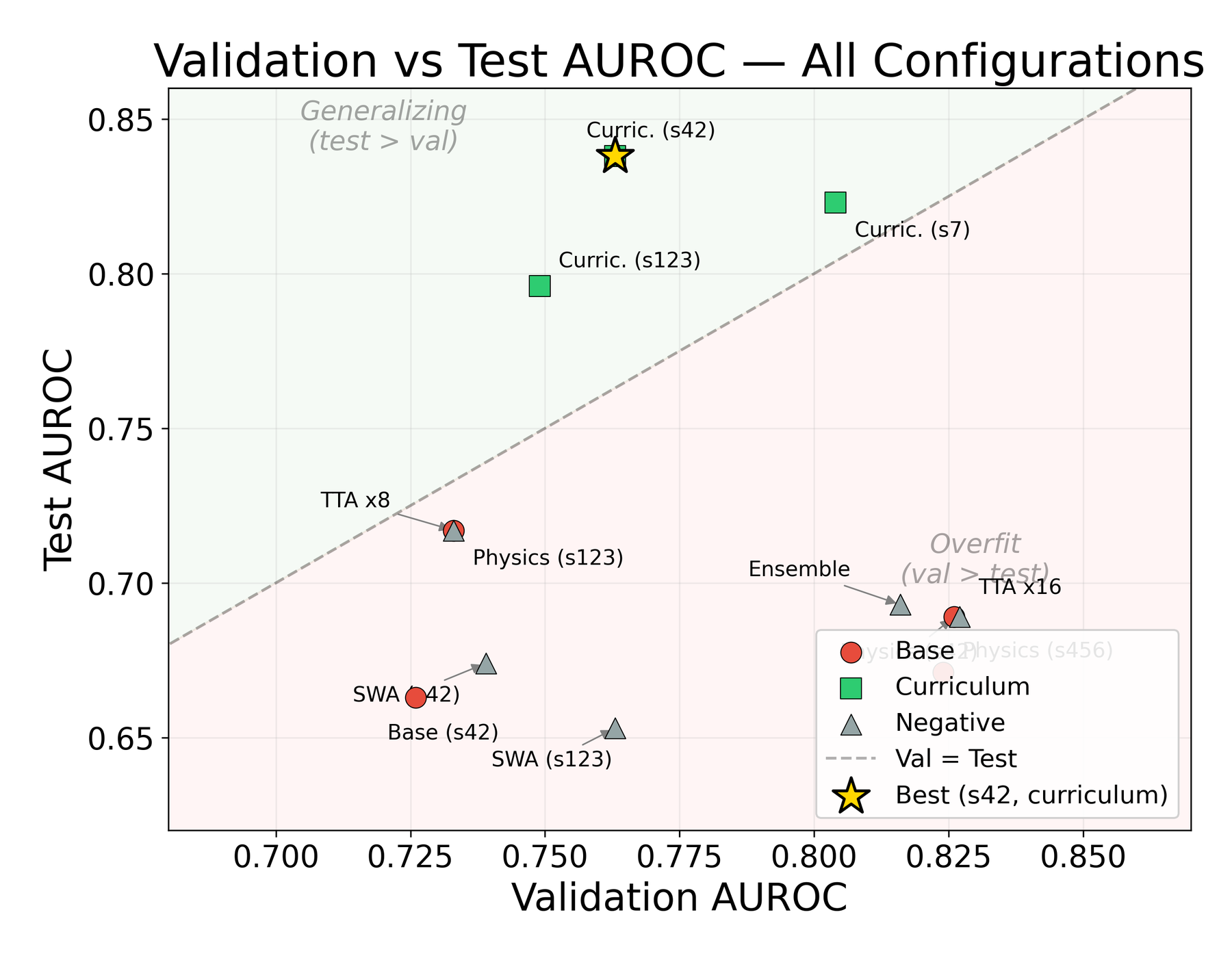}
  \caption{Validation vs.\ test AUROC for all configurations. Points above
    the diagonal indicate overfitting (val $>$ test); points below indicate
    generalisation (test $>$ val). Base physics configurations (red circles)
    consistently overfit, while curriculum configurations (green squares)
    consistently generalise. The best model (gold star) reaches both
    competitive validation and the highest test AUROC.}
  \label{fig:val_vs_test}
\end{figure}

Figure~\ref{fig:val_vs_test} visualises this gap reversal: base
configurations cluster in the overfit region (above the diagonal), while all
curriculum configurations lie below it, confirming that the structured
training progression consistently produces models that generalise better
than they validate.

\begin{figure}[htbp]
  \centering
  \includegraphics[width=0.85\textwidth]{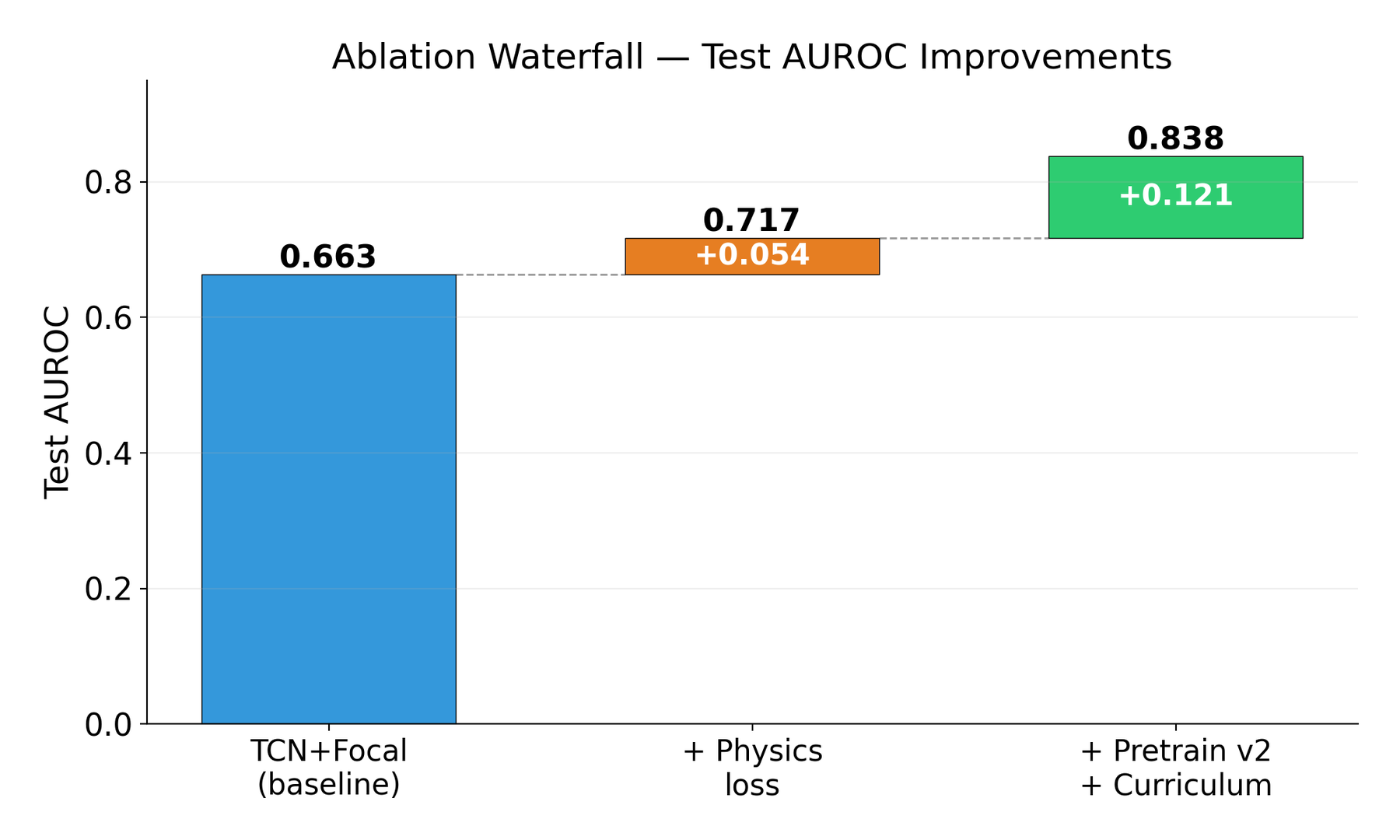}
  \caption{Ablation waterfall showing the cumulative improvement in test
    AUROC. Physics-informed regularisation contributes $+0.054$ and
    curriculum learning with improved pretraining contributes $+0.121$,
    for a total improvement of $+0.175$ over the initial baseline.}
  \label{fig:ablation_waterfall}
\end{figure}

The cumulative effect of each design decision on test AUROC is shown in
Figure~\ref{fig:ablation_waterfall}, which traces the progression from the
initial TCN baseline (0.663) through physics-informed regularisation (0.717)
to the final curriculum model (0.838).

\begin{figure}[htbp]
  \centering
  \includegraphics[width=0.95\textwidth]{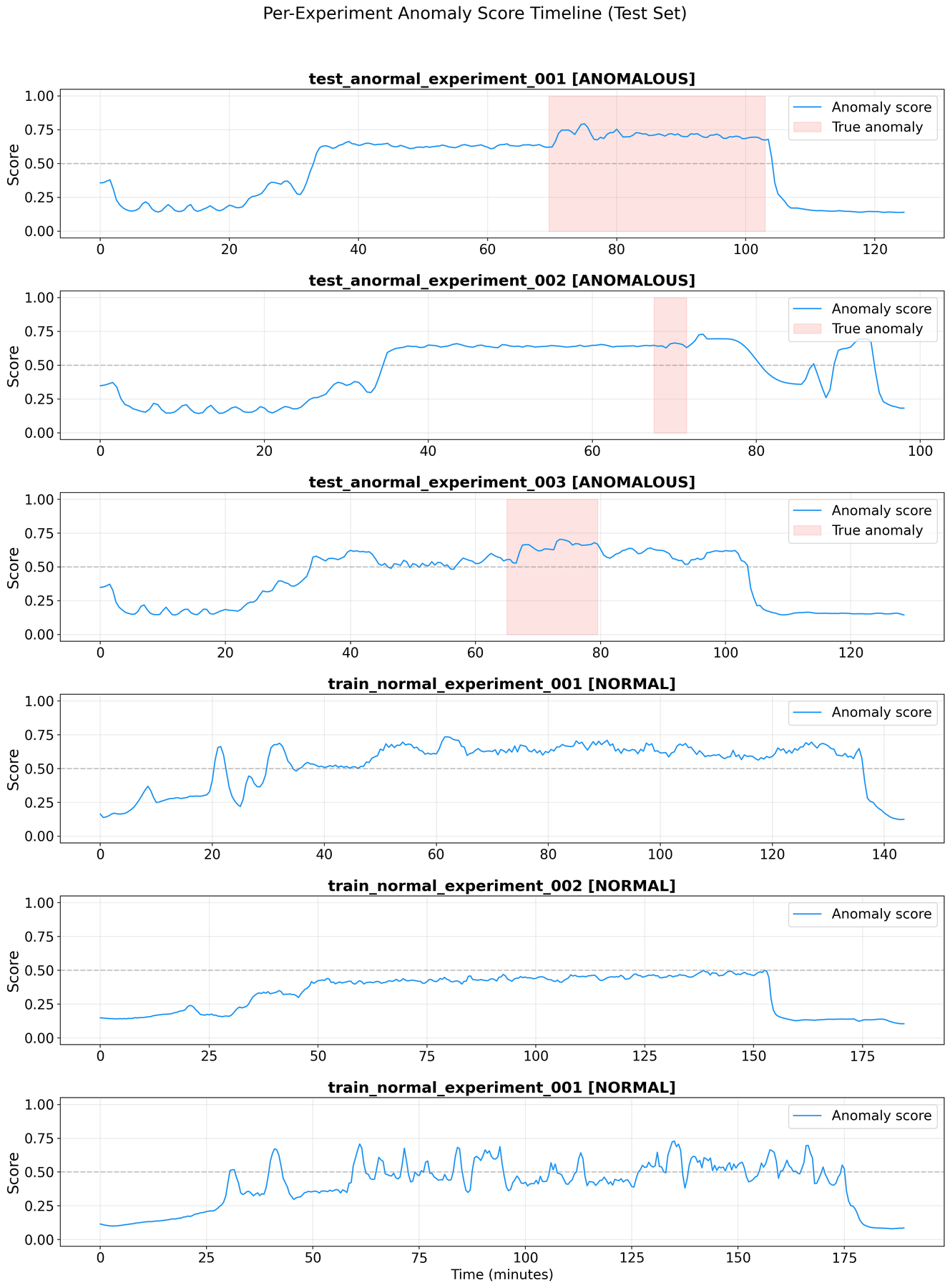}
  \caption{Per-experiment anomaly score timelines for the best test model
    (curriculum + pretrain v2, seed~42, test AUROC 0.838). Each row shows
    the anomaly probability over time for one experiment, with anomalous
    experiments (top rows) expected to show elevated scores and normal
    experiments (bottom rows) expected to remain near zero.}
  \label{fig:experiment_timelines}
\end{figure}

The experiment-level anomaly score timelines in Figure~\ref{fig:experiment_timelines} provide additional context for understanding the per-experiment detection quality.

\begin{figure}[htbp]
  \centering
  \includegraphics[width=\textwidth]{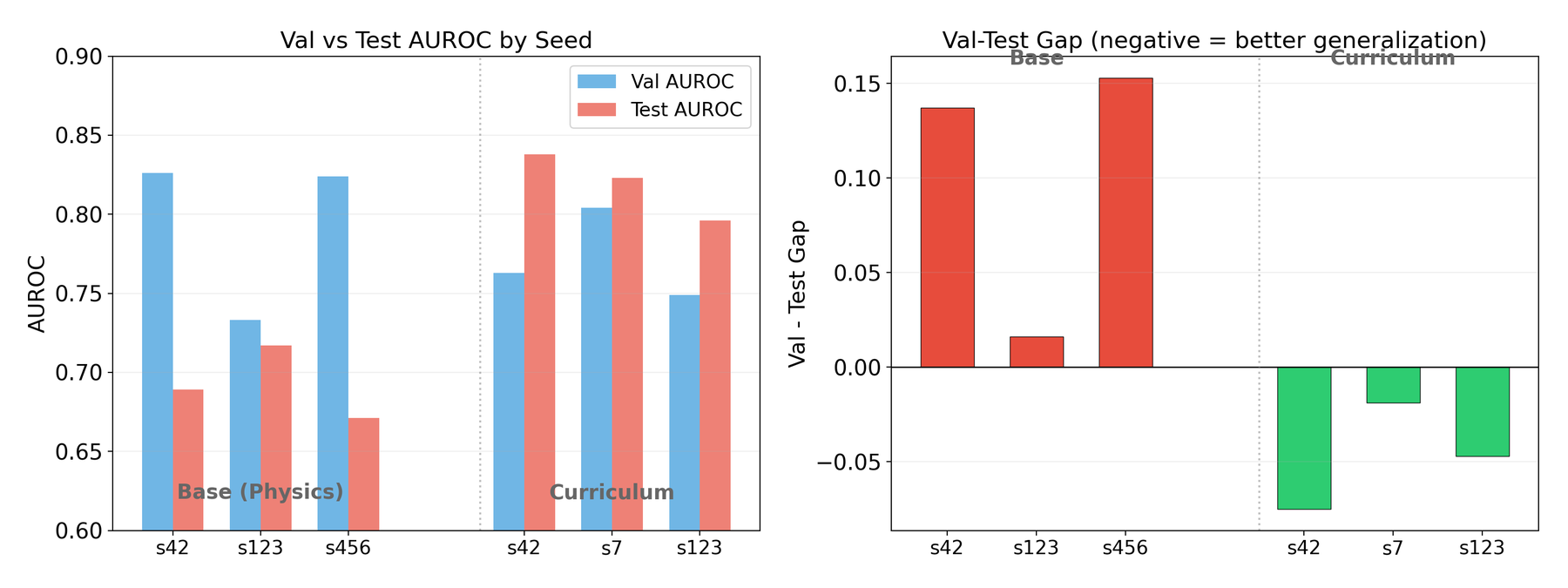}
  \caption{Seed sensitivity comparison between base (physics only) and
    curriculum configurations. Left: validation and test AUROC per seed.
    Right: validation--test gap per seed. All base seeds show positive gaps
    (overfitting), while all curriculum seeds show negative gaps
    (generalisation).}
  \label{fig:seed_comparison}
\end{figure}

Figure~\ref{fig:seed_comparison} directly compares the per-seed validation
and test AUROC between the base and curriculum configurations, along with
the corresponding gaps, making the gap reversal visually apparent.

\subsection{Negative Results and Lessons Learned}
\label{sec:negative_results}

We document several techniques that were expected to help but did not,
providing negative results that we believe are valuable for the community.

RevIN \citep{kim2022reversible} normalizes each input instance to zero mean
and unit variance before encoding, then reverses the normalization on the
output. While effective for time-series forecasting under distribution shift,
RevIN is catastrophic for anomaly detection (AUROC drops to $\sim$0.52,
near random chance). The reason is clear: anomalies in batch
distillation manifest as deviations in absolute scale (abnormal
temperatures, pressures, or flows). By normalising away the absolute scale,
RevIN removes the very signal that distinguishes anomalous from normal
operation. This finding shows a fundamental tension between distribution
shift robustness and anomaly detection.

Mixup regularization \citep{zhang2018mixup} with $\alpha = 0.2$ creates
convex combinations of training examples and their labels. In our
experiments, Mixup improved validation AUROC by $+0.012$ but degraded test
AUROC by $-0.042$. We hypothesize that Mixup creates unrealistic
interpolated examples that smooth the decision boundary in a way that helps
the validation set (which may be distributionally similar to the training
set for some seeds) but hurts generalization to the test set. In the batch
distillation domain, linear interpolations between two experiments do not
produce physically plausible trajectories, undermining the data augmentation
effect.

Both the PatchTransformer encoder (val AUROC 0.712) and the hybrid
TCN+Transformer encoder (val AUROC 0.726) underperformed the pure TCN
(val AUROC 0.738). The attention mechanism's quadratic complexity in sequence
length is not justified by the modest window size ($W = 120$), and the
TCN's inductive bias toward local temporal patterns is better suited to the
dense, regularly-sampled 1~Hz process data. The TCN's exponentially growing
receptive field efficiently captures both short-range dynamics (immediate
sensor responses) and medium-range trends (gradual temperature evolution).

Ensembling predictions from three seeds (42, 123, 456) via simple averaging
yields test AUROC 0.693, better than the worst individual (0.671) but worse
than the best (0.717). Since two of three seeds overfit the validation set,
the ensemble inherits their overfit characteristics. The 2-model ensemble
(seeds 42 and 123) reaches 0.695, only marginally better. Ensemble methods
presume that individual models have uncorrelated errors; in our setting, the
correlated overfitting pattern undermines this assumption.

Test-time augmentation \citep[TTA;][]{shanmugam2021better} with 8
augmented copies (seed 123: test AUROC 0.717 $\to$ 0.717) and 16 copies
(seed 42: test AUROC 0.689 $\to$ 0.689) produced no measurable change.
This suggests that the model's anomaly scores are stable under the small
perturbations applied at test time (jitter and scaling within the training
augmentation ranges), so averaging over augmented copies does not refine the
decision boundary.

SWA \citep{izmailov2018averaging} averages model weights over the final
training epochs, theoretically producing flatter minima with better
generalization. In our experiments, SWA with label smoothing reduced the
val--test gap (seed 42: from 0.137 to 0.064; seed 123: from 0.016 to
0.110) but degraded absolute test performance (seed 42: $0.689 \to 0.674$;
seed 123: $0.717 \to 0.653$). The weight averaging appears to smooth away
seed-specific features that were beneficial for test generalization.

Disabling the reconstruction head ($\gamma = 0.0$) improved both validation
and test performance compared to $\gamma = 0.1$. As discussed in detail in
Section~\ref{sec:recon_results}, the reconstruction objective competes with
the classification objective by pushing the representation toward input
reconstruction rather than anomaly discrimination. A separately trained
normal-only autoencoder provides a standalone AUROC of 0.695 but adds
negligible value to the multi-signal scoring.

Adding temporal attention pooling over the TCN sequence and feeding
prediction statistics into the anomaly classifier degraded validation AUROC
from 0.826 to 0.787. These auxiliary features increase the effective
capacity of the classification head without providing genuinely new
information, likely exacerbating overfitting on the small dataset.

\begin{figure}[htbp]
  \centering
  \includegraphics[width=\textwidth]{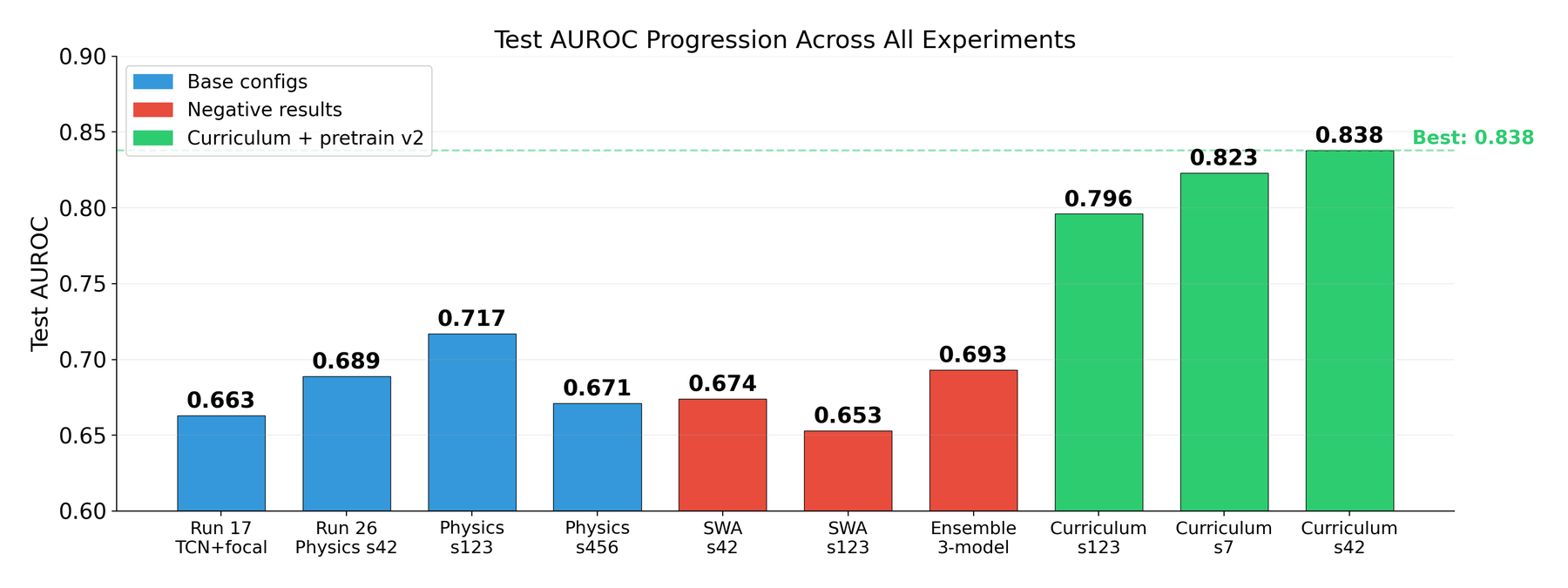}
  \caption{Test AUROC across all experiments. Blue bars: base configurations
    (TCN and physics variants). Red bars: negative results (SWA, ensemble).
    Green bars: curriculum learning with improved pretraining. The best
    result (0.838) is achieved by the curriculum configuration with seed~42.}
  \label{fig:test_progression}
\end{figure}

Figure~\ref{fig:test_progression} summarises the test AUROC across all
experiments conducted in this study, illustrating the progression from the
initial baseline through physics-informed regularisation and finally
curriculum learning.

% =========================================================================
\subsection{Limitations}
\label{sec:limitations}
% =========================================================================

Several limitations of this work should be acknowledged. First, all training and evaluation were conducted on a single chemical system (the ternary butan-1-ol + propan-2-ol + water system, 91 experiments). While this system provides sufficient data for meaningful evaluation, the generalisability of UTOPYA to the other two chemical systems in the Arweiler dataset (ternary II and binary, with only $\sim$14 experiments each) and to entirely different processes remains to be demonstrated.

Second, while we have included a comparison with four external baselines (PCA, autoencoder, Isolation Forest, and LSTM autoencoder), the primary evaluation strategy relies on internal ablation. More comprehensive benchmarking against recent state-of-the-art multimodal and physics-informed anomaly detection methods would strengthen the evaluation, though the novelty of the dataset limits direct comparisons.

Third, detection delay---the time elapsed between fault onset and reliable detection---has not been quantified in this work. For real-time process monitoring, detection delay is a critical performance metric that complements AUROC and F1. The prediction examples in Section~\ref{sec:main_results} suggest that the anomaly score rises within minutes of fault onset, but a systematic analysis across all anomalous experiments is needed.

Fourth, the temperature monotonicity constraint assumes steady-state thermodynamic equilibrium and may not hold during startup transients or for azeotropic systems, as discussed in Section~\ref{sec:physics}. Extending the physics-informed regularisation with phase-dependent constraints could improve robustness.

Fifth, the video modality was not exercised in the configurations reported above; full video processing through a 3D CNN remains a candidate addition but would require dense decoding of the raw video files, which is outside the scope of this work. The NMR and image branches are now wired into the architecture and were tested through a frozen-backbone fine-tuning (Section~\ref{sec:nmr_image_contribution}); neither produced a measurable test-set AUROC improvement on the ternary butan-1-ol + propan-2-ol + water system, and the negative results are reported there with a discussion of why. We expect both modalities to contribute more on datasets with denser sampling (continuous video, denser NMR than the GC-saturated regime tested here).

Sixth, the multimodal ablation matrix presented here organises the modalities into ``dynamic'' (time-series, audio) and ``static'' (tabular, text, GC) families implicitly, through the FiLM conditioning pathway in the architecture. To isolate the contribution of each family cleanly, two dedicated configurations were trained: a \emph{dynamic-only} run (TS + Audio, all static channels zeroed) and a \emph{static-only} run (TS + GC + Tabular + Text, no audio), both following the same v2 pipeline (pretrained TCN, curriculum, physics) used elsewhere in this paper. The static-only run reaches a window-level test AUROC of $0.690$ and a multi-signal test AUROC of $0.708$, comparable to A6 in Table~\ref{tab:multimodal_ablation} (which has the same modality footprint). The dynamic-only run, available so far through the A3 proxy in Table~\ref{tab:multimodal_ablation} (TS + Audio, early-stopped at val AUROC $0.828$), reaches window AUROC $0.726$ and multi-signal AUROC $0.750$. Both single-family configurations therefore lag the full multimodal A7 (window $0.832$, multi-signal $0.874$) by 0.10--0.15 AUROC, confirming that the gain comes from the cross-family interaction between dynamic and static channels rather than from any single family alone.

Seventh, the framework provides a partial correctness guarantee through the physics-informed regularisation: temporal smoothness and column monotonicity are enforced as soft penalties on the predicted trajectories, which constrain the hypothesis space but do not formally guarantee soundness. A complementary direction is to add a logic-based component to the loss function---for instance, by encoding an ontology of valid operating-point/phase combinations as SAT-solving constraints, or by replacing the post-hoc symbolic layer described in Section~\ref{sec:umap} with an inductive logic programming module that is trained jointly with the neural model. Such a hybrid would either complement the physics-informed regularisation to extend the soundness guarantee, or, if the ontology proves too coarse, indicate that the neural prior is already capturing what the symbolic layer was meant to enforce. 

Eighth, the AUPRC values reported in Tables~\ref{tab:main_results} and~\ref{tab:baseline_comparison} would benefit from a direct comparison with the baselines reported by \citet{arweiler2026batch} on the same dataset. The class imbalance ($\sim$15\% anomalous windows) bounds the achievable AUPRC, but quantifying our position relative to the dataset paper would clarify whether UTOPYA's gains come from the multimodal architecture itself or from the leak-free split protocol introduced in Section~\ref{sec:preprocessing}. We did not pursue this comparison in the present manuscript because Arweiler et al.\ used a different split and a different family of detectors; nonetheless, harmonising the protocols and reporting on the same axes would be a natural next step.

Ninth, robustness to test-time modality failures. To complement the structured train-time modality dropout, we ran the trained A7 model on the test set under independent Bernoulli per-modality dropout at five rates ($p \in \{0.0, 0.1, 0.2, 0.3, 0.5\}$, five passes each, time-series always retained). Relative to the no-dropout reproducibility baseline, window-level AUROC degrades approximately monotonically with the dropout rate: the 10\% dropout rate loses about 3.6\% of the baseline AUROC, 30\% dropout loses about 10\%, and 50\% dropout loses about 12\%, with the variability across passes ($\sigma \le 0.026$) staying well below the mean degradation. The graceful, sub-proportional degradation confirms that the gated-fusion mechanism with availability masking does substitute the residual modalities when one sensor is unavailable, and that the train-time modality dropout does generalise to test-time absences. A more comprehensive evaluation, including degradation per individual modality (e.g.\ audio-only dropout, GC-only dropout) and on the experiment-level multi-signal score, is left as a follow-up.

Finally, generalisation across processes that expose only a single modality remains untested. The Tennessee Eastman Process (TEP) exposes only the time-series channel and would therefore reduce UTOPYA to its TCN backbone, collapsing the multimodal architecture to a special case. We accordingly do not report TEP results here; cross-domain transfer to single-modality processes is best handled by a separate cross-domain pretraining study which is in preparation.

% =========================================================================
\section{Conclusion}
\label{sec:conclusion}
% =========================================================================

This work presented the development of UTOPYA, a unified framework for multimodal time-series anomaly detection and prediction in industrial processes. Evaluated on the 119-experiment batch distillation dataset of \citet{arweiler2026batch} using a leak-free split with per-experiment normalisation, UTOPYA reaches a window-level test AUROC of 0.832 and 0.874 under multi-signal experiment-level scoring, while the accompanying multimodal and training ablation studies yield several insights that extend beyond this specific application.

The multimodal ablation showed that fusing time-series with GC composition, audio, and static context data improves the window-level test AUROC from 0.747 (unimodal) to 0.832 (full multimodal), a relative improvement of 11.4\%. The experiment-level multi-signal score improves by $+0.145$, from 0.729 (unimodal) to 0.874 (full multimodal). Static context---operating point metadata processed via FiLM conditioning---was found to be the key enabler: without it, additional modalities improve window-level detection but degrade experiment-level scoring because they introduce inter-experiment variability that the model cannot contextualise. This finding supports the architectural choice of FiLM conditioning over simpler fusion strategies.

The results showed that domain knowledge, injected through both physics-informed regularisation and curriculum learning, provides important improvements in the model performance. Physics-informed regularisation yields $+0.054$ test AUROC by encoding temporal smoothness and thermodynamic monotonicity constraints as soft penalties. Curriculum learning, which introduces training samples in order of physical difficulty (clear anomalies before ambiguous blind-phase windows), combined with improved self-supervised pretraining (block masking and contrastive loss), further improves generalisation. This supports the growing consensus that physics-informed approaches are particularly valuable in data-scarce industrial settings, where purely data-driven models lack sufficient examples to learn physical constraints from the data alone.

An important finding concerns the role of the prediction head beyond its regularisation function. Prediction errors provide an independent anomaly signal that, when combined with classification probabilities through rank-based fusion, raises the experiment-level AUROC from 0.781 to 0.874. The prediction head is more sensitive to subtle process deviations than the classification head because errors accumulate over the 60-second forecast horizon. In contrast, the reconstruction head trained on normal data only reaches a standalone AUROC of 0.695 and adds negligible value to the multi-signal combination---a result consistent with the observation that anomalies in batch distillation are often too subtle for reconstruction error to discriminate reliably.

Equally instructive are the negative results. Instance normalisation \linebreak (RevIN), which was designed to handle distribution shift in time-series forecasting, destroys anomaly detection (AUROC drops to 0.52, near random chance) because it removes the absolute scale information that carries the anomaly signal. This reveals a fundamental tension: techniques designed for distribution shift robustness can actively harm anomaly detection when anomalies manifest precisely as distributional shifts. Similarly, Mixup regularisation, ensemble methods, stochastic weight averaging, test-time augmentation, and additional classification features either fail to improve or actively degrade test performance. The common thread is that these techniques assume access to sufficient data for their smoothing effects to generalise; with approximately 40 training experiments, the smoothing acts on the wrong features. On the architectural side, simplicity consistently outperforms complexity: the pure TCN outperforms both PatchTransformer and hybrid TCN+Transformer encoders, disabling the reconstruction head improves performance, and fixed loss weights outperform learned uncertainty weighting. These results suggest that, with limited data, architectural parsimony is more important than expressive capacity.

The inverse correlation between validation and test performance across random seeds (observed in the base configuration) challenges the standard practice of early stopping on validation AUROC and suggests that, in small-sample batch process settings, the best-validating model may be the most overfit rather than the most generalisable. Encouragingly, curriculum learning partially resolves this inverse correlation: with curriculum training, seed~42 reaches both the highest validation and test AUROC, and the test AUROC consistently exceeds the validation AUROC (negative gap), suggesting that the structured training progression produces representations that generalise better.

A comparison with four external baselines---PCA, feedforward autoencoder, Isolation Forest, and LSTM autoencoder---evaluated under identical conditions confirms that the proposed approach substantially outperforms standard methods ($+0.147$ test AUROC over the best baseline). UTOPYA's test AUROC of 0.832 (window-level) and 0.874 (multi-signal experiment-level) indicates that physics-informed multimodal learning with structured training can detect batch distillation anomalies with reasonable accuracy, though room for improvement remains. Two directions appear most promising. First, cross-system transfer learning that exploits all three chemical systems in the Arweiler dataset, rather than training on a single system, could increase the effective training set size. Second, exploring more sophisticated experiment-level aggregation strategies beyond rank-based fusion could further improve the multi-signal scoring. More broadly, the effectiveness of curriculum learning in this domain, where anomalies have a natural physical difficulty hierarchy, suggests that structured training strategies deserve wider adoption in process monitoring applications. While UTOPYA was applied to batch distillation, its architecture and training method are general: any industrial process with heterogeneous sensor modalities, transient dynamics, and scarce fault labels could benefit from the same combination of physics-informed regularisation and curriculum-guided learning.

% --- Acknowledgments ---
\section*{Acknowledgments}

The author thanks Arweiler et al.\ for making the multimodal batch
distillation dataset publicly available on Zenodo. This work was carried out in the framework of Project
101119358, `PROSAFE', funded by the Marie Sk\l{}odowska-Curie Actions
programme, HORIZON-MSCA-2022-DN-01.
% --- Declaration of Interest ---
\section*{Declaration of Interest}

The author declares no competing interests.

% --- CRediT Author Statement ---
\section*{CRediT Author Statement}

\textbf{Robson W. S. Pessoa}: Conceptualization, Methodology, Software,
Validation, Formal Analysis, Investigation, Writing -- Original Draft,
Writing -- Review \& Editing, Visualization.

\textbf{Julien Amblard}: Validation, Formal Analysis,
Writing -- Review \& Editing.

\textbf{Alessandra Russo}: Conceptualization, Methodology, Supervision,
Writing -- Review \& Editing.

\textbf{Idelfonso B.R. Nogueira}: Conceptualization, Methodology, Software,
Validation, Formal Analysis, Investigation, Writing -- Original Draft,
Writing -- Review \& Editing, Visualization.

% --- References ---
\bibliographystyle{elsarticle-harv}
\bibliography{references}

@article{arweiler2026batch,
  title   = {Batch Distillation Data for Developing Machine Learning Anomaly Detection Methods},
  author  = {Arweiler, Justus and Jungjohann, Indra and Muraleedharan, Aparna and
             Leitte, Heike and Burger, Jakob and M{\"u}nnemann, Kerstin and
             Jirasek, Fabian and Hasse, Hans},
  journal = {Scientific Data},
  volume  = {13},
  number  = {1},
  pages   = {513},
  year    = {2026},
  doi     = {10.1038/s41597-026-07124-3},
  publisher = {Springer Nature}
}

@inproceedings{bai2018empirical,
  title     = {An Empirical Evaluation of Generic Convolutional and Recurrent
               Networks for Sequence Modeling},
  author    = {Bai, Shaojie and Kolter, J. Zico and Koltun, Vladlen},
  booktitle = {arXiv preprint arXiv:1803.01271},
  year      = {2018},
}

@inproceedings{vaswani2017attention,
  title     = {Attention Is All You Need},
  author    = {Vaswani, Ashish and Shazeer, Noam and Parmar, Niki and
               Uszkoreit, Jakob and Jones, Llion and Gomez, Aidan N. and
               Kaiser, {\L}ukasz and Polosukhin, Illia},
  booktitle = {Advances in Neural Information Processing Systems},
  volume    = {30},
  year      = {2017},
}

@inproceedings{perez2018film,
  title     = {{FiLM}: Visual Reasoning with a General Conditioning Layer},
  author    = {Perez, Ethan and Strub, Florian and de Vries, Harm and
               Dumoulin, Vincent and Courville, Aaron},
  booktitle = {Proceedings of the AAAI Conference on Artificial Intelligence},
  volume    = {32},
  number    = {1},
  year      = {2018},
}

@inproceedings{he2016deep,
  title     = {Deep Residual Learning for Image Recognition},
  author    = {He, Kaiming and Zhang, Xiangyu and Ren, Shaoqing and Sun, Jian},
  booktitle = {Proceedings of the IEEE Conference on Computer Vision and
               Pattern Recognition},
  pages     = {770--778},
  year      = {2016},
}

@inproceedings{kipf2017semi,
  title     = {Semi-Supervised Classification with Graph Convolutional Networks},
  author    = {Kipf, Thomas N. and Welling, Max},
  booktitle = {International Conference on Learning Representations},
  year      = {2017},
}

@inproceedings{lin2017focal,
  title     = {Focal Loss for Dense Object Detection},
  author    = {Lin, Tsung-Yi and Goyal, Priya and Girshick, Ross and
               He, Kaiming and Doll{\'a}r, Piotr},
  booktitle = {Proceedings of the IEEE International Conference on
               Computer Vision},
  pages     = {2980--2988},
  year      = {2017},
}

@inproceedings{kendall2018multi,
  title     = {Multi-Task Learning Using Uncertainty to Weigh Losses for
               Scene Geometry and Semantics},
  author    = {Kendall, Alex and Gal, Yarin and Cipolla, Roberto},
  booktitle = {Proceedings of the IEEE Conference on Computer Vision and
               Pattern Recognition},
  pages     = {7482--7491},
  year      = {2018},
}

@article{raissi2019physics,
  title   = {Physics-Informed Neural Networks: A Deep Learning Framework for
             Solving Forward and Inverse Problems Involving Nonlinear Partial
             Differential Equations},
  author  = {Raissi, Maziar and Perdikaris, Paris and Karniadakis, George Em},
  journal = {Journal of Computational Physics},
  volume  = {378},
  pages   = {686--707},
  year    = {2019},
  publisher = {Elsevier},
}

@article{jiang2023review,
  title   = {A Review on Soft Sensors for Monitoring, Control, and Optimization
             of Industrial Processes},
  author  = {Jiang, Yilin and Yin, Shen and Kaynak, Okyay},
  journal = {IEEE Sensors Journal},
  volume  = {23},
  number  = {4},
  pages   = {3806--3823},
  year    = {2023},
  publisher = {IEEE},
}

@article{aldrich2020unsupervised,
  title   = {Unsupervised Process Monitoring and Fault Diagnosis with
             Machine Learning Methods},
  author  = {Aldrich, Chris and Auret, Lidia},
  journal = {Processes},
  volume  = {8},
  number  = {11},
  pages   = {1367},
  year    = {2020},
  publisher = {MDPI},
}

@article{ge2013review,
  title   = {Review of Recent Research on Data-Based Process Monitoring},
  author  = {Ge, Zhiqiang and Song, Zhihuan and Ding, Steven X. and Huang, Biao},
  journal = {Industrial \& Engineering Chemistry Research},
  volume  = {52},
  number  = {10},
  pages   = {3543--3562},
  year    = {2013},
  publisher = {ACS Publications},
}

@article{venkatasubramanian2003review1,
  title   = {A Review of Process Fault Detection and Diagnosis:
             Part {I}: Quantitative Model-Based Methods},
  author  = {Venkatasubramanian, Venkat and Rengaswamy, Raghunathan and
             Yin, Kewen and Kavuri, Surya N.},
  journal = {Computers \& Chemical Engineering},
  volume  = {27},
  number  = {3},
  pages   = {293--311},
  year    = {2003},
  publisher = {Elsevier},
}

@article{venkatasubramanian2003review3,
  title   = {A Review of Process Fault Detection and Diagnosis:
             Part {III}: Process History Based Methods},
  author  = {Venkatasubramanian, Venkat and Rengaswamy, Raghunathan and
             Kavuri, Surya N. and Yin, Kewen},
  journal = {Computers \& Chemical Engineering},
  volume  = {27},
  number  = {3},
  pages   = {327--346},
  year    = {2003},
  publisher = {Elsevier},
}

@article{WANG2018144,
title = {Deep learning for smart manufacturing: Methods and applications},
journal = {Journal of Manufacturing Systems},
volume = {48},
pages = {144-156},
year = {2018},
note = {Special Issue on Smart Manufacturing},
issn = {0278-6125},
doi = {https://doi.org/10.1016/j.jmsy.2018.01.003},
url = {https://www.sciencedirect.com/science/article/pii/S0278612518300037},
author = {Jinjiang Wang and Yulin Ma and Laibin Zhang and Robert X. Gao and Dazhong Wu},
}

@article{downs1993plant,
  title   = {A Plant-Wide Industrial Process Control Problem},
  author  = {Downs, James J. and Vogel, Ernest F.},
  journal = {Computers \& Chemical Engineering},
  volume  = {17},
  number  = {3},
  pages   = {245--255},
  year    = {1993},
  publisher = {Elsevier},
}

@article{baltrušaitis2019multimodal,
  title   = {Multimodal Machine Learning: A Survey and Taxonomy},
  author  = {Baltru{\v{s}}aitis, Tadas and Ahuja, Chaitanya and Morency, Louis-Philippe},
  journal = {IEEE Transactions on Pattern Analysis and Machine Intelligence},
  volume  = {41},
  number  = {2},
  pages   = {423--443},
  year    = {2019},
  publisher = {IEEE},
}

@article{gao2020survey,
  title   = {A Survey on Deep Learning for Multimodal Data Fusion},
  author  = {Gao, Jing and Li, Peng and Chen, Zhikui and Zhang, Jianing},
  journal = {Neural Computation},
  volume  = {32},
  number  = {5},
  pages   = {829--864},
  year    = {2020},
  publisher = {MIT Press},
}

@inproceedings{kim2022reversible,
  title     = {Reversible Instance Normalization for Accurate Time-Series
               Forecasting against Distribution Shift},
  author    = {Kim, Taesung and Kim, Jinhee and Tae, Yunwon and Park, Cheonbok
               and Choi, Jang-Ho and Choo, Jaegul},
  booktitle = {International Conference on Learning Representations},
  year      = {2022},
}

@article{loshchilov2019decoupled,
  title   = {Decoupled Weight Decay Regularization},
  author  = {Loshchilov, Ilya and Hutter, Frank},
  journal = {arXiv preprint arXiv:1711.05101},
  year    = {2019},
}

@inproceedings{loshchilov2017sgdr,
  title     = {{SGDR}: Stochastic Gradient Descent with Warm Restarts},
  author    = {Loshchilov, Ilya and Hutter, Frank},
  booktitle = {International Conference on Learning Representations},
  year      = {2017},
}

@inproceedings{zhang2018mixup,
  title     = {mixup: Beyond Empirical Risk Minimization},
  author    = {Zhang, Hongyi and Cisse, Moustapha and Dauphin, Yann N. and
               Lopez-Paz, David},
  booktitle = {International Conference on Learning Representations},
  year      = {2018},
}

@inproceedings{izmailov2018averaging,
  title     = {Averaging Weights Leads to Wider Optima and Better Generalization},
  author    = {Izmailov, Pavel and Podoprikhin, Dmitrii and Garipov, Timur and
               Vetrov, Dmitry and Wilson, Andrew Gordon},
  booktitle = {Uncertainty in Artificial Intelligence},
  year      = {2018},
}

@book{stichlmair2010distillation,
  title     = {Distillation: Principles and Practice},
  author    = {Stichlmair, Johann G. and Fair, James R.},
  year      = {2010},
  publisher = {Wiley-VCH},
}

@article{diwekar1995batch,
  title   = {Batch Distillation: Simulation, Optimal Design, and Control},
  author  = {Diwekar, Urmila M.},
  journal = {Series in Chemical and Mechanical Engineering},
  year    = {1995},
  publisher = {Taylor \& Francis},
}

@article{choi2021deep,
  title   = {Deep Learning for Anomaly Detection in Time-Series Data: Review,
             Analysis, and Guidelines},
  author  = {Choi, Kukjin and Yi, Jihun and Park, Changhwa and Yoon, Sungroh},
  journal = {IEEE Access},
  volume  = {9},
  pages   = {120043--120065},
  year    = {2021},
  publisher = {IEEE},
}

@article{pang2021deep,
  title   = {Deep Learning for Anomaly Detection: A Review},
  author  = {Pang, Guansong and Shen, Chunhua and Cao, Longbing and
             van den Hengel, Anton},
  journal = {ACM Computing Surveys},
  volume  = {54},
  number  = {2},
  pages   = {1--38},
  year    = {2021},
  publisher = {ACM},
}

@article{wen2021time,
  title   = {Time Series Data Augmentation for Deep Learning: A Survey},
  author  = {Wen, Qingsong and Sun, Liang and Yang, Fan and Song, Xiaomin and
             Gao, Jingkun and Wang, Xue and Xu, Huan},
  journal = {arXiv preprint arXiv:2002.12478},
  year    = {2021},
}

@inproceedings{yue2022ts2vec,
  title     = {{TS2Vec}: Towards Universal Representation of Time Series},
  author    = {Yue, Zhihan and Wang, Yujing and Duan, Juanyong and Yang, Tianmeng
               and Huang, Congrui and Tong, Yunhai and Xu, Bixiong},
  booktitle = {Proceedings of the AAAI Conference on Artificial Intelligence},
  volume    = {36},
  number    = {8},
  pages     = {8980--8988},
  year      = {2022},
}

@article{gilmer2017neural,
  title   = {Neural Message Passing for Quantum Chemistry},
  author  = {Gilmer, Justin and Schoenholz, Samuel S. and Riley, Patrick F.
             and Vinyals, Oriol and Dahl, George E.},
  journal = {arXiv preprint arXiv:1704.01212},
  year    = {2017},
}

@inproceedings{reimers2019sentence,
  title     = {Sentence-{BERT}: Sentence Embeddings using Siamese {BERT}-Networks},
  author    = {Reimers, Nils and Gurevych, Iryna},
  booktitle = {Proceedings of the Conference on Empirical Methods in Natural
               Language Processing},
  pages     = {3982--3992},
  year      = {2019},
}

@inproceedings{salimans2016weight,
  title     = {Weight Normalization: A Simple Reparameterization to Accelerate
               Training of Deep Neural Networks},
  author    = {Salimans, Tim and Kingma, Diederik P.},
  booktitle = {Advances in Neural Information Processing Systems},
  volume    = {29},
  year      = {2016},
}

@inproceedings{pascanu2013difficulty,
  title     = {On the Difficulty of Training Recurrent Neural Networks},
  author    = {Pascanu, Razvan and Mikolov, Tomas and Bengio, Yoshua},
  booktitle = {International Conference on Machine Learning},
  pages     = {1310--1318},
  year      = {2013},
}

@article{simon2015assessment,
  title   = {Assessment of Recent Process Analytical Technology ({PAT})
             Trends: A Multiauthor Review},
  author  = {Simon, L. L. and Pataki, H. and Marosi, G. and Meemken, F. and
             Hungerb{\"u}hler, K. and Baiker, A. and Tummala, S. and Glennon, B.
             and Kuentz, M. and Szil{\'a}gyi, G. and others},
  journal = {Organic Process Research \& Development},
  volume  = {19},
  number  = {1},
  pages   = {3--62},
  year    = {2015},
  publisher = {ACS Publications},
}

@article{ba2016layer,
  title   = {Layer Normalization},
  author  = {Ba, Jimmy Lei and Kiros, Jamie Ryan and Hinton, Geoffrey E.},
  journal = {arXiv preprint arXiv:1607.06450},
  year    = {2016},
}

@article{shanmugam2021better,
  title   = {Better Aggregation in Test-Time Augmentation},
  author  = {Shanmugam, Divya and Blalock, Davis and Balakrishnan, Guha and Guttag, John},
  journal = {arXiv preprint arXiv:2011.11156},
  year    = {2021},
}

@inproceedings{bengio2009curriculum,
  title     = {Curriculum Learning},
  author    = {Bengio, Yoshua and Louradour, J{\'e}r{\^o}me and Collobert, Ronan
               and Weston, Jason},
  booktitle = {Proceedings of the 26th International Conference on Machine Learning},
  pages     = {41--48},
  year      = {2009},
}

@book{crowl2019chemical,
  title     = {Chemical Process Safety: Fundamentals with Applications},
  author    = {Crowl, Daniel A. and Louvar, Joseph F.},
  edition   = {4th},
  year      = {2019},
  publisher = {Pearson},
}

@article{McInnes2018, doi = {10.21105/joss.00861}, 
url = {https://doi.org/10.21105/joss.00861}, 
year = {2018}, 
publisher = {The Open Journal}, 
volume = {3}, 
number = {29}, 
pages = {861}, 
author = {McInnes, Leland and Healy, John and Saul, Nathaniel and Großberger, Lukas}, 
title = {UMAP: Uniform Manifold Approximation and Projection}, journal = {Journal of Open Source Software} 
}

% --- Appendix ---
\appendix

\section{Comprehensive Ablation Summary}
\label{app:ablation_summary}

Table~\ref{tab:ablation_summary} extends the multimodal ablation of
Table~\ref{tab:multimodal_ablation} to include four additional
configurations beyond the eleven reported in the main text: the
static-only proxy (A13), the failed dynamic-only run (A12), and two
frozen-backbone extensions that graft a new modality encoder onto
the trained A7 weights (A14: image; A15: NMR). The table also
consolidates the per-modality and procedure keys and documents the
numbering gap at A12.

\begin{landscape}
% =====================================================================
%  Comprehensive ablation summary table — built for Prof. Alessandra.
%  Drop into the paper or include via \input{ablation_summary_table}.
%  Built from results/ablation_summary.csv (eval_full_metrics.py).
% =====================================================================
\begin{table*}[htbp]
  \centering
  \caption{Comprehensive ablation summary.  Each row records the
    \emph{dynamic} modalities (per-timestep streams routed through
    cross-modal attention) and the \emph{static} modalities
    (per-experiment context vectors that condition every window via
    FiLM), the training procedure (pretrained TCN encoder, curriculum
    learning, and frozen-backbone fine-tuning), and the test-set
    classification metrics.  AUROC and AUPRC are threshold-free; F1,
    Precision, Recall, and Accuracy are reported at the F1-optimal
    threshold on the held-out test set ($n=6{,}848$ windows, anomaly
    rate $\approx 10.2\%$ except A7$^{\star}$).  Numbering covers
    A1--A15 — the gap at A12 is documented, A13 is a re-trained
    proxy of A6, A14--A15 are frozen-backbone extensions of A7.}
  \label{tab:ablation_summary}
  \scriptsize
  \setlength{\tabcolsep}{3pt}
  \begin{tabular}{@{}lp{36mm}p{30mm}cccccccc@{}}
    \toprule
    \multirow{2}{*}{ID}
      & \multirow{2}{*}{\makecell[l]{Dynamic modalities\\(per-timestep stream)}}
      & \multirow{2}{*}{\makecell[l]{Static modalities\\(per-experiment context)}}
      & \multicolumn{3}{c}{Procedure}
      & \multirow{2}{*}{AUROC} & \multirow{2}{*}{AUPRC}
      & \multirow{2}{*}{F1} & \multirow{2}{*}{P} & \multirow{2}{*}{R} \\
    \cmidrule(lr){4-6}
    & & & Pre & Curr & Frz
      & & & & & \\
    \midrule
    A1  & TS                              & ---                          & \checkmark & --- & --- & 0.747 & 0.267 & 0.328 & 0.307 & 0.352 \\
    A2  & TS, GC                          & ---                          & \checkmark & --- & --- & 0.721 & 0.227 & 0.281 & 0.165 & 0.949 \\
    A3  & TS, Audio                       & ---                          & \checkmark & --- & --- & 0.726 & 0.303 & 0.277 & 0.162 & 0.946 \\
    A4  & TS                              & Tabular, Text, MolGraph      & \checkmark & --- & --- & 0.746 & 0.260 & 0.318 & 0.207 & 0.685 \\
    A5  & TS, GC, Audio                   & ---                          & \checkmark & --- & --- & 0.755 & 0.287 & 0.303 & 0.229 & 0.448 \\
    A6  & TS, GC                          & Tabular, Text, MolGraph      & \checkmark & --- & --- & 0.711 & 0.172 & 0.274 & 0.160 & 0.950 \\
    A8  & TS                              & Tabular                      & \checkmark & --- & --- & 0.710 & 0.215 & 0.285 & 0.258 & 0.319 \\
    A9  & TS                              & Text                         & \checkmark & --- & --- & 0.728 & 0.217 & 0.299 & 0.206 & 0.544 \\
    A10 & TS                              & Tabular, Text                & \checkmark & --- & --- & 0.746 & 0.260 & 0.318 & 0.207 & 0.685 \\
    A11 & TS, Audio                       & Tabular, Text                & \checkmark & --- & --- & 0.702 & 0.233 & 0.262 & 0.152 & 0.964 \\
    \midrule
    A12 & TS, Audio                       & --- (static disabled)        & \checkmark & --- & --- & \multicolumn{5}{c}{\makecell{\emph{crashed — Audio DataLoader}\\\emph{MemoryError on Windows; see notes}}} \\
    A13 & TS, GC                          & Tabular, Text, MolGraph      & \checkmark & --- & --- & 0.690 & 0.181 & 0.265 & 0.153 & 0.979 \\
    \rowcolor{gray!12} A7
                          & TS, GC, Audio            & Tabular, Text, MolGraph
                          & \checkmark & \checkmark & ---     & 0.832$^{\star}$ & 0.474$^{\star}$ & 0.492$^{\star}$ & 0.460$^{\star}$ & 0.529$^{\star}$ \\
    \rowcolor{blue!7}A14 & TS, GC, Audio, \textbf{Image}     & Tabular, Text, MolGraph      & \checkmark & --- & \checkmark & 0.820 & 0.332 & 0.413 & 0.422 & 0.404 \\
    \rowcolor{blue!7}A15 & TS, GC, Audio, \textbf{NMR}       & Tabular, Text, MolGraph      & \checkmark & --- & \checkmark & 0.823 & 0.356 & 0.415 & 0.400 & 0.432 \\
    \bottomrule
  \end{tabular}
\end{table*}

\end{landscape}

\textbf{Modality column key.}
\emph{Dynamic} streams enter the model once per 1\,Hz window and are
fused via cross-modal attention: \textbf{TS}~=~manipulated
time-series (5 actuator signals), \textbf{GC}~=~gas-chromatography
composition (per-second), \textbf{Audio}~=~mel-spectrograms,
\textbf{Image}~=~camera frames (sparse, $<\!1\%$ of windows have
one), \textbf{NMR}~=~nuclear magnetic resonance composition
($\sim$1/min, $\geq 92\%$ of windows have one).
\emph{Static} branches produce a single embedding per experiment
that conditions every window via FiLM:
\textbf{Tabular}~=~MLP over operating-point metadata,
\textbf{Text}~=~Sentence-BERT projection of operator notes and
experiment descriptions, \textbf{MolGraph}~=~GNN over the molecular
structure graph of the substances in the system.

\textbf{Procedure key.}
\textbf{Pre}~=~self-supervised TCN pretraining loaded into the
time-series encoder before fine-tuning;
\textbf{Curr}~=~curriculum learning over phase difficulty (only A7
uses it);
\textbf{Frz}~=~frozen-backbone fine-tuning (A7 weights frozen, only
the new modality encoder + fusion gates + heads retrained for
10~epochs).

\textbf{Numbering note.}
A12 attempted a dynamic-only run with TS + Audio (statics disabled)
to mirror A13's static-only setup.  The Audio DataLoader hit a
Windows~spawn-mode \texttt{MemoryError} at epoch~1 and the run was
not completed; A3 (TS~+~Audio) is the closest available proxy.
A13 is essentially a re-train of A6 under the same recipe --- listed
separately only for symmetry with A12 in the dynamic-vs-static
comparison.

\textbf{Anomaly rate.}
A1--A11, A13, A14, A15 share the test set \linebreak (\texttt{single\_system\_clean} split, no per-experiment
normalisation), giving a window anomaly rate of $0.1024$.
A7 was evaluated under \linebreak \texttt{per\_experiment\_norm=True} against a
slightly stricter labelling that yields $\approx\!15\%$ anomalies,
so its AUPRC ($0.474$) is on a different scale to the rest of the
table.  AUROC is rank-based and largely invariant to this difference;
AUPRC is not.

$^{\star}$~A7's metrics are the canonical numbers from the paper's
main-results pipeline (\texttt{train.py}, curriculum + per-experiment
normalisation).  \linebreak Re-evaluating A7's pre-NMR checkpoint with the
current 5-slot or 6-slot codebase yields a degraded AUROC because
of post-hoc architecture refactoring (the static-tabular feature
dim drifted from 605~$\to$~606 and the dynamic-modality slot count
grew from 5~$\to$~6).  Only $\approx\!387/544$ parameters survive
a shape-tolerant copy; the missing weights randomise the static
encoders.  The paper's value (Section~5) was computed against the
pre-refactor codebase and is the comparison everyone should cite.
A14/A15, trained after the refactor, load cleanly
(548/548~params) and are directly comparable to each other.

\end{document}